\PassOptionsToPackage{usenames,dvipsnames}{xcolor}
\documentclass{agujournal2019}
\journalname{Water Resources Research}
\usepackage[pdftex,tightpage]{preview}
\usepackage{comment}
\usepackage{url} %this package should fix any errors with URLs in refs.
\usepackage[inline]{trackchanges} %for better track changes. finalnew option will compile document with changes incorporated.
\usepackage{soul}
\usepackage[utf8]{inputenc}
\usepackage[export]{adjustbox}
\usepackage{amsmath, amsfonts, amssymb, amsthm, bm, mathtools}
\usepackage{algorithm, algpseudocode}
\usepackage{natbib}
\usepackage{booktabs, tabu, multicol, multirow, makecell, bigstrut}
\usepackage{caption}
\usepackage[subrefformat=parens]{subcaption}
\usepackage{hyperref}
\usepackage[capitalise]{cleveref}
\usepackage{xstring}
\usepackage{etoolbox}
\usepackage{tikz, pgfplots, pgfplotstable}
\usetikzlibrary{calc, matrix}
\usepgfplotslibrary{groupplots, statistics}
\PreviewEnvironment{tikzpicture}
\PreviewEnvironment{tabu}
\definecolor{seagreen}{RGB}{46, 139, 87}
\definecolor{midnight}{RGB}{25,25,112}

\pgfplotsset{
    compat=newest,
    every axis/.append style={
		thick,
		line width=1pt,
		x tick label style={
		    /pgf/number format/.cd,
		    set thousands separator={}
		}
	},
	all plot/.style={
        width=0.85*\textwidth,
        height=0.3*\textheight,
	},
	error plot/.style={
        all plot,
        cycle list={
            Maroon,every mark/.append style={fill=Maroon,solid},mark=*\\%
            MidnightBlue,every mark/.append style={fill=MidnightBlue,solid},mark=square*\\%
        },
        xlabel={Number of KL terms},
        ylabel={$\ell_2$ errors},
	},
    trend plot/.style={
        all plot,
        cycle multi list={
            {mark=*}, {mark=square*}\nextlist
            MidnightBlue, {Maroon, draw=none}\nextlist
            only marks, mark=none
        },
        xlabel={Number of FV cells},
        ylabel={Execution time (s)},
        xtick=data,
        xticklabel={
            \pgfkeys{/pgf/fpu=true}
            \pgfmathparse{exp(\tick)}
            \pgfmathprintnumber[fixed relative, precision=4]{\pgfmathresult}
            \pgfkeys{/pgf/fpu=false}
        },
    },
}

\newcommand\setv[3]{\expandafter\xdef\csname #1_#2\endcsname{#3}}
\newcommand\getv[2]{\csname #1_#2\endcsname}

\BeforeBeginEnvironment{equation}{\begin{linenomath}}
\AfterEndEnvironment{equation}{\end{linenomath}}
\BeforeBeginEnvironment{equation*}{\begin{linenomath}}
\AfterEndEnvironment{equation*}{\end{linenomath}}
\BeforeBeginEnvironment{gather}{\begin{linenomath}}
\AfterEndEnvironment{gather}{\end{linenomath}}
\BeforeBeginEnvironment{align}{\begin{linenomath}}
\AfterEndEnvironment{align}{\end{linenomath}}
\draftfalse
\addeditor{Yu-Hong}
\addeditor{David}

\renewcommand{\caption}[2][]{\ignorespaces}

\begin{document}

\title{Physics-Informed Machine Learning Method for Large-Scale Data Assimilation Problems}

\authors{Yu-Hong Yeung,\affil{1},  David A. Barajas-Solano,\affil{1}, Alexandre M. Tartakovsky\affil{2,1}}

\affiliation{1}{Physical and Computational Sciences Directorate, Pacific Northwest National Laboratory, Richland, WA 99354}
\affiliation{2}{Department of Civil and Environmental Engineering, University of Illinois Urbana-Champaign, Urbana, IL 61801}

\correspondingauthor{A.M. Tartakovsky}{amt1998@illinois.edu}

%% STYLE GUIDANCE
% Yu-Hong, please try to use the \verb+cleveref+ package (already included) for managing references.
% For Sections, use labels of the form \verb+sec:section-name+.  For figures, \verb+fig:figure-name+, and for tables, \verb+tbl:table-name+.
% For bibliographical references, use labels of the form \verb+firstauthorlastname-year-firstwordintitle+ (e.g., \verb+barajassolano-2019-physics+).

\begin{keypoints}
\item The modified physics-informed machine learning PICKLE method for large-scale data assimilation is proposed.  
\item PICKLE method is orders of magnitude faster than traditional a posteriori probability method for the considered high-resolution Hanford model. 
\item Trained for one set of boundary conditions, the PICKLE method can model data for different values of the boundary conditions.      
\end{keypoints}

\begin{abstract}
We develop a physics-informed machine learning approach for large-scale data assimilation and parameter estimation and apply it for estimating transmissivity and hydraulic head in the two-dimensional steady-state subsurface flow model of the Hanford Site given synthetic measurements of said variables.
In our approach, we extend the physics-informed conditional Karhunen-Lo\'{e}ve expansion (PICKLE) method for modeling subsurface flow with unknown flux (Neumann) and varying head (Dirichlet) boundary conditions.
We demonstrate that the PICKLE method is comparable in accuracy with the standard \emph{maximum a posteriori} (MAP) method, but is significantly faster than MAP for large-scale problems.
Both methods use a mesh to discretize the computational domain.
In MAP, the parameters and states are discretized on the mesh; therefore, the size of the MAP parameter estimation problem directly depends on the mesh size.
In PICKLE, the mesh is used to evaluate the residuals of the governing equation, while the parameters and states are approximated by the truncated conditional Karhunen-Lo\'{e}ve expansions with the number of parameters controlled by the smoothness of the parameter and state fields, and not by the mesh size.
For a considered example, we demonstrate that the computational cost of PICKLE increases near linearly (as $N_{FV}^{1.15}$) with the number of grid points $N_{FV}$, while that of MAP increases much faster as $N_{FV}^{3.28}$.
We demonstrated that once trained for one set of Dirichlet boundary conditions (i.e., one river stage), the PICKLE method provides accurate estimates of the hydraulic head for any value of the Dirichlet boundary conditions (i.e., for any river stage).
\end{abstract}

\section{Introduction}
\label{sec:intro}

The ability of numerical models to predict a complex behavior of natural and engineered systems have been the main goal of computational sciences. However, when applied to natural systems such as subsurface flow and transport, predictive modeling is complicated by the inherent uncertainty in the distribution of subsurface properties, including hydraulic conductivity, that enter the subsurface models as parameters.
Uniquely estimating the subsurface parameters from the measurements of parameters and states (e.g., the hydraulic head) without numerical regularization is not possible because of the ill-posedness of the arising inverse problems.
To further complicate the matter, the multiple length scales of heterogeneity and time scales of flow and transport processes create enough ambiguity such that the same data can be described with different models, including  deterministic and stochastic partial differential equation (PDE) models, non-local (integro-differential equation) models, and, most recently, machine learning and artificial intelligence.

In selecting the right modeling approach, one can rely on the Occam's razor or law of parsimony principal that the simplest explanation (model) is usually the right one.
However, given the abovementioned uncertainty in model parameters, the ability of models to be conditioned on spatially varying data is another critical criterion in selecting a computational model~\citep{neuman2009perspective}.
In theory, any model that involves observable parameters and states can be conditioned on the measurements of these variables if such are available. However, the computational cost of conditioning models on data may vary significantly. 

Conditioning models on direct measurements of space-varying parameters (e.g., conductivity) is relatively straightforward and can be achieved using Kriging or Gaussian process regression (GPR)~\citep{neuman1993eulerian,Tipireddy2020JCP}.
In Kriging, an estimate of the conductivity field is obtained by means of a nonlinear interpolation between its measured values with the interpolation coefficients given in terms of the covariance function of the conductivity field learned from data~\citep{matheron1963principles,rasmussen2003gaussian}.
Then, the flow (conditioned on the measurements of conductivity) can be found by solving the Darcy flow equation with the conductivity field given by the Kriging estimate.
The advantage of using Kriging for conditioning flow models on conductivity measurements is that 
it provides a Bayesian prediction, including the conditional mean (the most likely distribution of conductivity given its measurements) and the conductivity variance (a measure of uncertainty).
%\note[David]{Most likely in which sense?}
The conditional mean and varience of conductivity can be used to obtain probabilistic estimates of the hydraulic head and fluxes conditioned on the conductivity measurements.
However, conditioning deterministic or stochastic flow predictions on the measurements of both the hydraulic head and conductivity is more challenging because it requires solving an inverse problem that typically involves computing forward solutions of the Darcy equation multiple times for different realizations of the conductivity fields. 

The inverse problem of computing the deterministic conductivity and hydraulic head fields given sparse measurements of these fields can be solved via \emph{maximum a posteriori} (MAP) estimation, a Bayesian point estimation approach that consists of computing the largest mode of the posterior density of conductivity 
conditioned on the observations~\citep{kitanidis1996geostatistical,barajassolano-2014-linear}.
A recent review of other methods for solving deterministic inverse problems can be found in~\cite{barajassolano-2019-pickle}.
Inverse uncertainty quantification problems, where the distributions of the parameters and states conditioned on the measurements are thought, are usually obtained with Bayesian methods~\citep{Langevin,Yoon,Barajas2019JCP,LI2020JCP}.
%\note[David]{``Probabilistic solutions of the Darcy equations'' implies forward solutions. There's nothing Bayesian about that.}

In this work, we use the physics-informed conditional Karhunen-Lo\'{e}ve expansion (PICKLE) method for obtaining deterministic estimates of transmissivity and head fields conditioned on the said fields measurements. We apply this method for modeling steady-state two-dimensional groundwater flow at the Hanford Site given synthetic measurements of the transmissivity and hydraulic head. The synthetic measurements are generated using the hydraulic conductivity measurements and boundary conditions obtained in the Hanford Site calibration study of \cite{cole2001transient}. 
The PICKLE method was recently introduced in~\cite{barajassolano-2019-pickle} for parameter estimation in PDE models for a given set of  deterministically known fixed boundary conditions.
Here, we extend the PICKLE method for problems with uncertain flux boundary conditions and incrementally changing Dirichlet boundary conditions.
In the  Hanford Site groundwater models,  the latter may change in response to changing water stages in the Columbia and Yakima Rivers.
The former boundary condition is used to describe recharge from the Cold Creek Valley and the Rattlesnake Springs that is difficult to estimate with a high degree of certainty.
Another significant contribution of this work is testing the PICKLE methods for high-dimensional realistic conductivity fields (we find that more than 1000 terms in the KL expansion are needed to accurately approximate the log-conductivity field obtained from the Hanford Site calibration study). 
We compare the performance of the PICKLE and MAP methods  and show that the two methods have a comparable accuracy, while the computational cost of MAP increases significantly faster with the problem size (the number elements $N_{FV}$ in the finite volume discretization of the governing equations) than that of PICKLE, i.e., $N_{FV}^{3.28}$ in MAP versus $N_{FV}^{1.15}$ in PICKLE. In the considered examples, we observe that for $N_{FV}=5900$, the computational time of PICKLE is one order of magnitude smaller than that of MAP. For $N_{FV}=23600$, we estimate that PICKLE would be more than two orders of magnitude faster than MAP (the computational time of PICKLE is found to be approximately $2\times 10^3$ s, and the computational time of MAP of approximately $5 \times 10^5$ s is estimated from the scaling relationship because it is unpractical to perform simulations for this time). 
The choice of synthetic (as opposite to field) measurements of the hydraulic head allows us to have the ground truth (reference) transmissivity and hydraulic head fields to compare the accuracy of the MAP and PICKLE methods, while preserving the complexity of boundary conditions and the transmissivity field of the Hanford Site.

\section{Groundwater flow model and maximum a posteriori formulation of the inverse problem}
\label{sec:problem}

We consider a two-dimensional model of groundwater flow at the Hanford Site.
Our objective is to learn the spatial distribution of transmissivity $T(x)\colon D \to \mathbb{R}^+$
and hydraulic head $u(x)\colon D \to \mathbb{R}$ given the sparse measurements of $T(x)$ and $u(x)$, where $D \subset \mathbb{R}^2$ is the simulation domain encompassing the Hanford Site. 
We assume that the flow is governed by the boundary value problem (BVP)
\begin{align}
  \label{eq:pde}
  \nabla \cdot \left [ T(x) \nabla u(x) \right ] & = 0, && x \in D,\\
  \label{eq:pde-flux-bc}
  T(x) \nabla u(x) \cdot n(x) &= -q_\mathcal{N}(x), && x \in \Gamma_\mathcal{N},\\
  \label{eq:pde-head-bc}
  u(x) &= u_\mathcal{D}(x), && x \in \Gamma_\mathcal{D},
\end{align}
where $\partial D = \Gamma_\mathcal{N} \cup \Gamma_\mathcal{D}$ is the boundary of $D$ and $\Gamma_\mathcal{N}$ and $\Gamma_\mathcal{D}$ ($\Gamma_\mathcal{N} \cap \Gamma_\mathcal{D} = \varnothing$) are the portions of the boundary where the Neumann and Dirichlet boundary conditions are prescribed, respectively.
In Eq.~(\ref{eq:pde-flux-bc}), $q_\mathcal{N} \colon \Gamma_\mathcal{N} \to \mathbb{R}$ is the normal flux at the Neumann boundary $\Gamma_\mathcal{N}$, and $n \colon \Gamma_\mathcal{D} \to \mathbb{R}^2$ is the unit vector normal to $\Gamma_\mathcal{N}$. In Eq.~(\ref{eq:pde-head-bc}), $u_\mathcal{D} \colon \Gamma_\mathcal{D} \to \mathbb{R}$ is the prescribed hydraulic head on the Dirichlet boundary $\Gamma_\mathcal{D}$.

In this work, we assume that there are $N_{\mathbf{u}_{\mathrm{s}}}$ and $N_{\mathbf{y}_{\mathrm{s}}}$ observations of $u$ and $y = \log T$, respectively, organized into the vectors $\mathbf{u}_{\mathrm{s}}$ and $\mathbf{y}_{\mathrm{s}}$.
The locations of $u$ and $y$ observations are organized into the arrays $\mathbf{X}_u$ and $\mathbf{X}_y$, respectively.
%The boundary conditions are assigned following the Hanford Site calibration study of~\cite{cole2001transient}.
At the Hanford Site, $\Gamma_\mathcal{D}$ models the boundary formed by the Columbia and Yakima Rivers and $u_\mathcal{D}(x)$ is equal to the water levels in these  rivers that are easy to measure. Therefore, $u_\mathcal{D}(x)$ is treated here as a known function.
The homogeneous Neumann boundary condition ($q_\mathcal{N}=0$) is imposed at the boundaries of formed by the (impermeable) basalt layers, and we also treat this boundary condition as known.
The non-homogeneous Neumann boundary conditions are used to describe inflow/outflow of the groundwater through the boundaries formed by the Cold Creek Valley, Dry Creek Valley and Rattlesnake Spring Recharge Area. In subsurface flow models, including the Hanford Site model, (non-zero) fluxes $q_\mathcal{N}$ are difficult to measure directly.
%In the Hanford Site calibration study, these fluxes were estimated alongside other parameters.
In this work, we consider two cases, one where $q_\mathcal{N}$ is known and another where $q_\mathcal{N}$ is unknown and is estimated along with $y(x)$ and $u(x)$.

The transmissivity field $T$ in \cref{eq:pde} is a differentiable function of $x$. Inverse problems for $T(x)$ are typically solved in the context of numerical models where \cref{eq:pde} is discretized on a mesh and the values of $T$ are estimated at a finite set of points on this mesh. As an example, here we consider the MAP method for estimating $T$ values in a numerical model based on the cell-centered finite volumes (FV) discretization of Eqs.~(\cref{eq:pde})--(\ref{eq:pde-head-bc}) using the two-point flux approximation (TPFA) of fluxes across cell faces. In this setting, the domain $D$ is discretized with $N$ FV cells, and the fields $u$ and $y$ are discretized into values located at the cell centers $\{ x_i \}_{i=1}^N$. The discrete field values are organized into the column vectors $\mathbf{u}$ and $\mathbf{y}$, respectively, of a length $N$. The coordinates of the cell centers are organized into the array $\mathbf{X}_c$. In the problem formulation described below, we consider a more general case where $q_\mathcal{N}$ is unknown. In this case, the unknown field $q_\mathcal{N}(x)$ is discretized into $N_q$ values located at the centroids of the cell faces corresponding to $\Gamma_\mathcal{N}$. These values are organized into the vector $\mathbf{q}$. After discretizing the BVP using the TPFA-FV method, we obtain the set of algebraic equations
\begin{equation}
  \label{eq:pde-discretized}
  \mathbf{l}(\mathbf{u}, \mathbf{y}, \mathbf{q}) = \mathbf{A}(\mathbf{y}, \mathbf{q}) \mathbf{u} - \mathbf{b}(\mathbf{y}, \mathbf{q}) = 0,
\end{equation}
with the stiffness matrix $\mathbf{A} \colon \mathbb{R}^N \times \mathbb{R}^{N_q} \to \mathbb{R}^{N \times N}$ and right-hand side $\mathbf{b} \colon \mathbb{R}^N \times \mathbb{R}^{N_q} \to \mathbb{R}^N$ defined in~\ref{sec:FVD}.
In Eq.~(\ref{eq:pde-discretized}), $\mathbf{l} \colon \mathbb{R}^N \times \mathbb{R}^N \times \mathbb{R}^{N_q} \to \mathbb{R}^N$ denotes the vector of discretized BVP residuals.
The entries of the $\mathbf{l}$ vector correspond to the FV mass balance for each of the $N$ cells.
The set of cells can be split into three sets: $\mathcal{N}$, the $N_{\mathcal{N}}$ cells adjacent to $\Gamma_\mathcal{N}$, $\mathcal{D}$, the $N_{\mathcal{D}}$ cells adjacent to $\Gamma_\mathcal{D}$, and $\mathcal{I} = [1, N] \setminus (\mathcal{D} \cup \mathcal{N})$ of cardinality $N_{\mathcal{I}} = N - N_{\mathcal{N}} - N_{\mathcal{D}}$.
Only the mass balance for the cells in the $\mathcal{N}$ set explicitly includes the $\mathbf{q}$ contributions; therefore, it follows that $\mathbf{q}$ enters into $\mathbf{l}$ only on the $\mathcal{N}$ entries, and that $\mathbf{l}_{\mathcal{I} \cup \mathcal{D}}$ does not depend directly on $\mathbf{q}$.

In the MAP method, the vectors of unknown parameters $\mathbf{y}$  and $\mathbf{q}$ are estimated by minimizing the $\ell_2$-norm of the discrepancy between observations and model predictions, that is,
\begin{equation}
  \label{eq:pde-constrained-opt-reg}
  \begin{aligned}
    \min_{\mathbf{u}, \mathbf{y}, \mathbf{q}} \quad & \frac{1}{2} \| \mathbf{u}_{\mathrm{s}} - \mathbf{H}_{\mathbf{u}} \mathbf{u} \|^2_2 + \frac{1}{2} \| \mathbf{y}_{\mathrm{s}} - \mathbf{H}_{\mathbf{y}} \mathbf{y} \|^2_2 + \frac{\gamma}{2} \| \mathbf{D} \mathbf{y} \|^2_2 + \frac{\gamma}{2} \| \mathbf{q} \|^2_2, \\
    \text{s.t.} \quad & \mathbf{l}(\mathbf{u}, \mathbf{y}, \mathbf{q}) = 0,
  \end{aligned}
\end{equation}
where $\mathbf{H}_{\mathbf{u}} \colon \mathbb{R}^{N_{u_s} \times N}$ and $\mathbf{H}_{\mathbf{y}} \colon \mathbb{R}^{N_{y_s} \times N}$ are observation matrices that downsample the vectors $\mathbf{u}$ and $\mathbf{y}$ into the vectors of $u$ and $y$ values at the locations  where $\mathbf{u}_{\mathrm{s}}$ and $\mathbf{y}_{\mathrm{s}}$ are measured. Specifically, $\mathbf{H}_{\mathbf{u}} = \mathbf{I}_N [\mathbf{X}_u, :]$ and $\mathbf{H}_{\mathbf{y}} = \mathbf{I}_N [\mathbf{X}_y, :]$ are submatrices of the identity matrix of dimension $N$, $\mathbf{I}_N$, whose rows are selected at the indices of the observation locations $\mathbf{X}_u$ and $\mathbf{X}_y$, respectively.
The inverse problem (\ref{eq:pde-constrained-opt-reg}) is ill-posed; therefore, it is necessary to introduce regularization penalties to the cost function.
Here, we choose to penalize the $\ell_2$-norm of the discrete gradient of the $y$ field, and the $\ell_2$-norm of $\mathbf{q}$
%The regularized inversion problem reads
%
%\begin{equation}
%  \label{eq:pde-constrained-opt-reg}
%  \begin{aligned}
%    \min_{\mathbf{u}, \mathbf{y}, \mathbf{q}} \quad & \frac{1}{2} \| \mathbf{u}_{\mathrm{s}} - \mathbf{H}_{\mathbf{u}} \mathbf{u} \|^2_2 + \frac{1}{2} \| \mathbf{y}_{\mathrm{s}} - \mathbf{H}_{\mathbf{y}} \mathbf{y} \|^2_2 + \frac{\gamma}{2} \| \mathbf{D} \mathbf{y} \|^2_2 + \frac{\gamma}{2} \| \mathbf{q} \|^2_2, \\
%    \text{s.t.} \quad & \mathbf{l}(\mathbf{u}, \mathbf{y}, \mathbf{q}) = 0,
%  \end{aligned}
%\end{equation}
%
with the regularization penalty coefficient $\gamma$. In the first regularization term, $\mathbf{D} \colon \mathbb{R}^{N_{\mathcal{I}} \times N}$ is the TPFA approximation of the gradient operator.
The estimates of $y$ and $q$ fields obtained from  Eq.~(\ref{eq:pde-constrained-opt-reg}) are equivalent to the mode of the posterior distributions of these fields in a Bayesian interpretation of the inverse problem, with the data misfit terms corresponding to a Gaussian log-likelihood, and the regularization penalties terms to a Gaussian log-prior.
Here, we use MAP to benchmark the PICKLE method.
The details of MAP implementation are given in~\ref{MAP}.

\section{PICKLE method for inverse problems}
\label{sec:pickle}

\subsection{Method formulation}

The PICKLE method was proposed in~\cite{barajassolano-2019-pickle} for solving inverse diffusion equations with unknown diffusion coefficients.
In PICKLE, the unknown parameter field $y(x)$ as well as the hydraulic head $u(x)$ are represented with the so-called conditional Karhunen-Lo\'{e}ve expansion (CKLEs)~\citep{tipireddy-2020} as
\begin{gather}
  \label{eq:ckle_u}
  u^c(x, \boldsymbol{\eta}) = \bar{u}^c(x) + \sum_{i=1}^{N_u} \phi_i^u (x) \sqrt{\lambda_i^u} \eta_i,\\
  \label{eq:ckle_y}
  y^c(x, \boldsymbol{\xi}) = \bar{y}^c(x) + \sum_{i=1}^{N_y} \phi_i^y(x) \sqrt{\lambda_i^y} \xi_i,
\end{gather}
where $\boldsymbol{\eta} = (\eta_1, \eta_2, \ldots)^\top$ and $\boldsymbol{\xi} = (\xi_1, \xi_2, \ldots)^\top$ are the vectors of unknown CKLE coefficients, and the eigenpairs $\{\phi_i^u(x), \lambda_i^u\}_{i=1}^{N_u}$ and $\{\phi_i^y(x), \lambda_i^y\}_{i=1}^{N_y}$ are the solutions of the eigenvalue problems
\begin{equation}
  \label{eigenproblems}
  \int_D C^c_u(x; x')\phi^u(x') dx' = \lambda^u \phi^u(x), \quad
  \int_D C^c_y(x; x')\phi^y(x') dx' = \lambda^y\phi^y(x).
\end{equation}
Here, the superscript $c$ denotes conditioning on the measurements of $y$. Methods for computing $\bar{y}^c(x)$, $\bar{u}^c(x)$, $C^c_u(x; x')$, and $C^c_y(x; x')$ are described in Section \ref{covariances}.

The number of KL terms $N_u$ and $N_y$ are selected to satisfy the following conditions:
\begin{equation}
  \label{eq:truncation}
  \sum^{\infty}_{i = N_u + 1} \lambda_i^u \leq \text{rtol}_u \int_D  C^c_u(x,x) \, \mathrm{d} x, \quad \sum^{\infty}_{i = N_y + 1} \lambda_i^y \leq \text{rtol}_y \int_D C_y^c(x,x) \, \mathrm{d} x.
\end{equation}
For the eigenproblems solved via the eigendecomposition of the covariance matrices $C^c_y$ and $C^c_u$ evaluated on the cell-centered FV scheme with $N$ cells, these conditions can be approximated as
\begin{equation}
  \label{eq:truncation_disc}
  \sum^{N}_{i = N_u + 1} \lambda_i^u \leq \text{rtol}_u \sum^{N}_{i = 1} \lambda_i^u, \quad \sum^{N}_{i = N_y + 1} \lambda_i^y \leq \text{rtol}_y \sum^{N}_{i = 1} \lambda_i^y.
\end{equation}

The inverse problem is solved by minimizing the residual $\mathbf{l}(\mathbf{u}, \mathbf{y}, \mathbf{q})$ with the substitutions of $\mathbf{u}$ and $\mathbf{y}$
in Eq.~(\ref{eq:pde-discretized}) by the CKLEs (\ref{eq:ckle_u}) and (\ref{eq:ckle_y}) evaluated at the FV cell centroids.
Together with regularization penalty $\mathcal{R}$ on $\hat{\mathbf{y}}(\boldsymbol{\xi})$ and $\hat{\mathbf{u}}(\boldsymbol{\eta})$, as well as
a data misfit for $u$ predictions, the minimization problem becomes
\begin{equation}
  \label{eq:pickle-general}
  \min_{\bm{\xi}, \bm{\eta}, \mathbf{q}} \quad \frac{1}{2} \left \| \mathbf{l} \left [ \hat{\mathbf{u}}(\bm{\eta}), \hat{\mathbf{y}}(\bm{\xi}), \mathbf{q} \right ] \right \|^2_2 + \frac{\beta}{2} \left \| \mathbf{u}_{\mathrm{s}} - \mathbf{H}_{\mathbf{u}} \hat{\mathbf{u}}(\bm{\eta}) \right \|^2_2 + \frac{\alpha}{2} \mathcal{R}(\hat{\mathbf{y}}(\boldsymbol{\xi}),\hat{\mathbf{u}}(\boldsymbol{\eta})),
\end{equation}
where $\hat{\bm{u}}$ and $\hat{\bm{y}}$ denote the PICKLE estimates of $\bm{u}$ and $\bm{y}$, respectively.
Here, we consider two forms of $\mathcal{R}$. The first form,
\begin{equation}
 \label{eq:pickle-l2reg}
\mathcal{R} = \left \| \bm{\xi} \right \|^2_2 + \left \| \bm{\eta} \right \|^2_2,
\end{equation}
was proposed in \cite{barajassolano-2019-pickle} and shown to perform well when the reference $y$ field is generated as a realization of a Gaussian field. We also consider a form of $\mathcal{R}$ that penalizes the gradients of $\hat{\bm{u}}$ and $\hat{\bm{y}}$ (computed with the approximate gradient operator $\mathbf{D}$) as in the MAP method:
\begin{equation}
    \label{eq:pickle-h1reg}
\mathcal{R}= \left \| \mathbf{D} \hat{\mathbf{y}} \right \|^2_2 + \left \| \mathbf{D} \hat{\mathbf{u}} \right \|^2_2.
\end{equation}
% and $\hat{\mathbf{u}}_m(\bm{\eta})$ is the vector of $\hat{{u}}(x,\bm{\eta})$ evaluated at $\mathbf{X}_u$.

The PICKLE formulation (\ref{eq:pickle-general}) is different from that of~\cite{barajassolano-2019-pickle} because it allows for unknown flux boundary conditions.
Below, we propose a cost-effective treatment of this problem.  
% has the following additional features compared to the PICKLE formulation presented in~\cite{barajassolano-2019-pickle}:
As noted in Section \ref{sec:problem}, the FV mass balances for the subset $\mathcal{I} \cup \mathcal{D}$ of cells (which includes the vast majority of cells) do not directly involve the vector of normal fluxes $\mathbf{q}$.
This allows us to exclude $\mathbf{q}$ from the inverse problem by penalizing the $\ell_2$-norm of $\mathbf{l}_{\mathcal{I} \cup \mathcal{D}}$, i.e., to only minimize residuals within $\mathcal{I}$ and $\mathcal{D}$ cells:
\begin{equation}
  \label{eq:pickle-final}
  \min_{\bm{\xi}, \bm{\eta}} \quad \frac{1}{2} \left \| \mathbf{l}_{\mathcal{I} \cup \mathcal{D}} \left [ \hat{\mathbf{u}}(\bm{\eta}), \hat{\mathbf{y}}(\bm{\xi}) \right ] \right \|^2_2 + \frac{\beta}{2} \left \| \mathbf{u}_{\mathrm{s}} - \mathbf{H}_{\mathbf{u}} \hat{\mathbf{u}} (\bm{\eta}) \right \|^2_2 +
  \frac{\alpha}{2} \mathcal{R}(\hat{\mathbf{y}}(\boldsymbol{\xi}),\hat{\mathbf{u}}(\boldsymbol{\eta})).
\end{equation}

% This simplifies the CKLE-based inverse problem as it is then not necessary to construct a KL or CKL model for $\mathbf{q}$.
A PICKLE estimate of $\mathbf{q}$ is then obtained by solving
\begin{equation}
  \label{eq:pickle-estimate-q-indirect}
  \mathbf{l}_{\mathcal{N}} \left[ \hat{\mathbf{u}}(\hat{\bm{\eta}}), \hat{\mathbf{y}}(\hat{\bm{\xi}}), \hat{\mathbf{q}} \right] = 0,
\end{equation}
where $\hat{\bm{\xi}}$ and $\hat{\bm{\eta}}$ are the solutions of the minimization problem~(\ref{eq:pickle-final}).
We note that $\mathbf{l}_{\mathcal{N}}$ is linear in $\mathbf{q}$, so that the solution of Eq.~(\ref{eq:pickle-estimate-q-indirect}) is trivial. 
  
Because we use a CKLE model $\hat{\mathbf{y}}(\bm{\xi})$ for $\mathbf{y}$, it satisfies the $y$ observations by construction.
The $u$ field is approximated with the KLE model $\hat{\mathbf{u}}(\bm{\eta})$ that is not conditioned on the $u$ measurements.
Therefore, we have included a data misfit term with coefficient $\beta$ in Eq~(\ref{eq:pickle-final}) penalizing the deviation between $u$ predictions and observations.
  
We use the Trust Region Reflective algorithm~\citep{branch1999} to solve minimization problems in both PICKLE and MAP methods.
  To do this, we cast both Eqs.~(\ref{eq:pde-constrained-opt-reg}) and (\ref{eq:pickle-final}) as least-squares problems.
%We initialize the L-M algorithm by setting unknown parameters to zero.
%We find that a random initialization does not lead to better results.
The least-squares minimization algorithm requires the evaluations of the Jacobian matrix $\mathbf{J}$ of the objective vector, which is also the most computationally demanding part of the least-squares minimization.
Jacobian evaluation in the PICKLE method only requires computing the derivatives of the PDE residuals with respect to the CKLE coefficients. This is in general significantly less expansive  than solving the BVP and the corresponding adjoint problem in the MAP's Jacobian evaluation procedure ~\citep{barajassolano-2014-linear}.
  In addition, the cost of Jacobian evaluation depends on the Jacobian matrix size, which is $(N_{\mathcal{I}} + N_{u_s} + N_{u_y}) \times N$ in MAP ($N$ is the number of the FV cells) and $(N + N_{u_s} + N_u + N_y) \times (N_u + N_y)$ in PICKLE.
  Depending on the field smoothness and the required resolution, $N_u$ and $N_y$ can be much smaller than $N$.
The details of the optimization algorithm are given in~\ref{sec:optimization}.

%Finally, we note that the PICKLE method yields a functional approximation of the unknown parameter $y(x)$ rather than a set of $y$ values as in standard inverse methods, such as MAP. However, the PICKLE functional approximation in general depends on the mesh size that is used to evaluate the residual vector $\mathbf{l}$. Also, in this work we compute the eigenfunctions in the KL expansion of a mesh which allows evaluating $y(x)$ on the same mesh.
  
\subsection{Computing covariance functions}\label{covariances}

To construct $C^c_y(x,y)$, we first estimate the unconditional covariance of $y$, $C_y(x,y)$ by assuming it has the $5/2$-Mat\'{e}rn kernel
\begin{equation*}
  C_y(x,y) = \sigma^2 \left(1 + \sqrt{5}\frac{|x-y|}{\lambda} + \frac{5}{3}\frac{|x-y|^2}{l^2}\right) \exp\left(- \sqrt{5} \frac{|x-y|}{l}\right),
\end{equation*}
where $\sigma$ and $l$ are the standard deviation and the correlation length of $y$ that are estimated from $\mathbf{y}_{\mathrm{s}}$ by minimizing the marginal log-likelihood function~\citep{rasmussen2003gaussian}.
The conditional mean and covariance of $y$ are then computed from the GPR (or Kriging) equations
\begin{align}
  \label{eq:gpr-mean}
  \bar{y}^c(x) &=  \mathbf{C}(x) \mathbf{C}^{-1}_{\mathrm{s}}  \mathbf{y}_{\mathrm{s}},\\
  \label{eq:gpr-var}
  C^c_y(x, x') &= C_y(x, x') - \mathbf{C}(x) \mathbf{C}^{-1}_{\mathrm{s}} \mathbf{C}(x'),
\end{align}
where $\mathbf{C}_{\mathrm{s}}$ is the $N_{\mathbf{y}_{\mathrm{s}}} \times N_{\mathbf{y}_{\mathrm{s}}}$ observation covariance matrix with elements $C_{\mathrm{s},ij} = C_y(x_i,x_j)$ and $\mathbf{C}(x)$ is the  $N_{\mathbf{y}_{\mathrm{s}}}$-dimensional vector with the components  $C_i(x)= C_y(x,x_i)$, where $x_i, x_j \in\mathbf{X}_y$.  
The evaluation of $\overline{u}^c$ and $C_u^c(x,x')$ from $u$ measurements using the marginal likelihood maximization is not adequate for two related reasons: (1) the $u(x)$ field is not stationary (i.e., the covariance kernel of $u$ depends on $x$ and $y$ and not just the distance between $x$ and $y$ as in, e.g., the Mat\'{e}rn kernel), and guessing a nonstationary covariance kernel for $u$ and then training it could be very challenging; and (2) this purely data-driven approach does not enforce the governing equations and boundary conditions on the mean and covariance function even approximately. 
Therefore, in this work we employ a Monte Carlo (MC) simulation-based method for computing the conditional mean and covariance of $u$.

In the MC method, we treat the partially known $u(x)$ and $y(x)$ as random variables $\tilde{u}^c(x,\omega)$ and $\tilde{y}^c(x,\omega)$, $\{ \tilde{u}^c, \tilde{y}^c \} : D \times \Omega \to \mathbb{R}$ (where $\Omega$ is the corresponding random outcome space) conditioned on observed measurements of $y$.
We model $\tilde{y}^c(x,\omega)$ using the \emph{stochastic} truncated CKLEs
\begin{equation}
  \label{eq:ckle_y_stoch}
  \tilde{y}^c(x, \tilde{\boldsymbol{\xi}}(\omega)) = \bar{y}^c(x) + \sum_{i=1}^{N'_y} \phi_i^y(x) \sqrt{\lambda_i^y} \tilde{\xi}_i(\omega),
\end{equation}
where $\tilde{\boldsymbol{\xi}}(\omega) = (\tilde{\xi}_1(\omega); \tilde{\xi}_2(\omega); \ldots)^\top$ is the vectors of independent and identically distributed Gaussian random variables. The eigenpairs $\{\phi_i^y(x); \lambda_i^y\}_{i=1}^{N'_y}$ satisfy the same eigenvalue problem (\ref{eigenproblems}) as those in the deterministic CKLE of Eq.~(\ref{eq:ckle_y}). We note that, $N'_y$ does not need to be the same as $N_y$ in Eq.~(\ref{eq:ckle_u}), e.g., it can be chosen with smaller $rtol_y$ to obtain a more accurate MC solution. However, in this study we set $N'_y=N_y$. 
Next, we construct an ensemble of $N_{\mathrm{ens}}$ realizations of $\tilde{y}^c$, $\{ {y}^{c,(i)} \}^{N_{\mathrm{ens}}}_{i = 1}$ by sampling $\bm{\xi}^{(i)}$ from $\mathcal{N}(0, \mathbf{I}_{N'_y})$ and evaluating the CKLE model~(\ref{eq:ckle_y_stoch}) with $\tilde{\bm{\xi}} = \bm{\xi}^{(i)}$.
% , that is,
% \begin{equation}
%     \label{eq:ckle_y_real}
%     {y}^{c,(i)}(x, \boldsymbol{\xi}^{(i)}) = \bar{y}^c(x) + \sum_{j=1}^{N'_y} \phi_j^y(x) \sqrt{\lambda_j^y} {\xi}_j^{(i)}.
% \end{equation}
% %

%\subsection{Boundary Conditions}
The Dirichlet and Neumann boundaries $\Gamma_\mathcal{D}$ and $\Gamma_\mathcal{N}$ are defined in the Hanford Site calibration study and are shown in Figure~\ref{fig:hanford_site}. This calibration study also provides the estimates of the head $u_\mathcal{D}(x)$ and and fluxes $q_\mathcal{N}$ at these boundaries.
In this work we assume that $u_\mathcal{D}(x)$ is known (in Sections~\ref{sec:RF1_field} and~\ref{sec:RF2_field}, $u_\mathcal{D}(x)$ is given by the aforementioned calibration study, and in Section~\ref{sec:varying_Dirichlet}, $u_\mathcal{D}(x)$ is modified from the calibrated values to simulate the changing water levels in the Columbia and Yakima Rivers).
In cases with unknown $q_\mathcal{N}$, we assume that $q_\mathcal{N}$ is a Gaussian random field with the known mean and variance that we compute from the $q_\mathcal{N}$ values in the aforementioned calibration study. %In the case study where we assume that $q_\mathcal{N}(x)$ is known, we take the values of $q_\mathcal{N}$ from the calibration study.

Next, we generate $N_{\text{ens}}$ random realizations of $q_\mathcal{N}(x)$, $\{ q_\mathcal{N}^{(i)}(x) \}^{N_{\mathrm{ens}}}_{i = 1}$.  
For each member of the ensemble $ y^{c,(i)}$ and the corresponding Neumann boundary condition $q_\mathcal{N}^{(i)}(x)$, we calculate $u^{(i)}$ by solving the PDE problem (\ref{eq:pde})--(\ref{eq:pde-head-bc}).
The resulting ensemble $\{ u^{(i)} \}^{N_{\mathrm{ens}}}_{i = 1}$ is used to compute the $N$-dimensional vector of mean values of $u$ in each FV cell,
\begin{equation}
  \label{eq:ens_mean}
  \overline{u}^c(x_j) = \frac{1}{N_{\mathrm{ens}}} \sum_{i=1}^{N_{\mathrm{ens}}} u^{(i)}(x_j),\quad x_j\in\mathbf{X}_c,
\end{equation}
and the $N \times N$ covariance matrix of $u$ with elements
\begin{equation}
  \label{eq:ens_covariance}
  C^c_u(x_p,x_q) = \frac{1}{N_{\mathrm{ens}} - 1} \sum_{i=1}^{N_{\mathrm{ens}}} \left [ u^{(i)}(x_p) - \overline{u}(x_p) \right ] \left [ u^{(i)}(x_q) - \overline{u}(x_q) \right ], \quad(x_p,x_q)\in\mathbf{X}_c.
\end{equation}
%
%Here, the superscript $c$ denotes that the mean and covariance of $u$ are conditioned on the measurements of $y$.

%We note that the accuracy of the mean and covariance estimates of $u$ depends on $N_{\mathrm{ens}}$. This is important because the mean and covariance estimates of $u$ are used to construct the KLE approximation of $u$. 
In this work, we set  $N_{\mathrm{ens}}$ large enough to assure that the PICKLE estimates of $y$ do not change with further increase of $N_{\mathrm{ens}}$. 
In general, for $C^c_u(x_p,x_q)$ to have at least $N_u$ non-zero eigenvalues $\lambda_i^u$, the ensemble size should be $N_{\mathrm{ens}} > N_{\mathbf{u}_s}$. 
When it is not feasible to perform $N_{\mathrm{ens}} > N_{\mathbf{u}_s}$ MC simulations, shrinkage estimators can be employed to regularize the covariance matrix estimation~\citep{chen_2009_shrinkage}.
The accuracy of the covariance estimation with a small number of MC simulations can be increased by performing additional less-expensive coarser-resolution simulations using the Multilevel MC approach~\citep{giles-2015-multilevel,yang2018physics,Yang2019JCP}.
Also, there are several computationally efficient alternatives to MC methods, including the moment equation method (e.g., \citep{neuman1993eulerian,Tart2003JH,Jarman2013JUQ}) and polynomial-chaos-based approaches~\citep{Lin2010JSC,Tipireddy2020JCP,LI2020JCP}, surrogate models~\citep{YangJUQ2018}, and generative physics-informed machine learning methods~\citep{yang2019highly}. 

\section{Numerical Experiments}
\label{sec:experiments}

\subsection{Synthetic data sets}

We compare the performance of the PICKLE and MAP methods for parameter and state estimation in the steady-state two-dimensional groundwater model of the Hanford Site. In this comparison study, we use two
reference transmissivity fields $y(x)=\ln T(x)$ and boundary conditions $u_\mathcal{D}(x)$ and $q_\mathcal{N}(x)$ that are based on the three-dimensional Hanford Site calibration study~\citep{cole2001transient}.
\begin{figure}
    \centering%
    \begin{tabu} to 0.97\textwidth {*{2}{X[cm]}}
    %\begin{subfigure}[b]{0.45\textwidth}
        \includegraphics[width=0.45\textwidth]{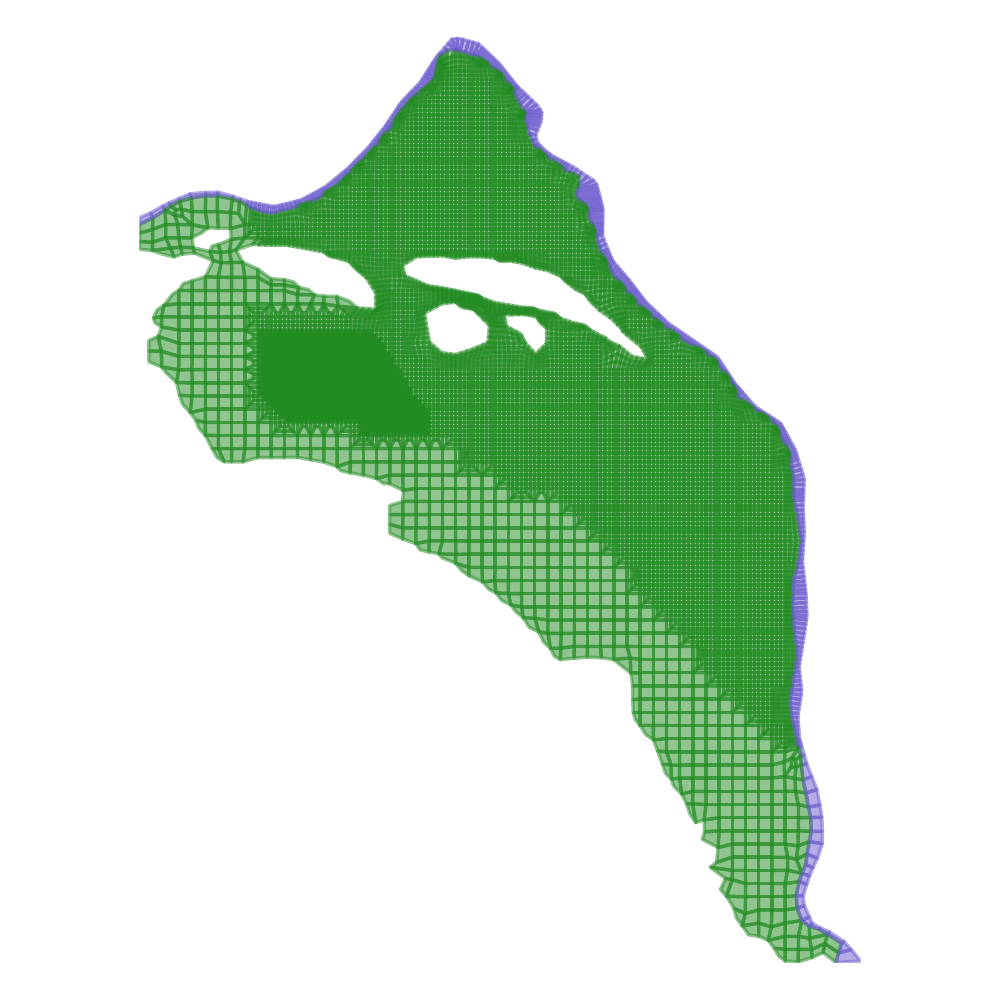}
        %\caption{\label{fig:hanford_preprocess}}
    %\end{subfigure}
    %\hfill
    %\begin{subfigure}[b]{0.45\textwidth}
    &
        \includegraphics[width=0.45\textwidth]{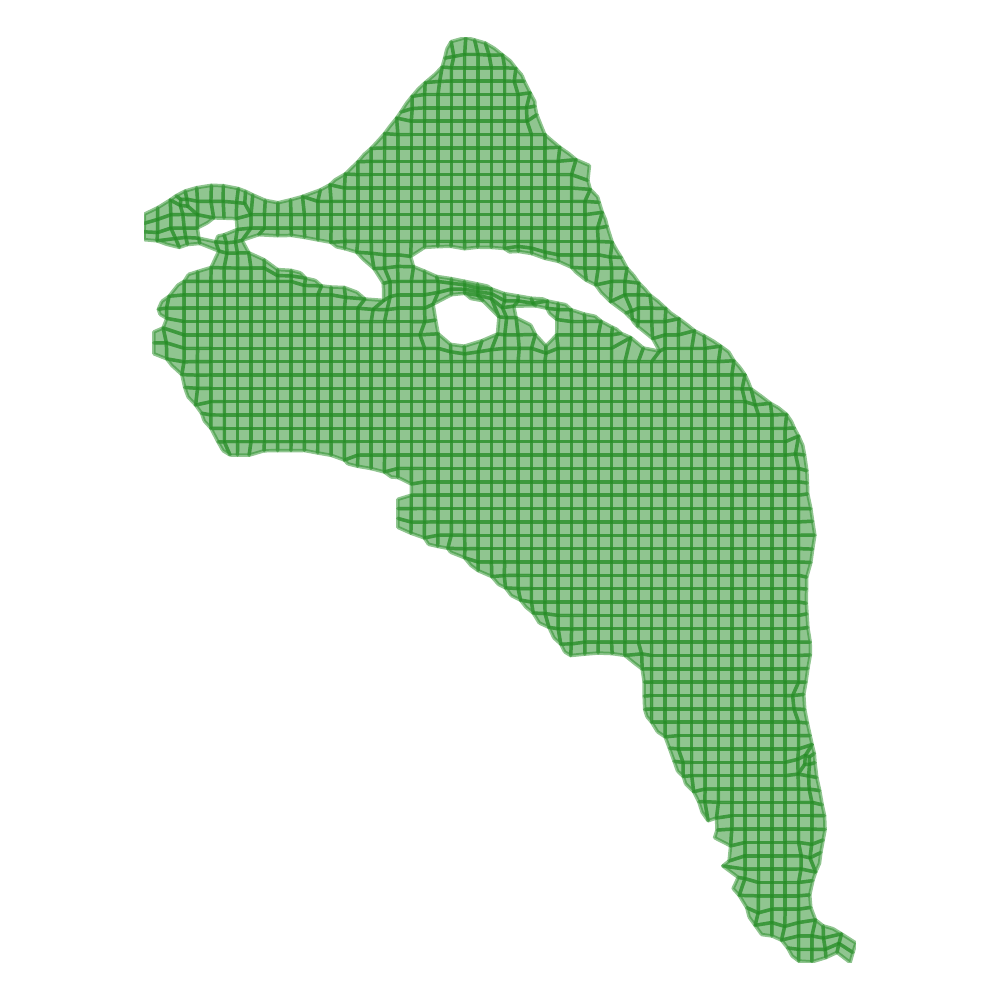}\\
        %\caption{\label{fig:hanford_postprocess}}
    (a) & (b)
    %\end{subfigure}
    \end{tabu}
    \caption{\subref{fig:hanford_preprocess} Mesh of the Hanford site subsurface flow model in the calibration study of \cite{cole2001transient}. Cells that represent Columbia River are highlighted in blue with the river head boundary conditions prescribed at the vertices of these cells. \subref{fig:hanford_postprocess} A quasi-uniform coarse mesh with the $N_{FV}=1475$ cells used in this study.}
    \label{fig:hanford_geometry}
\end{figure}
\begin{comment}
\begin{figure}
    \centering
    \includegraphics{pickle-hanford-paper/figures/Fig_Hanford.pdf}
    \caption{(a) Mesh of the Hanford Site subsurface flow model in the calibration study of \cite{cole2001transient}. Cells that represent the Columbia River are highlighted in blue with the river head boundary conditions prescribed at the vertices of these cells. (b) A quasi-uniform coarse mesh with the $N_{FV}=1475$ cells used in this study.}
    \label{fig:hanford_geometry}
\end{figure}
\end{comment}
%
This calibration study was performed on the unstructured quadrilateral grid shown in Figure~\ref{fig:hanford_geometry}a with 4 to 17 horizontal layers depending on the Cartesian plane coordinates and produced an estimate of the three-dimensional conductivity field. 
We obtain the first reference transmissivity field by depth averaging the conductivity field over the unstructured mesh.  

\begin{figure}
    \centering
    \includegraphics{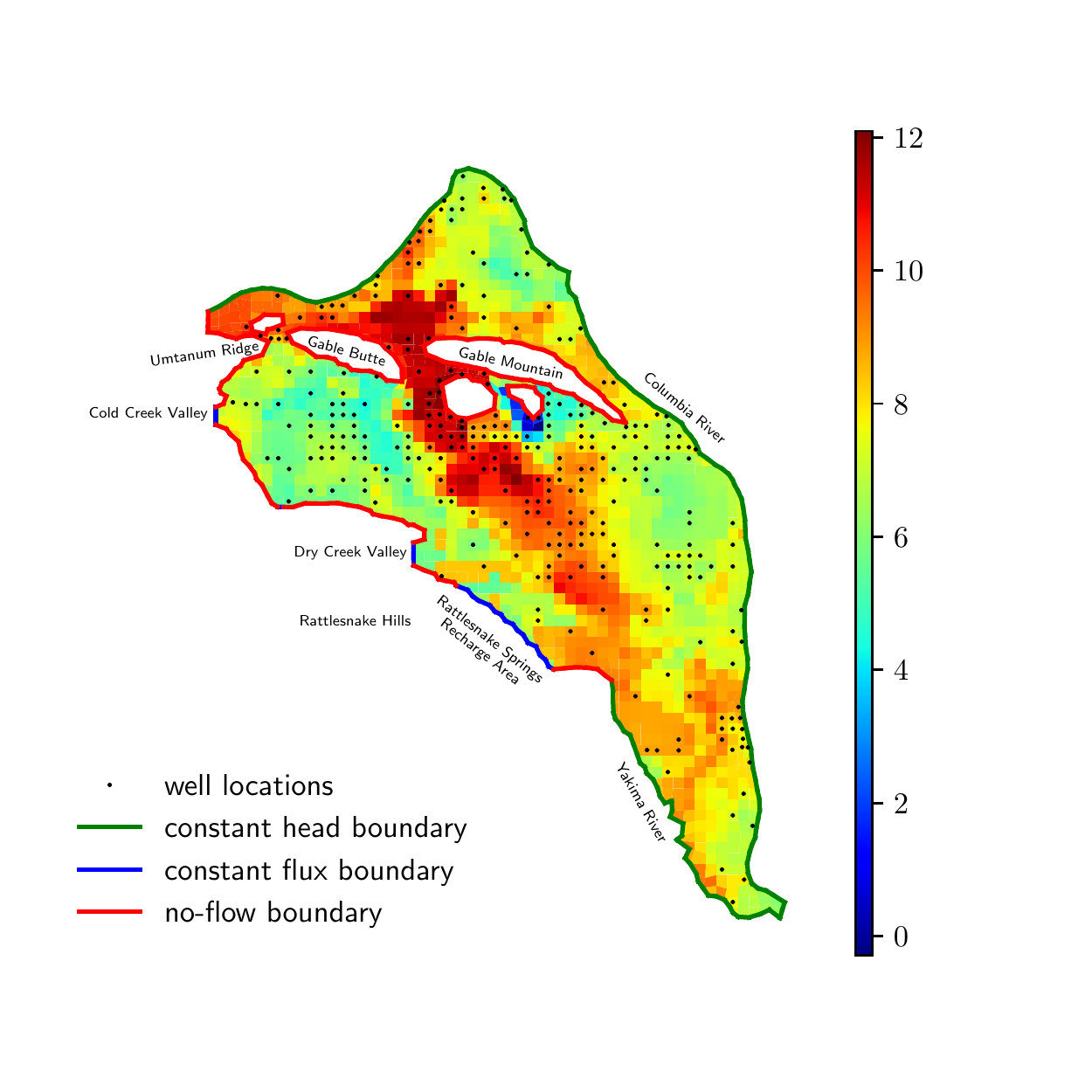}
    \caption{The coarse-resolution ($N_{FV}=1475$) RF1 $y$ field, well locations, and the parts of boundaries where different types of boundary conditions are prescribed.}
    \label{fig:hanford_site}
\end{figure}

The original lateral mesh contains three different mesh resolutions and includes both the western and eastern banks of the Columbia River (the ``Columbia River'' cells are highlighted in blue in Figure~\ref{fig:hanford_geometry}a).
We simplify the mesh by removing the river cells and prescribing Dirichlet BC on the western side of the Columbia River and by coarsening the mesh to achieve a uniform resolution, as shown in Figure~\ref{fig:hanford_geometry}b. 
For mesh coarsening, we use a semi-automatic algorithm to merge groups of finer cells into a single cell while maintaining that the mesh is boundary-conforming and each cell is a quadrilateral. The resulting mesh has 1475 cells. The transmissivity of each coarse cell is computed as the geometric average of the transmissivities of the replaced finer cells. %, except when multiple non-colinear edges are merged into a single straight edge.
%The latter helps us to avoid possible artifacts in parameter estimation caused by a multiresolution mesh.
%\note[David]{Why this sentence?}
%Next, we recompute the transmissivity field on this uniform mesh as local averages over finer cells that are replaced by the coarse cell. by replacing 
The transmissivity field corresponding to the coarse mesh is shown in Figure~\ref{fig:hanford_site}.
We refer to this field as \emph{reference field 1} (``RF1'').
We note that the PICKLE method can employ the FV (as in this study) or finite elements discretization to evaluate the residuals and, therefore, can utilize a multiresolution mesh.

Figure~\ref{fig:hanford_site} also shows the locations of some of the wells at the Hanford Site. We note that the calibration study ~\citep{cole2001transient} gives coordinates of $558$ wells at the Hanford Site but that some of these wells are located in the same coarse or fine cells.
Because our model uses exclusively cells and not points to denote spatial locations, multiple wells are considered as one measurement if they are located in the same cell. As a result, there are 323 in the FV model shown in Figure~\ref{fig:hanford_site}.  

% Although both PICKLE and MAP can handle multiple resolutions in a mesh, it is more desirable to make the mesh more uniform to reduce the number of cells in the higher resolution regions and the
 %

%\subsection{Higher Resolution Meshes}
We hypothesize that the accuracy of the PICKLE method depends on the smoothness of the reference transmissivity field. To test this hypothesis, we generate the \emph{reference field 2} (``RF2'') transmissivity field in addition to RF1 field using GPR (\cref{eq:gpr-mean}) and
50 measurements drawn from the RF1 field at the locations randomly picked from the locations of the wells. By construction, the RF2 field is smoother than the RF1 field. 

We also study the performance of PICKLE relative to MAP as a function of the size of the FV model resolution.
For this, we generate a higher-resolution mesh by splitting each cell in the mesh in Figure~\ref{fig:hanford_geometry}b ($1\times$ resolution) into four ($4\times$ resolution) equiareal cells, resulting in 5,900 cells.  We note that there are  $408$ wells at this resolution. 

%Then, we study the performance of the PICKLE and MAP methods for estimating $y$ field using the measurements drawn from the RF2 field using the coarse and fine FV discretization of the flow equation.

For each reference transmissivity field $y_{ref}(x) = \ln T_{ref}(x)$,  we generate the hydraulic head field $u_{ref}(x)$ by solving the Darcy flow equation on the corresponding mesh with the deterministic Dirichlet and Neumann boundary conditions from the calibration study ~\citep{cole2001transient}.
Then, we randomly pick $N_{\mathbf{y}_{\mathrm{s}}}$ well locations and treat the values of $y_{ref}$ at these locations as $y$ measurements.
Similarly, we draw $N_{\mathbf{u}_{\mathrm{s}}}$ measurements of the hydraulic head $u$ from $u_{ref}(x)$. These measurements are treated as synthetic data sets and used in the PICKLE and MAP methods to estimate the $y(x)$ and $u(x)$ fields.   

All PICKLE and MAP simulations are performed using a 3.2~GHz 8-core Intel Xeon W CPU and 32~GB of 2666~MHz DDR4 RAM. Codes are written in Python using the NumPy and SciPy packages.
%\note[David]{This last bit is not true. It doesn't follow from using numpy/scipy. The \texttt{lm} solver does only use one core, but \texttt{trf} uses multiple cores. I suggest we remove this sentence and we remove ``8-core'' above.} \note[Yu-Hong]{I removed the single-core sentence. It was for the \texttt{lm} method. I keep the ``8-core'' term since it's needed to reproduce the timing results.}

The weights in the PICKLE and MAP minimization problems are empirically found to minimize the error with respect to the reference $y$ fields as $\beta=10$, $\alpha=10^{-4}$, and $\gamma=10^{-4}$. When a reference field is not known, these weights could be found using the standard cross-validation methods \citep{picard1984cross}.

\subsection{RF1 reference field}\label{sec:RF1_field}

First, we use PICKLE to estimate $y$ and $u$ on the coarse mesh with $1475$ cells with the measurements of $y$ and $u$ drawn from the synthetic set generated with the RF1 reference field.
We start with the unknown Neumann boundary condition case.
The number of terms in the KL expansions of $y$ and $u$ are set to $N_y=1000$ and $N_u=1000$, respectively.
The corresponding relative tolerances are for these choices of $N_y$ and $N_u$ are 
$\text{rtol}_u =  \sum^{N}_{i = N_u + 1} \lambda_i^u / \sum^{N}_{i = 1} \lambda_i^u = 6.4 \times 10^{-6}$
and
$\text{rtol}_y =  \sum^{N}_{i = N_y + 1} \lambda_i^y / \sum^{N}_{i = 1} \lambda_i^y = 2.8 \times 10^{-8}$, respectively.
We assume that $u$ measurements are available at all wells, i.e., $N_{\mathbf{u}_\mathrm{s}}=323$. For the RF1 field, we find that Eq. (\ref{eq:pickle-h1reg}) regularization in the PICKLE method provides more accurate results than Eq. (\ref{eq:pickle-l2reg}) regularization. 
For example, for 10 different spatial distributions of 50 observations of $y$, the relative $l_2$ errors in the estimated $y$ field are in the ranges of 0.39-0.88 and 0.41-1.06 for regularizers given by Eqs. (\ref{eq:pickle-h1reg}) and (\ref{eq:pickle-l2reg}), respectively.  
The relative $\ell_2$ errors are  computed on the FV mesh as
\begin{equation*}
    \varepsilon_y = \frac{ \|\hat{\mathbf{y}} - \mathbf{y}_{ref}\|_2 }{ \|\mathbf{y}_{ref}\|_2 }.
\end{equation*}
Therefore, in all cases considered in this section we are using the Eq. (\ref{eq:pickle-h1reg}) regularization.  

\begin{figure}[!htbp]
    \centering
    \begin{tabu} to \textwidth {X[cm]|c*{2}{X[cm]}c}
        reference & $N_{\mathbf{y}_{\mathrm{s}}}$ & $|\text{PICKLE} - \text{reference}|$ & $|\text{MAP} - \text{reference}|$ & \\
        \multirow{4}{*}{\includegraphics[scale=0.33]{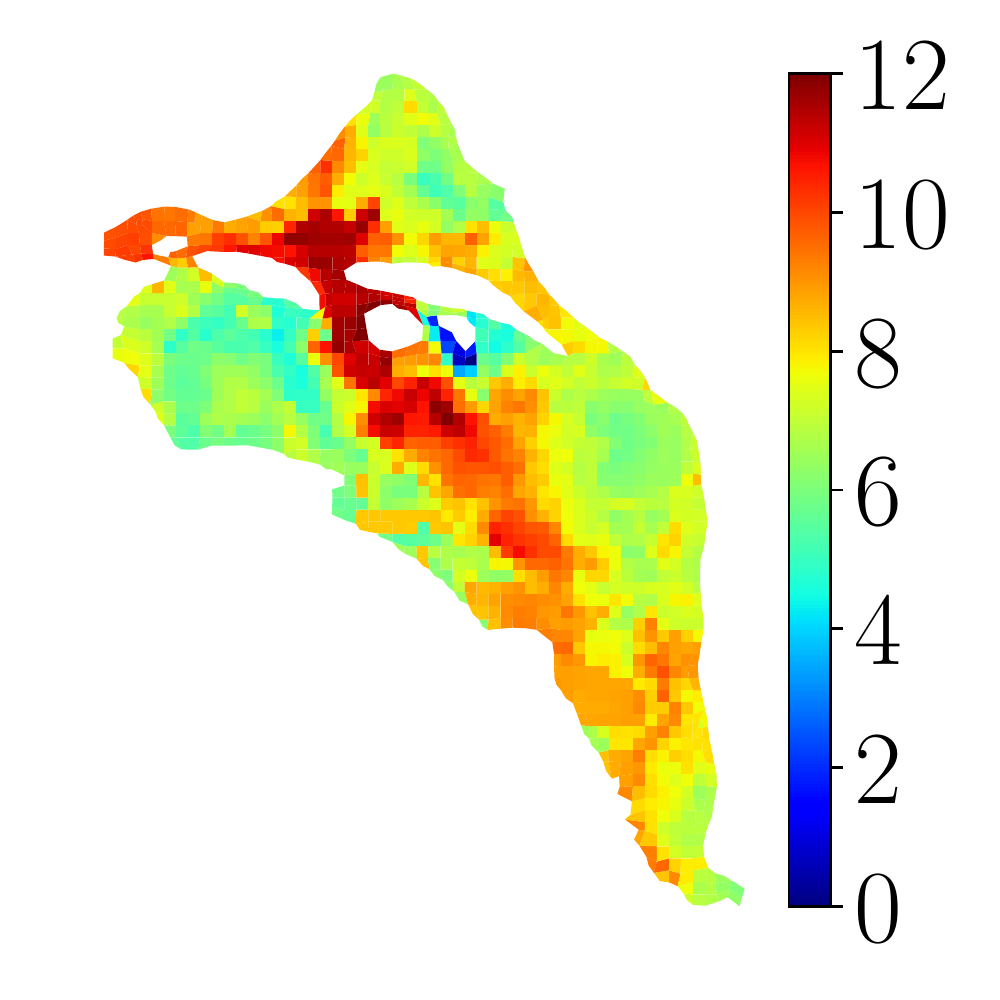}}
        & 200 &
        \includegraphics[scale=0.33]{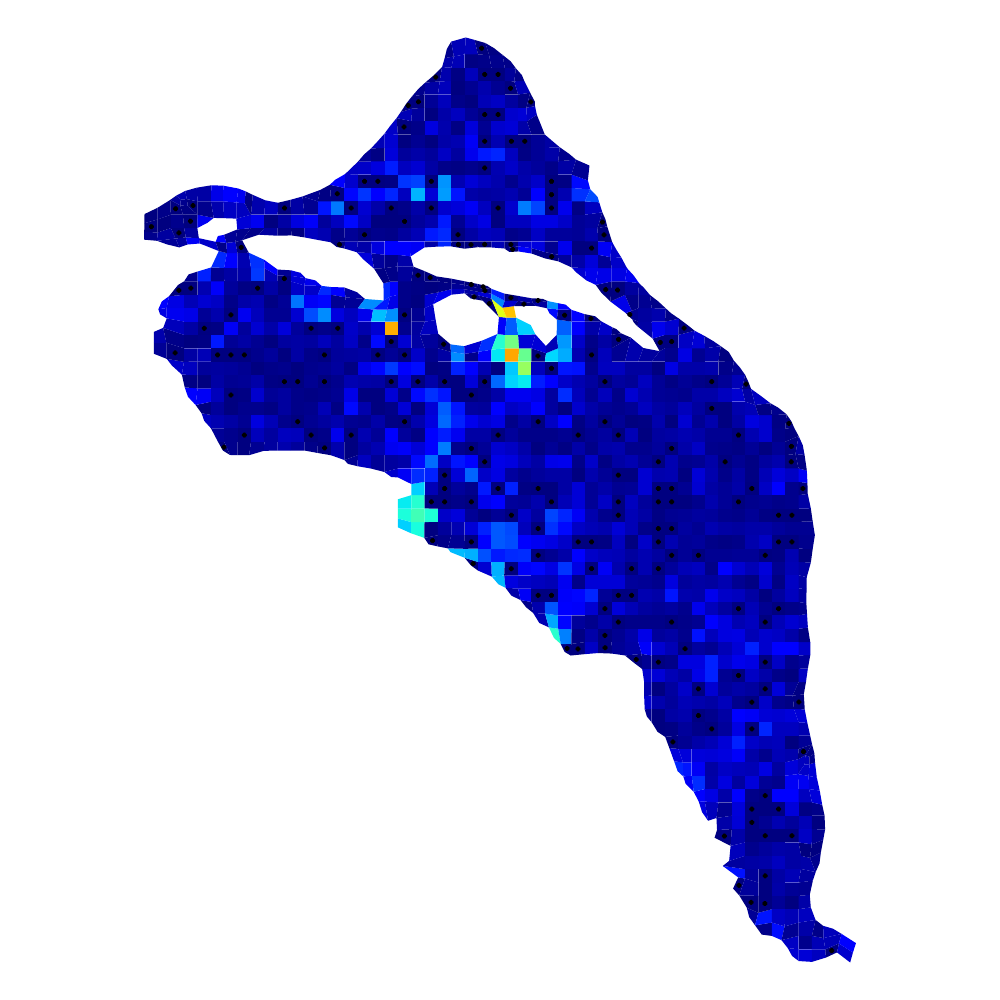} &
        \includegraphics[scale=0.33]{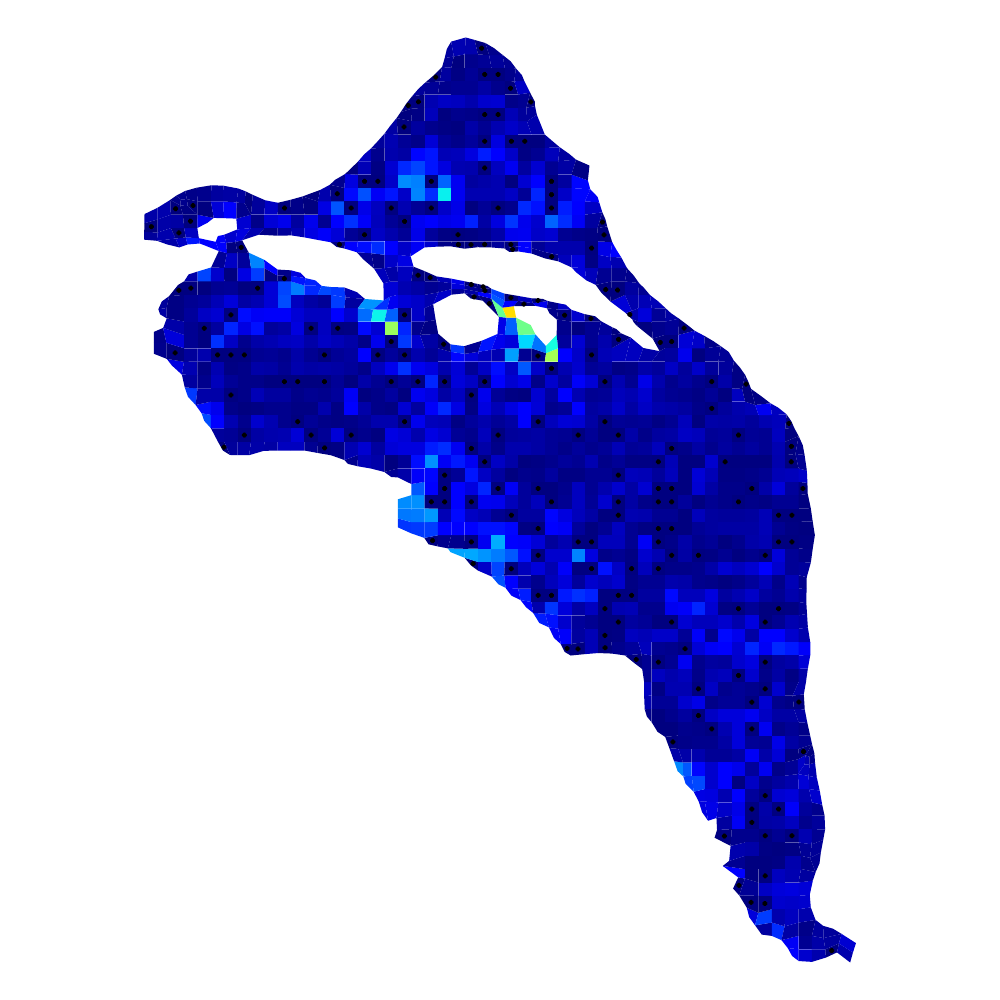} &
        \multirow{4}{*}{\includegraphics[scale=0.44]{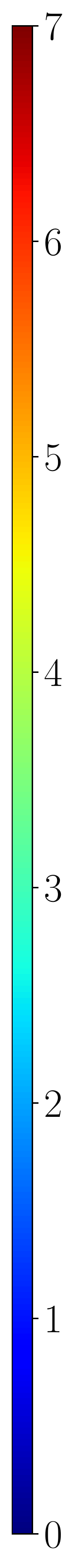}} \\
        & 100 &
        \includegraphics[scale=0.33]{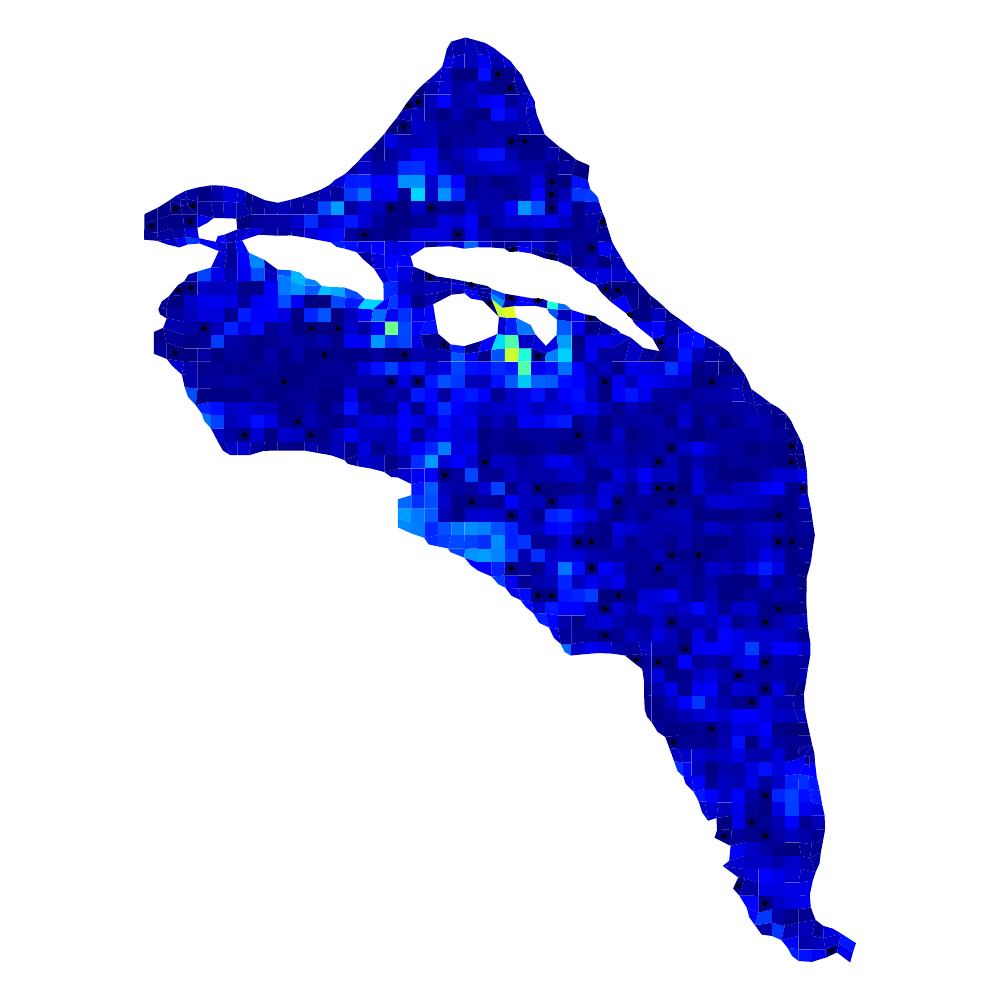} &
        \includegraphics[scale=0.33]{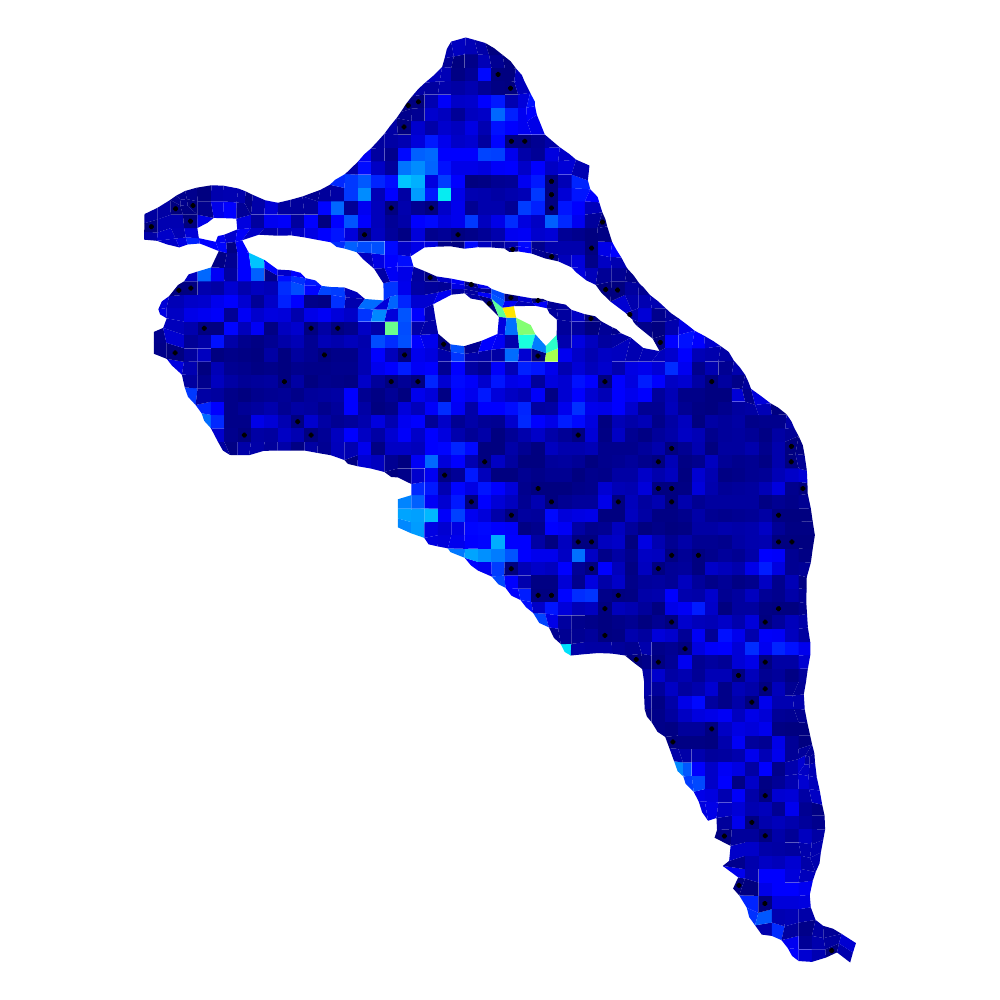} &
        \\
        & 50 &
        \includegraphics[scale=0.33]{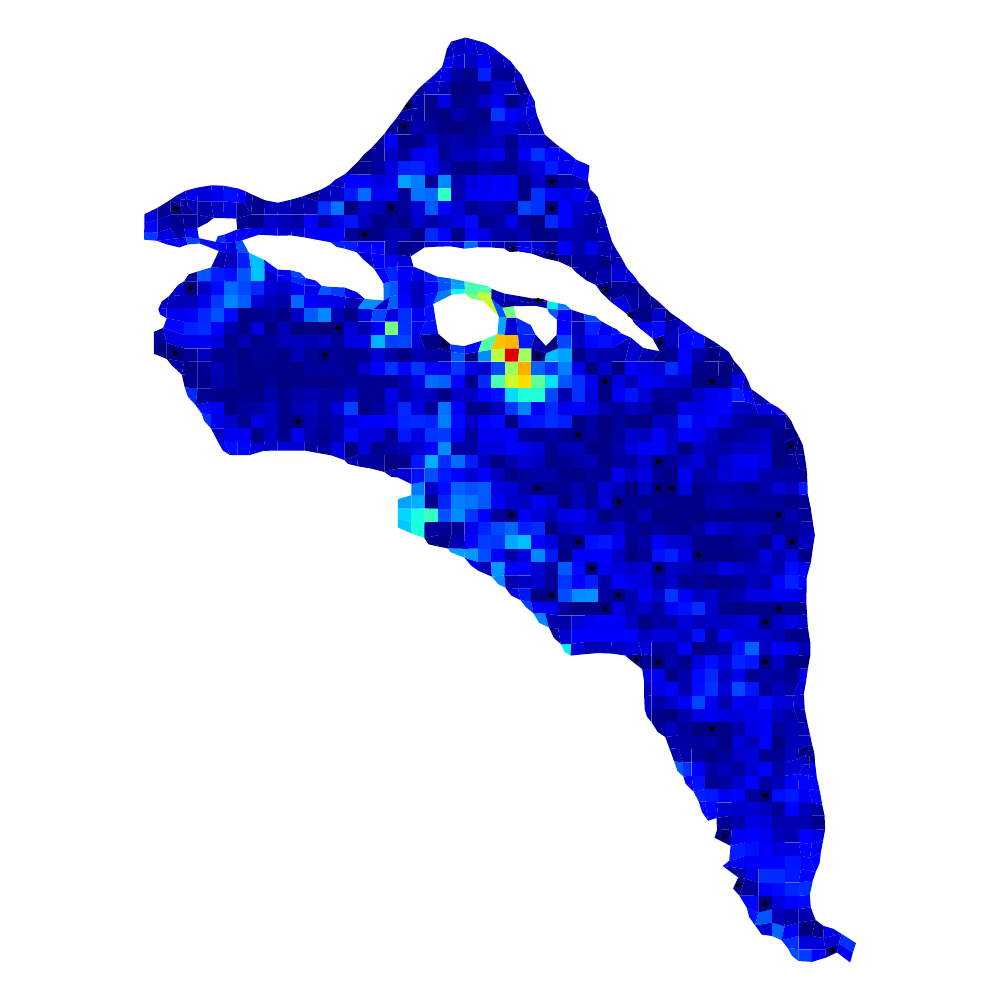} &
        \includegraphics[scale=0.33]{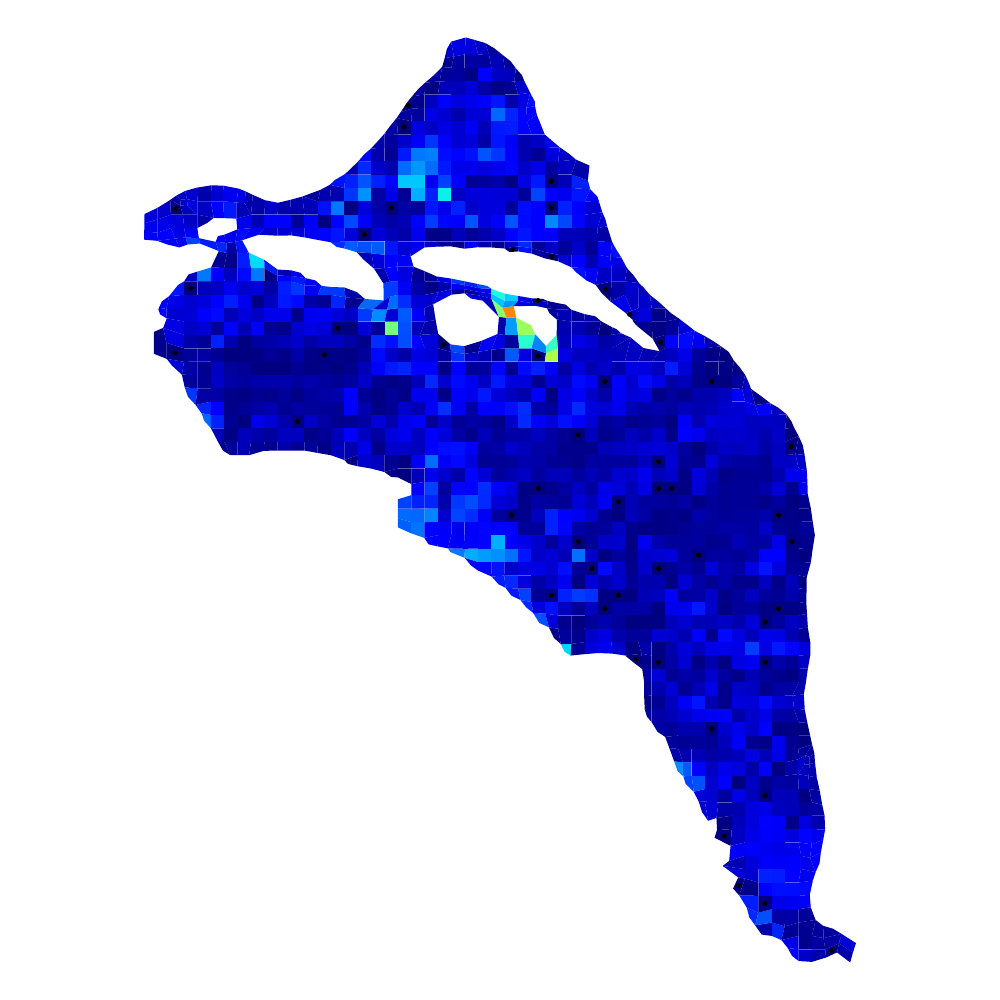} &
        \\
        & 25 &
        \includegraphics[scale=0.33]{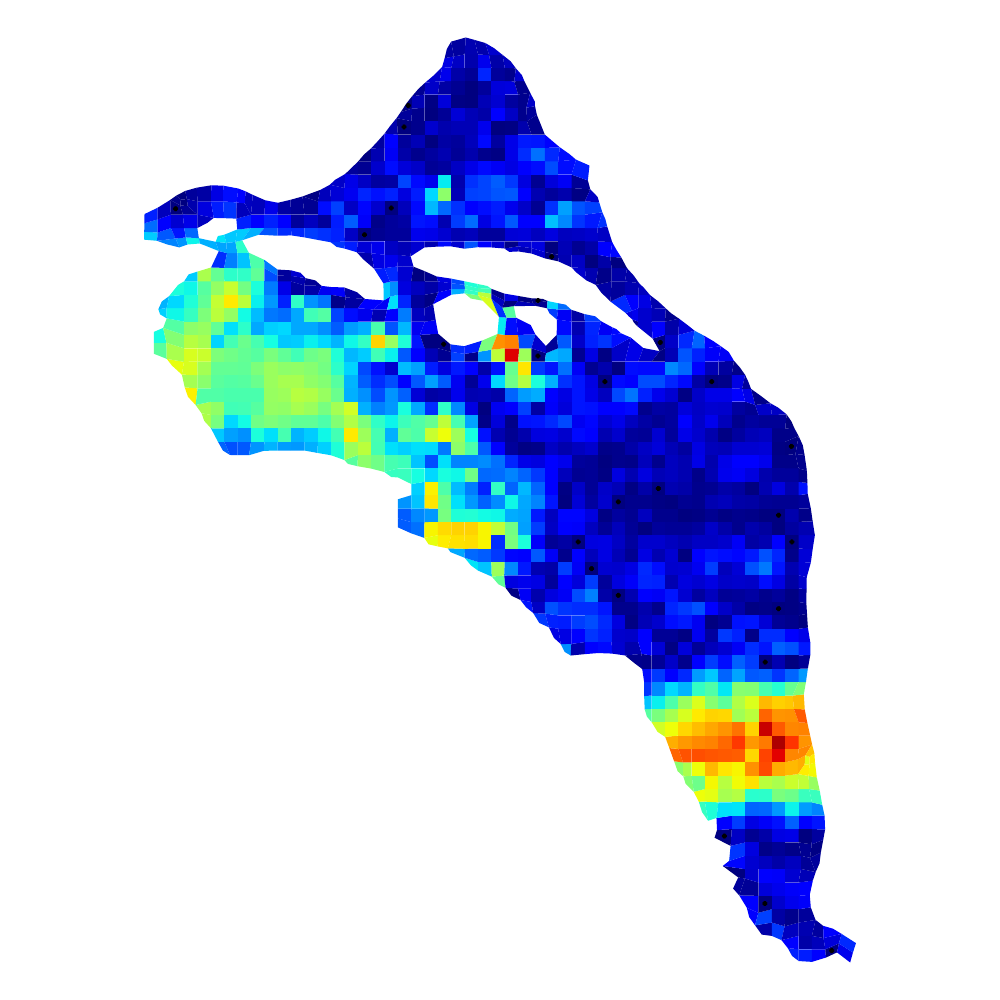} &
        \includegraphics[scale=0.33]{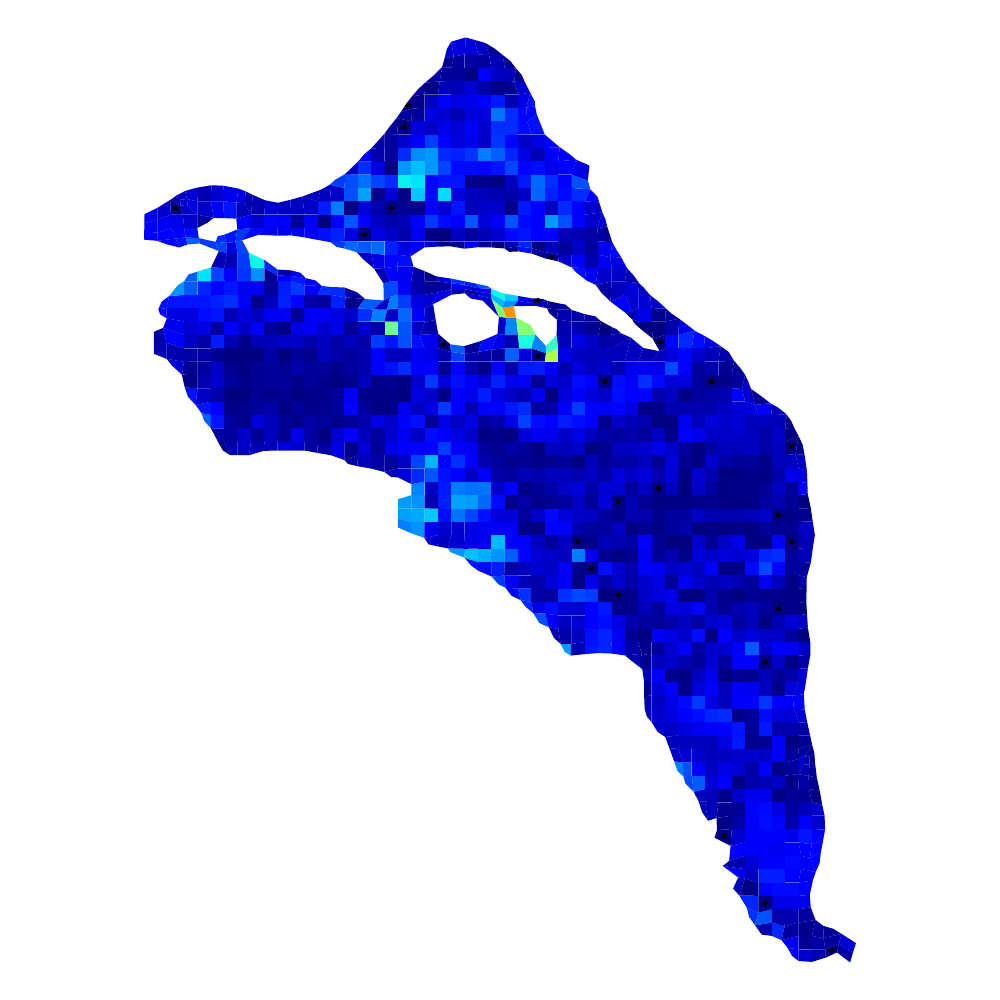} &
        \\
    \end{tabu}
    \caption{The coarse-resolution ($N_{FV}=1475$) RF1 reference $y$ field and the point errors in the PICKLE and MAP estimates of the $y$ field as functions of $N_{\mathbf{y}_{\mathrm{s}}}$  given the unknown Neumann boundary conditions. The dots are the locations of the 100 observations of $y$.}
    \label{fig:hanford_Yref=orig_1x}
\end{figure}

Figure~\ref{fig:hanford_Yref=orig_1x} shows the distribution of point errors in the PICKLE and MAP estimates of $y$ relative to the RF1 $y$ field obtained with $N_{\mathbf{y}_\mathrm{s}}=25$, 50, 100, and 200  $y$ observations. 
%The locations of the measurements are randomly selected from well locations. 
For the considered measurement locations, the PICKLE and MAP methods have comparable accuracy for $N_{\mathbf{y}_\mathrm{s}}\ge 50$, with MAP being more accurate for $N_{\mathbf{y}_\mathrm{s}}=25$.

Because the inverse problem for $y$ is ill-posed, the regularized PICKLE and MAP solutions depend not only on the number of measurements but also on the measurement locations.
To study the effect of the measurement locations on the PICKLE and MAP estimation errors, for each value $N_{\mathbf{y}_\mathrm{s}}$, we randomly generate 10 distributions of $y$ measurement locations and estimate $y$ for each of these locations distributions.
\cref{tab:RF1_1x_unknown_flux_results} shows the ranges of relative $\ell_2 $ and absolute $\ell_\infty$ errors in the PICKLE and MAP $y$ estimates as well as the number of iterations in the minimization algorithm and the execution time (in seconds) for $N_{\mathbf{y}_{\mathrm{s}}}$ 
 ranging from 25 to 400.
For comparison, we also show errors in $y$ estimated with the GPR Eq (\ref{eq:gpr-mean}).
The $\ell_\infty$ error is defined as the maximum of $|\hat{y}(x_i) - y_{ref}(x_i)|$ ($i=1,\ldots,N$), where $\hat{y}(x_i)$ and $y_{ref}(x_i)$ are the values of the estimated and reference $y$ fields at the center of the $i$th FV cell, respectively.     
 
As expected, the accuracy of all methods increases with $N_{\mathbf{y}_{\mathrm{s}}}$.
The PICKLE method is on average slightly less accurate than MAP in terms of both $\ell_2$ and $\ell_\infty$ errors.
However, MAP is more sensitive to the measurements locations.
For example, for $N_{\mathbf{y}_{\mathrm{s}}} = 100$ and 200, we observe that in MAP the maximum $\ell_2$ errors are 2.43 and 1.25, respectively, versus 0.5 and 0.38 in PICKLE.
We attribute the higher robustness of PICKLE relative to MAP with respect to measurement locations to the regularization effect of the CKLE  representation of $y$.
We also note that GPR has significantly larger errors than those in PICKLE and MAP for all considered examples. 
%Also, in all considered cases, PICKLE improves the GRP estimate of $y$.  

\begin{comment}
\begin{figure}
    \centering
    \includegraphics{pickle-hanford-paper/figures/Fig_Ydiff_RF1_1x.pdf}
    \caption{The coarse-resolution ($N_{FV}=1475$) RF1 reference $y$ field and the point errors in the PICKLE and MAP estimates of the $y$ field as functions of $N_{\mathbf{y}_{\mathrm{s}}}$  given the unknown Neumann boundary conditions. The dots are the locations of the 100 observations of $y$.}
    \label{fig:hanford_Yref=orig_1x}
\end{figure}
\end{comment}

Table~\ref{tab:RF1_1x_unknown_flux_results} also shows that the computational cost of PICKLE is significantly smaller than the cost of MAP, and the cost difference increases with increasing $N_{\mathbf{y}_{\mathrm{s}}}$. % 
Note that we give the total execution time of PICKLE that includes the cost of MC evaluation of the mean and covariance of $u$ (approximately 28~s), GPR (approximately 0.4~s), and eigendecomposition (approximately 2.9~s).
The computational cost of GPR for the considered problem is negligible relative to both PICKLE and MAP, and we do not show it in this table. 
As with the estimation errors, we observe that the computational cost of PICKLE is significantly less sensitive to the measurement locations than that of MAP.
For example, the ratio between the PICKLE maximum and minimum execution times in 10 realizations for  $N_{\mathbf{y}_{\mathrm{s}}}=25$ and 400 are 2.74 and 2.64, respectively.
In MAP, for the same values of  $N_{\mathbf{y}_{\mathrm{s}}}$, these ratios are 4.01 and 16.51. The larger variability in the MAP computational time corresponds to the larger variability in the number of iterations in the MAP's least-square minimization algorithm.      

\begin{table}[!htbp]
    \centering
    \caption{Performance of PICKLE and MAP in estimating the coarse-resolution ($N_{FV}=1475$) RF1 as functions of $N_{\mathbf{y}_{\mathrm{s}}}$ with \subref{tab:RF1_1x_unknown_flux_results} unknown and \subref{tab:RF1_1x_known_flux_results} known Neumann boundary conditions.}
    \label{tab:RF1_1x_flux_results}
    \begin{subtable}{\textwidth}
        \caption{Unknown Neumann boundary conditions}
        \label{tab:RF1_1x_unknown_flux_results}%
        \begin{tabu} to \textwidth {*{2}{r}*{5}{X[cm]}}
            \toprule
            & & \multicolumn{5}{c}{$N_{\mathbf{y}_{\mathrm{s}}}$} \\
            & solver & 400 & 200 & 100 & 50 & 25 \\
            \midrule
            \multirow{2}{*}{\shortstack[r]{least square\\iterations}} & PICKLE & 12--35 & 13--23 & 14--30 & 17--64 & 18--51 \\
            & MAP & 25--400 & 27--603 & 22--535 & 33--483 & 93--357 \\
            \midrule
            \multirow{2}{*}{\shortstack[r]{execution\\time (s)}} & PICKLE & 143.26--378.35 & 144.73--253.47 & 142.71--300.60 & 159.94--592.47 & 180.43--495.34 \\
            & MAP & 116.74--1928.08 & 121.87--2737.72 & 70.40--2283.90 & 126.52--2134.91 & 402.08--1613.52 \\
            \midrule
            \multirow{3}{*}{\shortstack[r]{relative\\$\ell_2$ error}} & GPR & 0.29--0.35 & 0.40--0.48 & 0.50--0.66 & 0.62--0.76 & 0.74--1.03 \\
            & PICKLE & 0.24--0.29 & 0.32--0.38 & 0.35--0.50 & 0.42--0.64 & 0.55--1.20 \\
            & MAP & 0.22--0.26 & 0.28--1.25 & 0.32--2.43 & 0.36--0.42 & 0.40--0.55 \\
            \midrule
            \multirow{3}{*}{\shortstack[r]{absolute\\$\ell_\infty$ error}} & GPR & 3.74--5.87 & 3.79--7.39 & 4.24--8.00 & 5.64--8.27 & 6.21--9.71 \\
            & PICKLE & 3.68--5.16 & 4.17--6.57 & 4.36--6.10 & 4.52--6.40 & 4.61--7.57 \\
            & MAP & 3.68--6.30 & 3.72--39.49 & 4.06--79.01 & 4.07--6.51 & 3.91--6.29 \\
            \bottomrule
        \end{tabu}
    \end{subtable}\\
    \vspace{1em}%
    \begin{subtable}{\textwidth}
        \caption{Known Neumann boundary conditions}
        \label{tab:RF1_1x_known_flux_results}%
        \begin{tabu} to \textwidth {*{2}{r}*{5}{X[cm]}}
            \toprule
            & & \multicolumn{5}{c}{$N_{\mathbf{y}_{\mathrm{s}}}$} \\
            & solver & 400 & 200 & 100 & 50 & 25 \\
            \midrule
            \multirow{2}{*}{\shortstack[r]{least square\\iterations}} & PICKLE & 28--54 & 29--55 & 37--85 & 31--78 & 47--158 \\
            & MAP & 1081--2626 & 1488--4143 & 3497--19048 & 5343--13247 & 7662--15694 \\
            \midrule
            \multirow{2}{*}{\shortstack[r]{execution\\time (s)}} & PICKLE & 119.43--216.47 & 119.43--216.47 & 137.37--324.25 & 120.67--305.00 & 155.06--510.83 \\
            & MAP & 232.30--545.90 & 320.33--836.32 & 726.96--3649.97 & 1072.23--2582.03 & 1544.52--3133.03 \\
            \midrule
            \multirow{2}{*}{\shortstack[r]{relative\\$\ell_2$ error}} & PICKLE & 0.24--0.27 & 0.30--0.35 & 0.34--0.46 & 0.39--0.88 & 0.44--1.47 \\
            & MAP & 0.21--0.29 & 0.27--0.30 & 0.31--0.36 & 0.35--0.42 & 0.39--0.45 \\
            \midrule
            \multirow{2}{*}{\shortstack[r]{absolute\\$\ell_\infty$ error}} & PICKLE & 3.48--4.99 & 3.62--5.43 & 4.32--5.45 & 4.87--8.76 & 4.55--9.51 \\
            & MAP & 3.15--5.55 & 3.65--6.45 & 3.64--6.40 & 4.91--6.39 & 5.37--6.57 \\
            \bottomrule
        \end{tabu}
    \end{subtable}%
\end{table}

\begin{comment}
\begin{table}[!htbp]
    \centering
    \caption{Performance of PICKLE and MAP for the RF1 field with $N_{FV}=1475$ and known Neumann boundary conditions. 1000 KL terms are used in the CKLEs of $y$ and $u$.}
    \label{tab:low_res_known_flux_results}
    \begin{tabu} to \textwidth {*{2}{r}*{6}{X[cm]}}
        \toprule
        & & \multicolumn{6}{c}{$N_{\mathbf{y}_{\mathrm{s}}}$} \\
        & solver & 600 & 400 & 200 & 100 & 50 & 25 \\
        \midrule
        \multirow{2}{*}{\shortstack[r]{least square\\iterations}} & PICKLE & 14 & 23 & 14 & 18 & 23 & 28 \\
        & MAP & 105 & 41 & 26 & 29 & 44 & 73 \\
        \midrule
        \multirow{2}{*}{\shortstack[r]{execution\\time (s)}} & PICKLE & 175.97 & 272.93 & 161.59 & 194.42 & 239.63 & 272.54 \\
        & MAP & 520.61 & 167.58 & 121.25 & 121.01 & 184.31 & 315.83 \\
        \midrule
        \multirow{2}{*}{\shortstack[r]{relative\\$\ell_2$ error}} & PICKLE & 0.23 & 0.28 & 0.33 & 0.42 & 0.49 & 0.99 \\
        & MAP & 0.19 & 0.23 & 0.29 & 0.33 & 0.42 & 0.43 \\
        \midrule
        \multirow{2}{*}{\shortstack[r]{absolute\\$\ell_\infty$ error}} &  PICKLE & 3.35 & 4.45 & 4.55 & 4.46 & 5.32 & 6.84 \\
        & MAP & 3.99 & 4.30 & 4.22 & 4.32 & 6.55 & 6.43 \\
        \bottomrule
    \end{tabu}
\end{table}
\end{comment}

Next, we investigate the performance of the PICKLE and MAP methods as functions of $N_{\mathbf{y}_{\mathrm{s}}}$ when the Neumann boundary conditions are known.
We note that the GPR method for estimating $y$ is based solely on $y$ measurements and, therefore, is independent of the boundary conditions. Therefore, we do not present GPR errors in this comparison study.  
Table~\ref{tab:RF1_1x_known_flux_results} shows the errors and execution time in the PICKLE and MAP methods for the same sets of $y$ measurements as in the unknown Neumann boundary condition cases.
We find that the errors of both methods only slightly decrease (less than 5\%) relative to the unknown Neumann boundary condition cases. 
The execution time of PICKLE is practically not affected by whether the Neumann boundary conditions are known deterministically or stochastically, while the MAP execution time is increased.

Theoretically, increasing the number of KL terms in the CKLE of $y$ should increase the accuracy of PICKLE because it allows capturing more accurately the spatial correlation structure of $y(x)$. However, increasing the number of KL terms also increases the number of unknown parameters and, therefore, the computational cost of PICKLE.
In~\cref{tab:RF1_1x_diff_kl_unknown_flux_results,tab:RF1_1x_diff_kl_known_flux_results},
we compare the errors and execution time of PICKLE with 1000 and 1400 terms in the $y$ CKLE for the cases with unknown and known boundary conditions, respectively.
We observe that increasing the number of KL terms does not lead to a significant increase in the accuracy of PICKLE.
The $\ell_2$ error decreases slightly, with only significant (10\%) improvement for the smallest considered number ($N_{\mathbf{y}_{\mathrm{s}}} = 50$) of $y$ measurements.
This is because $N_y = 1000$ already corresponds to a very small value of $\text{rtol}_y = 2.8 \times 10^{-8}$.
The further increase in $N_y$ does not significantly improve the approximation power of the CKLE, but makes the minimization problem computationally more difficult and costly. We observe a slight increase in the $\ell_\infty$ errors due to the fact that the larger number of KL terms might require a stronger regularization (i.e., larger values of $\beta$). On the other hand, the increase in $N_y$ leads to a significant increase in the execution time of PICKLE, by approximately a factor of 4 for $N_{\mathbf{y}_{\mathrm{s}}} = 50$ and a factor of 2 for $N_{\mathbf{y}_{\mathrm{s}}} = 100$, 200, and 323. 

\begin{table}[!htbp]
    \centering
    \caption{Performance of PICKLE as a function of the number of KL terms for estimating the coarse-resolution RF1 field with \subref{tab:RF1_1x_diff_kl_unknown_flux_results} unknown and \subref{tab:RF1_1x_diff_kl_known_flux_results} known Neumann boundary conditions.}
    \label{tab:RF1_1x_diff_kl_results}
    \begin{subtable}{\textwidth}
        \caption{Unknown Neumann boundary conditions}
        \label{tab:RF1_1x_diff_kl_unknown_flux_results}
        \begin{tabu} to \textwidth {*{2}{r}*{4}{X[cm]}}
            \toprule
            & & \multicolumn{4}{c}{$N_{\mathbf{y}_{\mathrm{s}}}$} \\
            & KL terms & 323 & 200 & 100 & 50 \\
            \midrule
            \multirow{2}{*}{least square iterations} & 1000 & 9 & 13 & 19 & 36 \\
            & 1400 & 9 & 10 & 15 & 67 \\
            \midrule
            \multirow{2}{*}{execution time (s)} & 1000 & 81.59 & 115.46 & 166.63 & 317.18 \\
            & 1400 & 179.37& 202.58 & 306.18 & 1374.94 \\
            \midrule
            \multirow{2}{*}{relative $\ell_2$ error} & 1000 & 0.32 & 0.39 & 0.44 & 0.58 \\
            & 1400 & 0.32 & 0.37 & 0.42 & 0.48 \\
            \midrule
            \multirow{2}{*}{absolute $\ell_\infty$ error} & 1000 & 5.20 & 5.36 & 5.07 & 6.49 \\
            & 1400 & 5.32 & 5.50 & 5.49 & 6.57 \\
            \bottomrule
        \end{tabu}
    \end{subtable}\\
    \vspace{1em}%
    \begin{subtable}{\textwidth}
        \caption{Known Neumann boundary conditions}
        \label{tab:RF1_1x_diff_kl_known_flux_results}
        \centering
        \begin{tabu} to \textwidth {*{2}{r}*{4}{X[cm]}}
            \toprule
            & & \multicolumn{4}{c}{$N_{\mathbf{y}_{\mathrm{s}}}$} \\
            & KL terms & 323 & 200 & 100 & 50 \\
            \midrule
            \multirow{2}{*}{least square iterations} & 1000 & 9 & 13 & 18 & 23 \\
            & 1400 & 9 & 11 & 12 & 42 \\
            \midrule
            \multirow{2}{*}{execution time (s)} & 1000 & 80.88 & 116.97 & 169.18 & 199.28 \\
            & 1400 & 179.82 & 226.61 & 238.54 & 887.05 \\
            \midrule
            \multirow{2}{*}{relative $\ell_2$ error} & 1000 & 0.31 & 0.38 & 0.43 & 0.56 \\
            & 1400 & 0.31 & 0.37 & 0.43 & 0.49 \\
            \midrule
            \multirow{2}{*}{absolute $\ell_\infty$ error} & 1000 & 5.37 & 5.38 & 5.14 & 6.43 \\
            & 1400 & 5.39 & 5.42 & 5.51 & 6.51 \\
            \bottomrule
        \end{tabu}
    \end{subtable}
\end{table}

\begin{table}[!htbp]
  \caption{Performance of PICKLE and MAP for estimating the coarse-resolution ($N_{FV}=1475$) RF2 as functions of $N_{\mathbf{y}_{\mathrm{s}}}$ with \subref{tab:RF2_1x_unknown_flux_results} unknown and \subref{tab:RF2_1x_known_flux_results} known Neumann boundary conditions.
  }
    \label{tab:RF2_1x_flux_results}
    \begin{subtable}{\textwidth}
        \caption{Unknown Neumann boundary conditions}%
        \label{tab:RF2_1x_unknown_flux_results}%
        \begin{tabu} to \textwidth {*{2}{r}c*{3}{X[cm]}}
            \toprule
            & & \multicolumn{4}{c}{$N_{\mathbf{y}_{\mathrm{s}}}$} \\
            & solvers & 100 & 50 & 25 & 10\\
            \midrule
            \multirow{2}{*}{\shortstack[r]{least square\\ iterations}} & PICKLE & 9--11 & 11--14 & 11--18 & 15--47\\
            & MAP & 14--144 & 14--62 & 14--87 & 16--346 \\
            \midrule
            \multirow{2}{*}{\shortstack[r]{execution\\ time (s)}} & PICKLE & 83.4--114 & 81.5--105 & 85.3--138 & 56.7--136 \\
            & MAP & 38.1--306 & 29.5--138 & 28.4--191 & 19.2--310 \\
            \midrule
            \multirow{3}{*}{\shortstack[r]{relative\\ $\ell_2$ error}} & GPR & 0.118--0.175 & 0.207--0.398 & 0.353--0.513 & 0.468--0.875 \\
            & PICKLE & 0.0300-0.0450 & 0.0539--0.145 & 0.0876--0.232 & 0.174--0.816 \\
            & MAP & 0.0892--0.107 & 0.109--0.163 & 0.150--0.227 & 0.198-0.300 \\
            \midrule
            \multirow{3}{*}{\shortstack[r]{absolute\\ $\ell_\infty$ error}} & GPR & 1.04--2.37 & 2.11--4.36 & 2.05--4.29 & 3.34--4.45 \\
            & PICKLE & 0.781--0.854 & 0.792--1.20 & 0.831--1.51 & 1.18--3.61 \\
            & MAP & 0.790--1.09 & 0.858--1.49 & 1.08--1.44 & 1.09--1.49 \\
            \bottomrule
        \end{tabu}
    \end{subtable}\\
    \vspace{1em}%
    \begin{subtable}{\textwidth}
        \caption{Known Neumann boundary conditions}%
        \label{tab:RF2_1x_known_flux_results}%
        \begin{tabu} to \textwidth {*{2}{r}c*{3}{X[cm]}}
            \toprule
            & & \multicolumn{4}{c}{$N_{\mathbf{y}_{\mathrm{s}}}$} \\
            & solvers & 100 & 50 & 25 & 10\\
            \midrule
            \multirow{2}{*}{\shortstack[r]{least square\\ iterations}} & PICKLE & 9--11 & 11--14 & 11--17 & 14--30 \\
            & MAP & 14--47 & 14--53 & 17--193 & 61--109 \\
            \midrule
            \multirow{2}{*}{\shortstack[r]{execution\\ time (s)}} & PICKLE & 72.8--94 & 53.5--68.2 & 53.3--71.8 & 58.2--107 \\
            & MAP & 39.5--326 & 66.4--121 & 72.1--122 & 66.6--126 \\
            \midrule
            \multirow{2}{*}{\shortstack[r]{relative\\ $\ell_2$ error}} & PICKLE & 0.0291--0.0411 & 0.0423--0.145 & 0.0773--0.245 & 0.129--0.547 \\
            & MAP & 0.0857--0.102 & 0.103--0.156 & 0.155--0.251 & 0.174--0.280 \\
            \midrule
            \multirow{2}{*}{\shortstack[r]{absolute\\ $\ell_\infty$ error}} & PICKLE & 0.734--0.985 & 0.794--1.28 & 0.834--1.46 & 1.06--2.45 \\
            & MAP & 0.800--1.08 & 0.873--1.47 & 1.02--1.45 & 1.02--1.49 \\
            \bottomrule
        \end{tabu}
    \end{subtable}
\end{table}

\subsection{RF2 reference field}\label{sec:RF2_field}
Here, we estimate $y$ and $u$ using the synthetic measurements of $y$ and $u$ generated on the coarse and fine meshes for the RF2 reference field. 
%The reference $u$ fields are computed as the FV solutions of Eqs. (\ref{eq:pde})-(\ref{eq:pde-head-bc}) with $K(x)=\ln y_{RF2}$ on the coarse and fine mesh, respectively. 
We assume that $u$ measurements are available at all wells, i.e.,  $N_{\mathbf{u}_s}=323$ and $408$ on the coarse and fine meshes, respectively. 
As in Section \ref{sec:RF1_field}, the number of KL terms in the $y$ and $u$ expansions is set to $N_y= N_u=1000$. 
 The corresponding relative tolerances for these choices of $N_y$ and $N_u$ are 
$\text{rtol}_u = 3.01 \times 10^{-9}$
and
$\text{rtol}_y = 7.9 \times 10^{-6}$, respectively.
Opposite to our results for the RF1 field, here we find that the Eq. (\ref{eq:pickle-l2reg}) regularization in the PICKLE method provides more accurate results than the Eq. (\ref{eq:pickle-h1reg})  regularization. 
For 10 different spatial distributions of 50 observations of $y$, the relative $l_2$ errors in the estimated $y$ field are in the ranges of 0.21--0.40 and 0.0423--0.145 for regularizers given by Eqs. (\ref{eq:pickle-h1reg}) and (\ref{eq:pickle-l2reg}), respectively. Therefore, in this section we are using the Eq. (\ref{eq:pickle-l2reg}) regularization.

\begin{figure}[!htbp]
    \centering
    \begin{tabu} to \textwidth {X[cm]|c*{2}{X[cm]}c}
        reference & $N_{\mathbf{y}_{\mathrm{s}}}$ & $|\text{PICKLE} - \text{reference}|$ & $|\text{MAP} - \text{reference}|$ & \\
        \multirow{4}{*}{\includegraphics[scale=0.33]{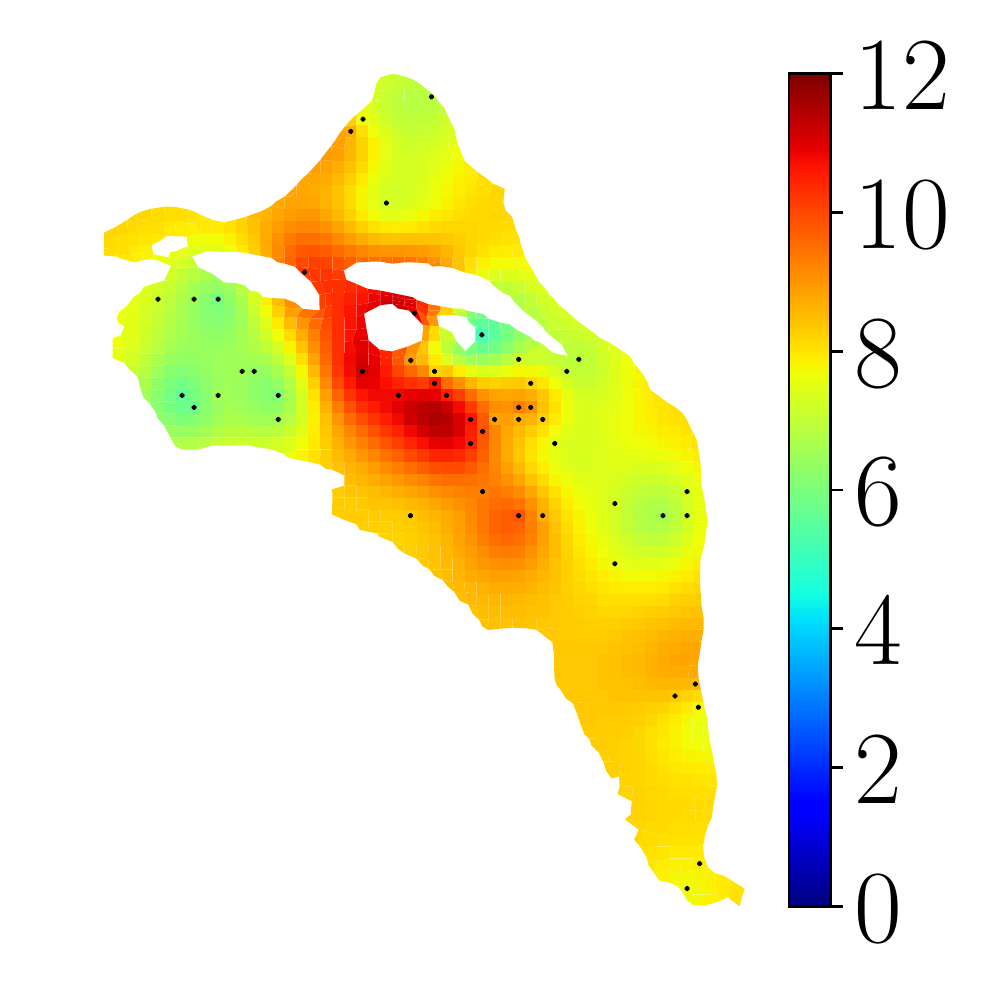}} & 100 &
        \includegraphics[scale=0.33]{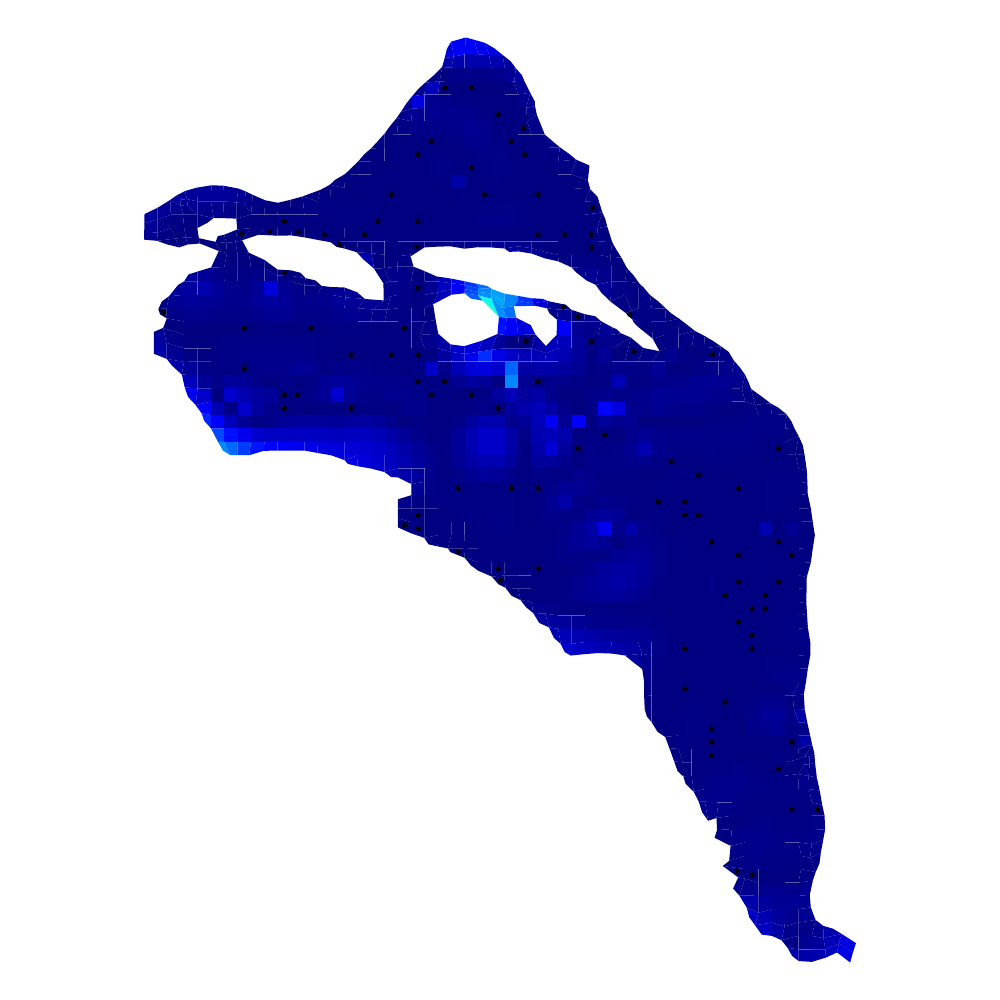} &
        \includegraphics[scale=0.33]{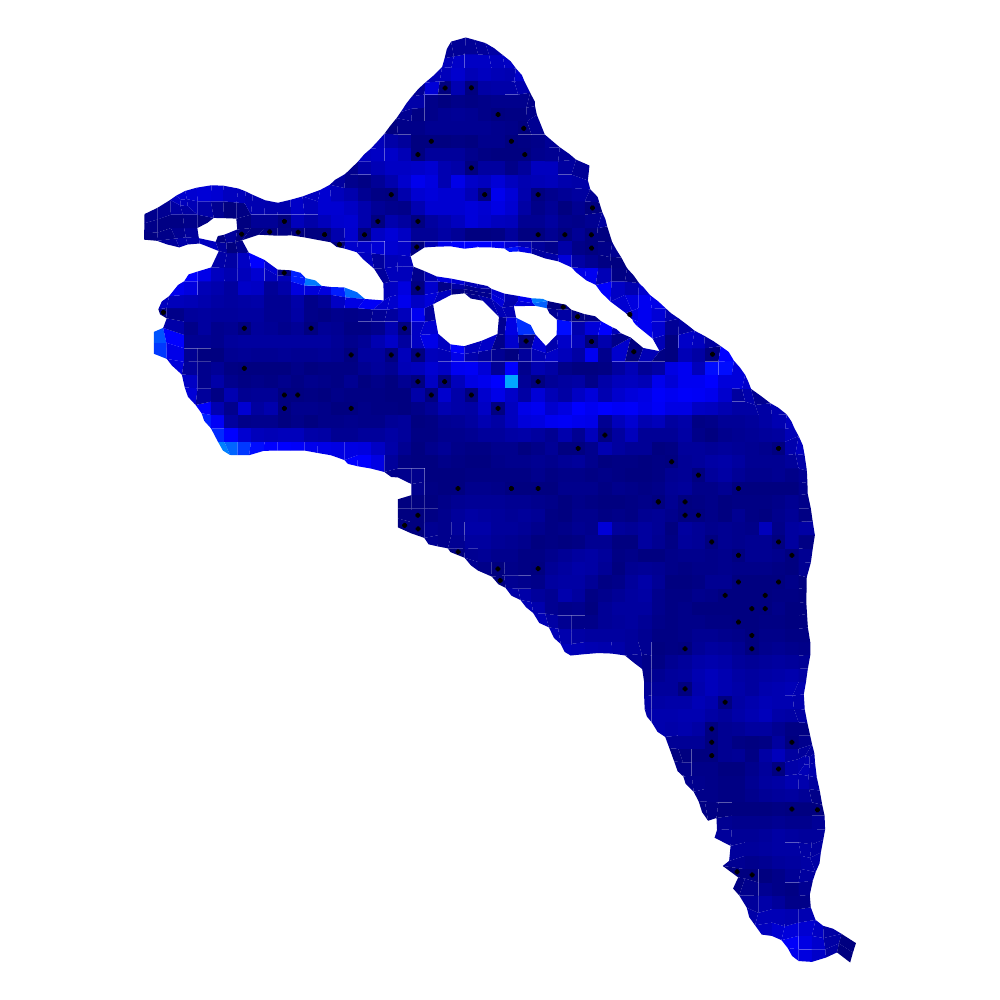} &
        \multirow{4}{*}{\includegraphics[scale=0.44]{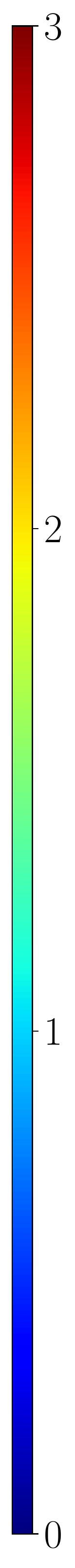}} \\
        & 50 &
        \includegraphics[scale=0.33]{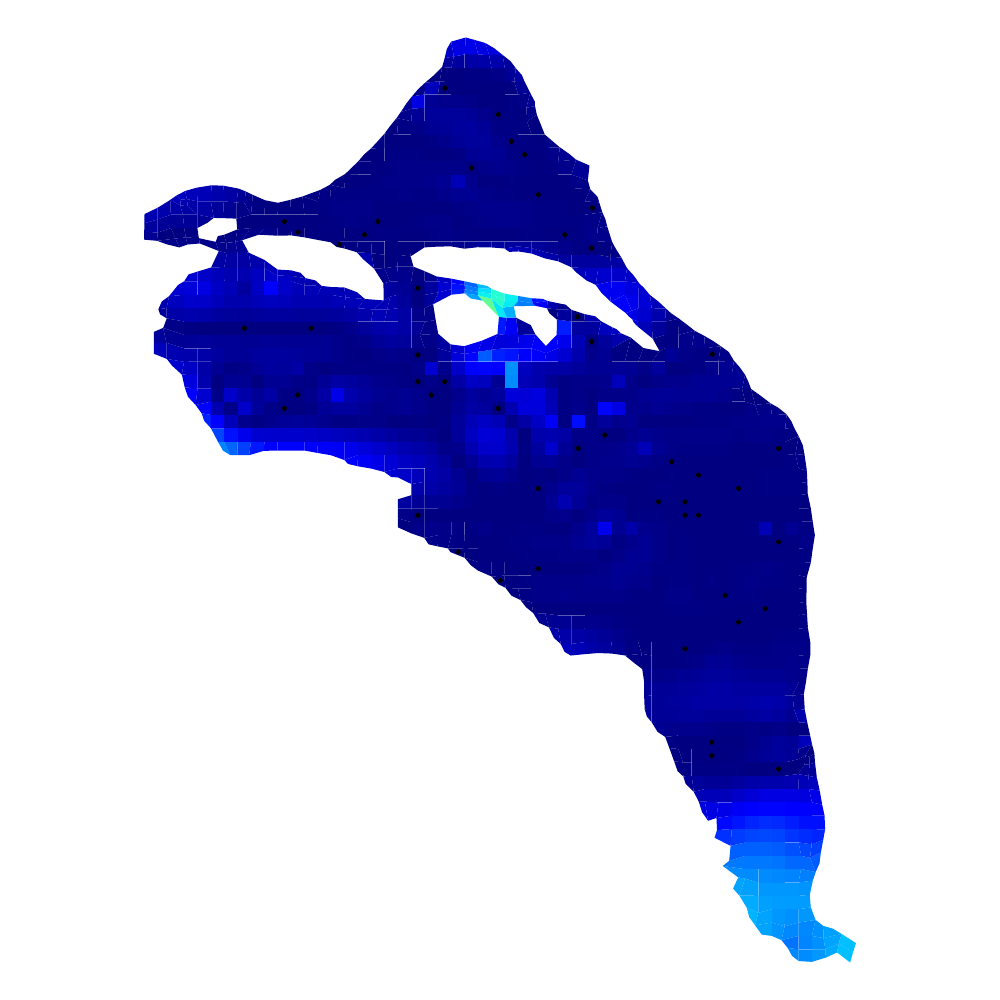} &
        \includegraphics[scale=0.33]{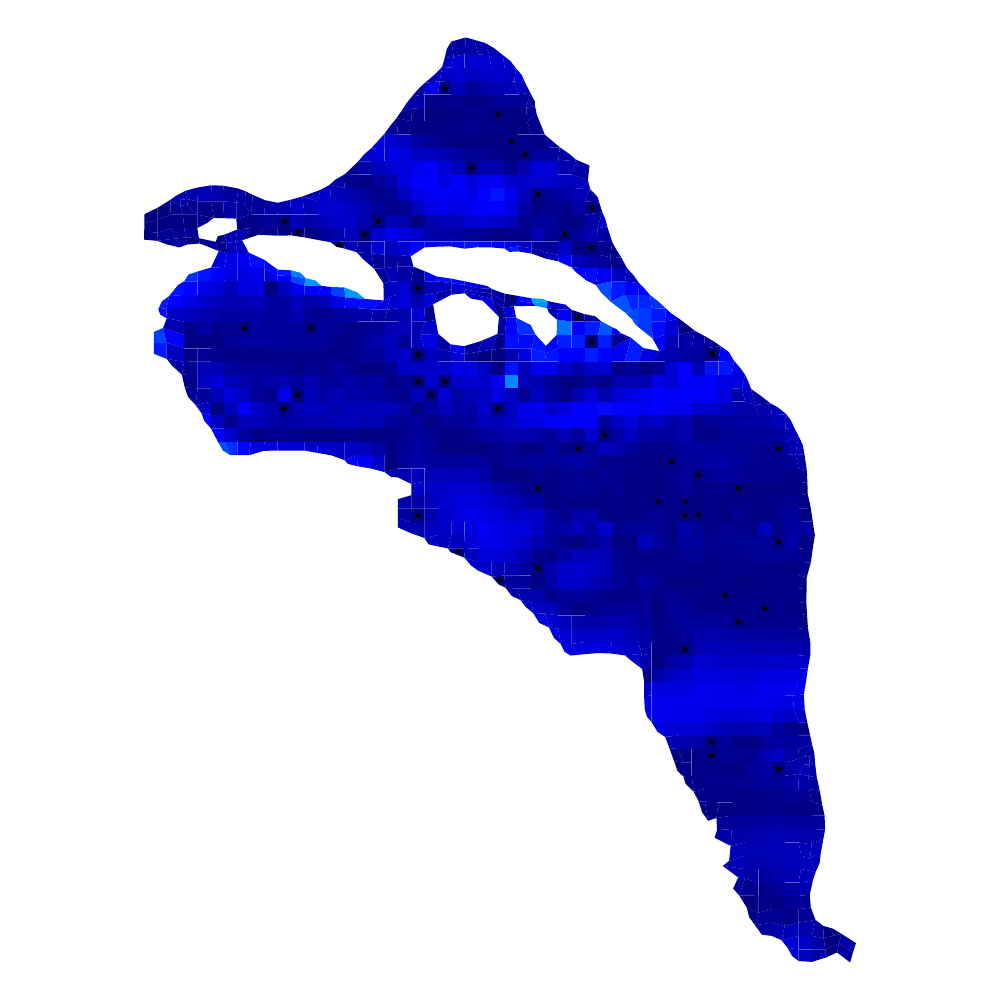} &
        \\
        & 25 &
        \includegraphics[scale=0.33]{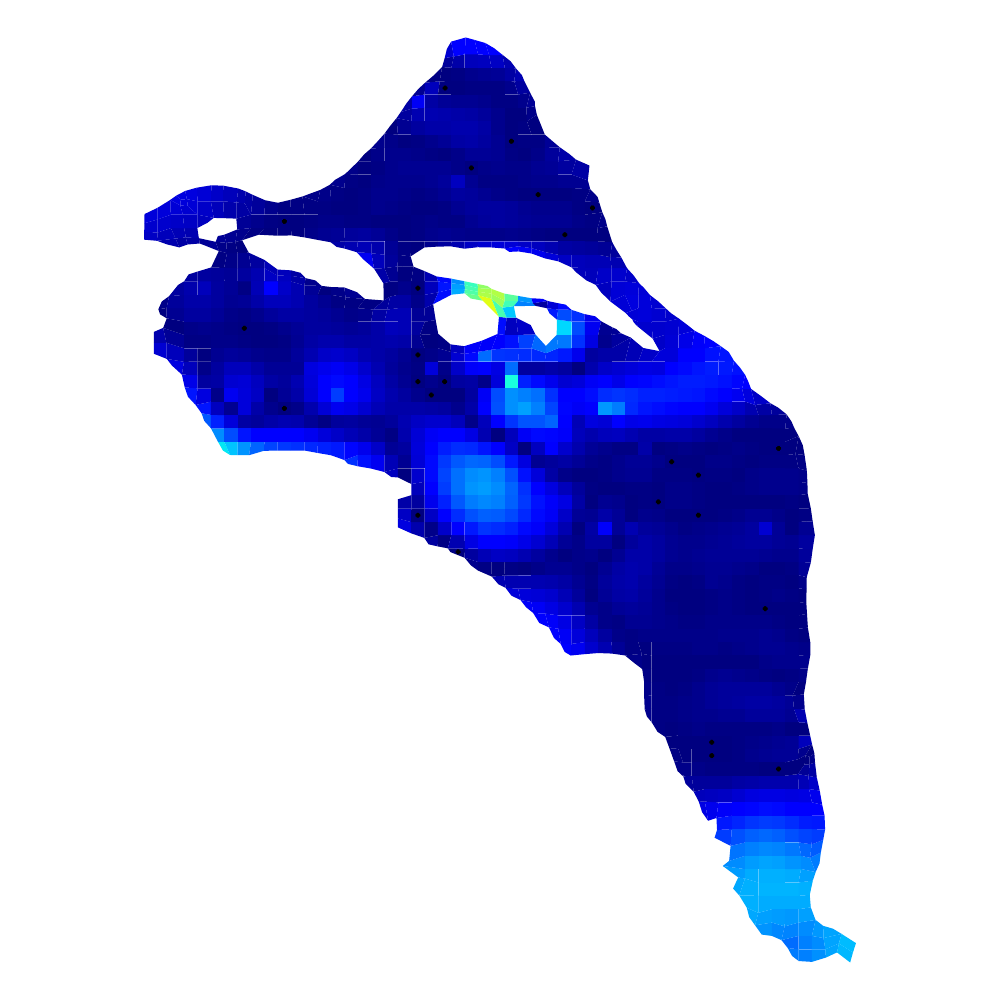} &
        \includegraphics[scale=0.33]{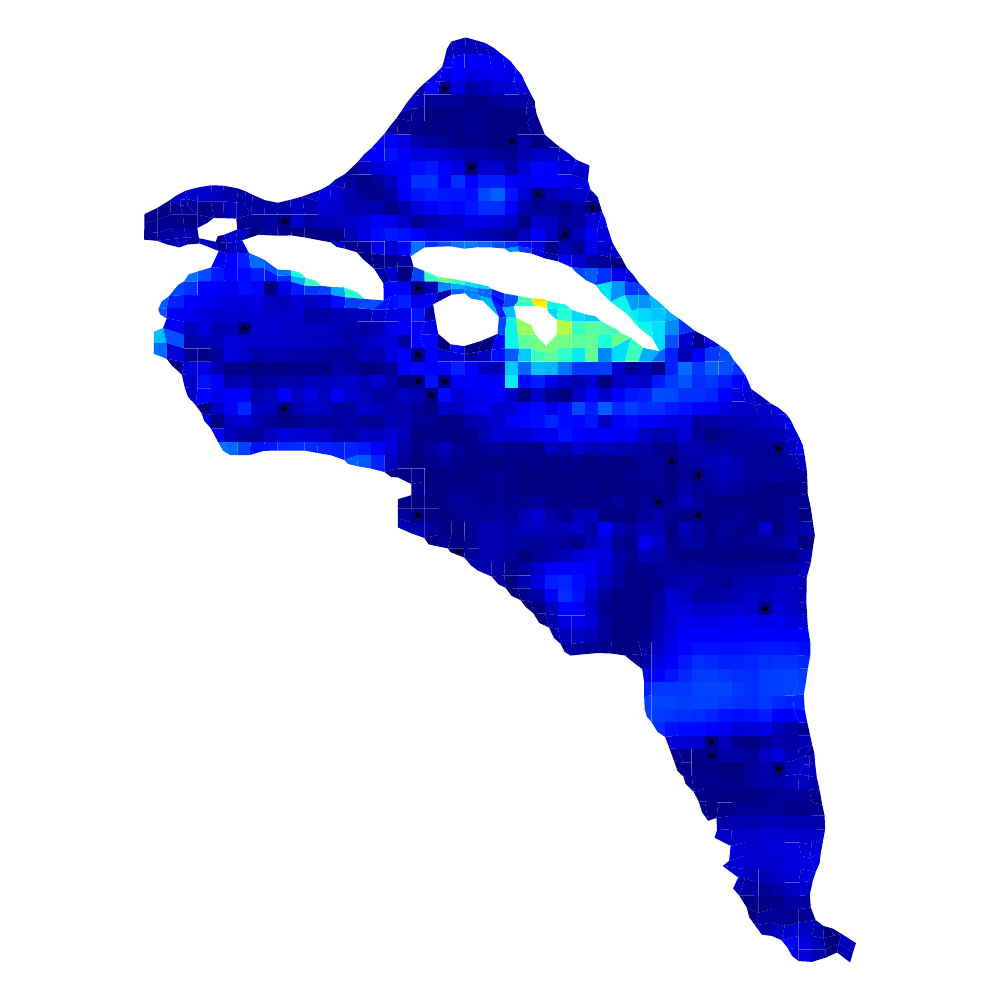} &
        \\
        & 10 &
        \includegraphics[scale=0.33]{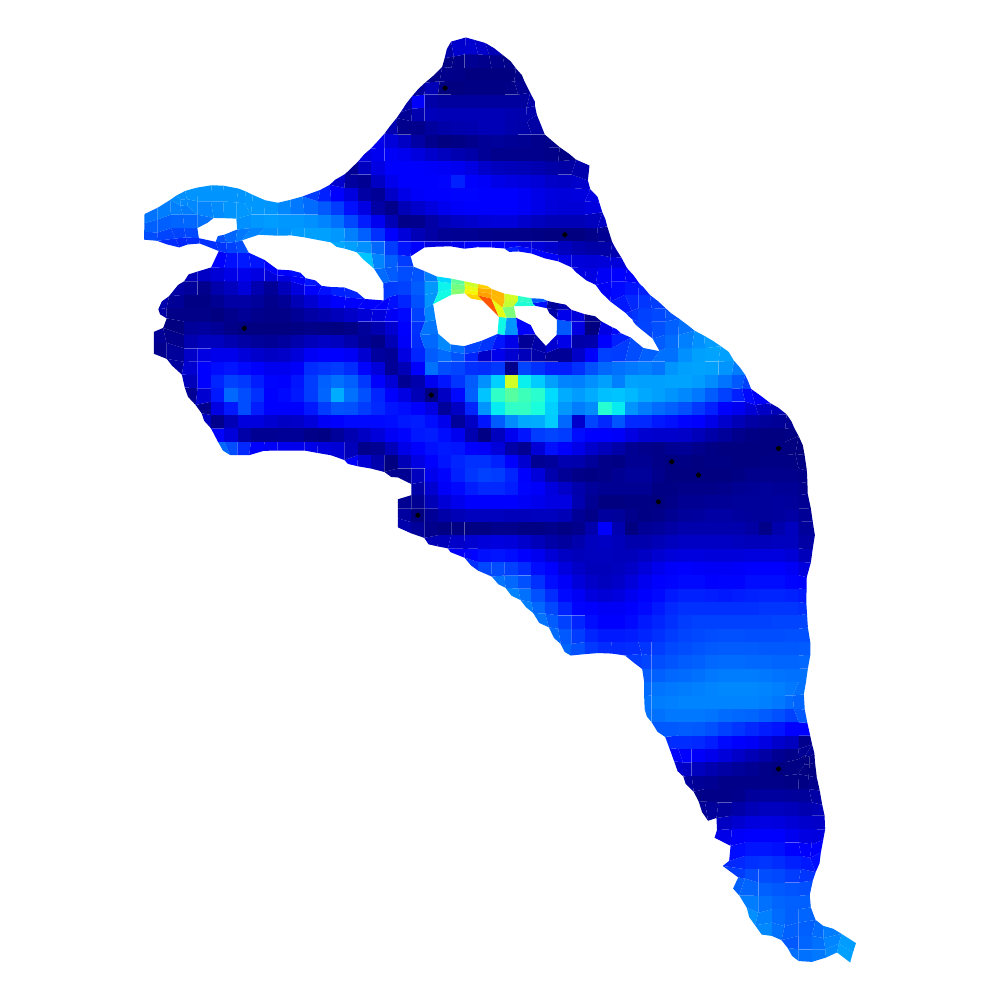} &
        \includegraphics[scale=0.33]{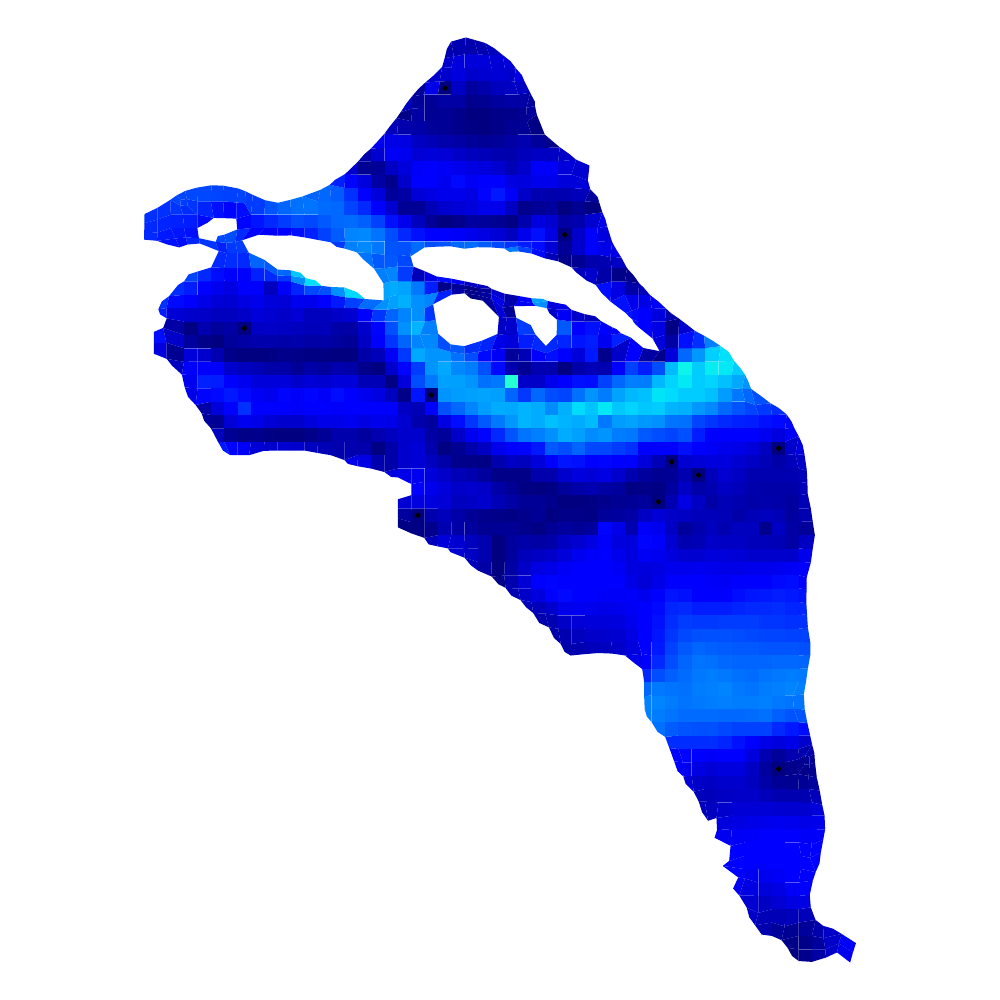} & \\
    \end{tabu}
        \caption{The coarse-resolution ($N_{FV}=1475$) RF2 reference $y$ field and the point errors in the PICKLE and MAP estimates of the $y$ field as functions of $N_{\mathbf{y}_{\mathrm{s}}}$ given the unknown Neumann boundary conditions. The dots are the locations of the 100 observations of $y$.}
    \label{fig:hanford_1x}
\end{figure}
\begin{comment}
\begin{figure}[!htbp]
    \centering
    \includegraphics{pickle-hanford-paper/figures/Fig_Ydiff_RF2_1x.pdf}
    \caption{The coarse-resolution ($N_{FV}=1475$) RF2 reference $y$ field and the point errors in the PICKLE and MAP estimates of the $y$ field as functions of $N_{\mathbf{y}_{\mathrm{s}}}$ given the unknown Neumann boundary conditions. The dots are the locations of the 100 observations of $y$.}
    \label{fig:hanford_1x}
\end{figure}
\end{comment}

Figure~\ref{fig:hanford_1x} shows the RF2 reference $y$ field and the point errors in the PICKLE and MAP estimates of the $y$ field on the coarse mesh obtained with $N_{\mathbf{y}_{\mathrm{s}}}=10$, 25, 50, and 100 for the unknown Neumann boundary conditions. The locations of $y$ measurements  are randomly selected from the well locations. 
Table~\ref{tab:RF2_1x_flux_results} lists the ranges of  $\ell_2$ and $\ell_\infty$ errors in the $y$ estimates as functions of $N_{\mathbf{y}_{\mathrm{s}}}$ obtained with the PICKLE, GPR, and MAP methods. For each $N_{\mathbf{y}_{\mathrm{s}}}$, 10 different random configurations of the measurement locations are selected to compute the ranges. Subtables~\subref{tab:RF2_1x_unknown_flux_results} and \subref{tab:RF2_1x_known_flux_results} give results for unknown and known Neumann boundary conditions, respectively. 
PICKLE's $\ell_2$ errors  are smaller than those in MAP for  $N_{\mathbf{y}_{\mathrm{s}}}=50$  and 100. For $N_{\mathbf{y}_{\mathrm{s}}}=10$  and 25, the lower bounds of $\ell_2$ errors are smaller in PICKLE and the upper bounds are smaller in MAP. 
The absolute $\ell_\infty$ errors follow the same pattern as the $\ell_2$ errors. 

For this coarse resolution, the execution time of PICKLE is larger than that of MAP.
Both PICKLE and MAP perform well for unknown Neumann boundary conditions with estimation errors being only slighter larger than those in the case with known Neumann boundary conditions.   
The $\ell_2$ and $\ell_\infty$ errors in estimating the RF2 field are significantly smaller than those in estimating the RF1 field, which is not surprising given the relative smoothness of the RF2 field. 
For the same reason, the execution time of both PICKLE and MAP methods is significantly smaller for modeling measurements from RF2 than RF1. 

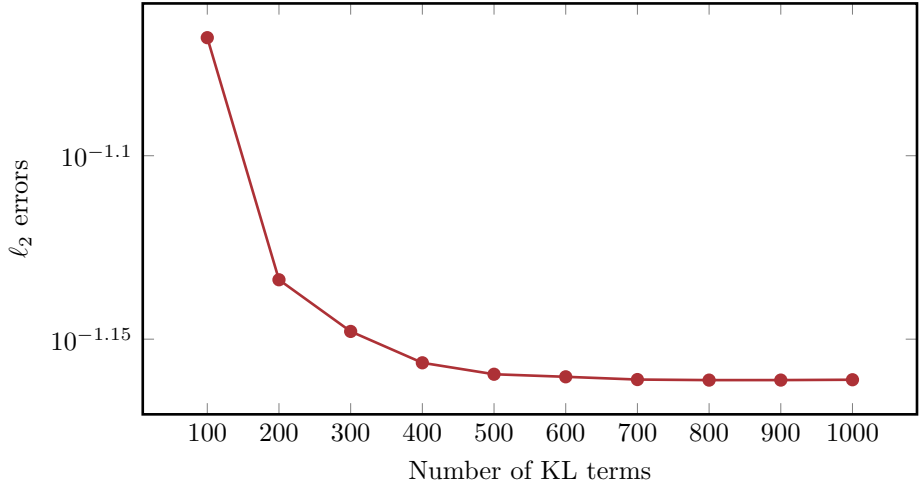
\begin{figure}[!htbp]
    \centering
    \begin{tikzpicture}
        \begin{semilogyaxis}[error plot]
            \pgfplotstableread{figures/rel_errors_vs_kl.txt}{\tabledata}
            \addplot table [x=kl, y=rel_errors] {\tabledata};
        \end{semilogyaxis}
    \end{tikzpicture}
    \caption{Relative $\ell_2$ errors versus the number of KL terms.}
    \label{fig:rel_errors_vs_kl}
\end{figure}
\begin{comment}
\begin{figure}[!htbp]
    \centering
    \includegraphics{pickle-hanford-paper/figures/Fig_L2_vs_KL.pdf}
    \caption{Relative $\ell_2$ errors versus the number of KL terms.}
    \label{fig:rel_errors_vs_kl}
\end{figure}
\end{comment}
Next, we study the relative $\ell_2$ error in the PICKLE solution for $y(x)$ as a function of $N_y$ and $N_u$, the number of terms in the CKLE of $y$ and $u$, respectively, for $N_{\mathbf{y}_{\mathrm{s}}} = 100$.
For simplicity, we set $N_y=N_u$.
Figure~\ref{fig:rel_errors_vs_kl} shows that error decreases as $N_y$ increases and, for the considered RF2 field, reaches the asymptotic value of less than 0.07 at $N_y \approx 800$.
Therefore, the $\text{rtol}_y$ on the order of $10^{-6}$, which is used in this, and the corresponding $N_y=1000$ are  sufficient to obtain an accurate approximating of the RF2 $y$ field and the corresponding reference $u$ field. We note that for the (diffusion-type) Darcy equation, the solution $u(x)$ is always smoother than the parameter field $y(x)$. Therefore, the computational cost of PICKLE can be reduced  by setting $\text{rtol}_u = \text{rtol}_y$, which for diffusion equations would result in $N_u < N_y$.

\begin{table}[!htbp]
    \caption{Performance of PICKLE and MAP for estimating the fine-resolution ($N_{FV}=5900$) RF2 as functions of $N_{\mathbf{y}_{\mathrm{s}}}$ with
  \subref{tab:RF2_4x_unknown_flux_results} unknown and \subref{tab:RF2_4x_known_flux_results} known Neumann boundary conditions.}
    \label{tab:RF2_4x_flux_results}
    \begin{subtable}{\textwidth}
        \caption{Unknown Neumann boundary conditions}
        \label{tab:RF2_4x_unknown_flux_results}
        \begin{tabu} to \textwidth {*{2}{r}c*{3}{X[cm]}}
            \toprule
            & & \multicolumn{4}{c}{$N_{\mathbf{y}_{\mathrm{s}}}$} \\
            & solvers & 100 & 50 & 25 & 10\\
            \midrule
            \multirow{2}{*}{\shortstack[r]{least square\\ iterations}} & PICKLE & 9--11 & 11--15 & 11--16 & 14--35 \\
            & MAP & 69--205 & 78--199 & 102--234 & 210--288 \\
            \midrule
            \multirow{2}{*}{\shortstack[r]{execution\\ time (s)}} & PICKLE & 206--265 & 196--217 & 188--203 & 208--290 \\
            & MAP & 4072--8186 & 3977--10121 & 6031--12374 & 12459--16944 \\
            \midrule
            \multirow{3}{*}{\shortstack[r]{relative\\ $\ell_2$ error}} & GPR & 0.103--0.162 & 0.219--0.395 & 0.349--0.503 & 0.458--0.846 \\
            & PICKLE & 0.0211-0.0650 & 0.0560--0.150 & 0.0681--0.323 & 0.205--0.915 \\
            & MAP & 0.119--0.140 & 0.157--0.188 & 0.179--0.247 & 0.211--0.336 \\
            \midrule
            \multirow{3}{*}{\shortstack[r]{absolute\\ $\ell_\infty$ error}} & GPR & 0.80--2.10 & 2.18--3.80 & 2.06--4.19 & 2.95--4.40 \\
            & PICKLE & 0.348--0.834 & 0.476--2.04 & 0.715--2.32 & 1.70--4.19 \\
            & MAP & 0.877--1.30 & 0.961--1.48 & 1.18--1.55 & 1.32--1.53 \\
            \bottomrule
        \end{tabu}
    \end{subtable}\\
    \vspace{1em}%
    \begin{subtable}{\textwidth}
        \caption{Known Neumann boundary conditions}
        \label{tab:RF2_4x_known_flux_results}
        \begin{tabu} to \textwidth {*{2}{r}c*{3}{X[cm]}}
            \toprule
            & & \multicolumn{4}{c}{$N_{\mathbf{y}_{\mathrm{s}}}$} \\
            & solvers & 100 & 50 & 25 & 10\\
            \midrule
            \multirow{2}{*}{\shortstack[r]{least square\\ iterations}} & PICKLE & 8--11 & 11--15 & 11--15 & 14--37 \\
            & MAP & 86--129 & 84--136 & 69--128 & 65--109 \\
            \midrule
            \multirow{2}{*}{\shortstack[r]{execution\\ time (s)}} & PICKLE & 192--223 & 192--217 & 181--200 & 193--200 \\
            & MAP & 4520--6330 & 4228--7124 & 6277--10348 & 6096--9333 \\
            \midrule
            \multirow{2}{*}{\shortstack[r]{relative\\ $\ell_2$ error}} & PICKLE & 0.0193--0.0526 & 0.0497--0.145 & 0.0638--0.324 & 0.154--0.594 \\
            & MAP & 0.104--0.128 & 0.121--0.168 & 0.150--0.211 & 0.179--0.308 \\
            \midrule
            \multirow{2}{*}{\shortstack[r]{absolute\\ $\ell_\infty$ error}} & PICKLE & 0.302--0.823 & 0.476--1.93 & 0.709--2.29 & 1.65--3.91 \\
            & MAP & 0.922--1.37 & 0.900--1.62 & 1.06--1.69 & 1.41--1.71 \\
            \bottomrule
        \end{tabu}
    \end{subtable}
\end{table}

\begin{figure}[!htbp]
    \centering
    \begin{tabu} to \textwidth {X[cm]|c*{2}{X[cm]}c}
        reference & $N_{\mathbf{y}_{\mathrm{s}}}$ & $|\text{PICKLE} - \text{reference}|$ & $|\text{MAP} - \text{reference}|$ & \\
        \multirow{4}{*}{\includegraphics[scale=0.33]{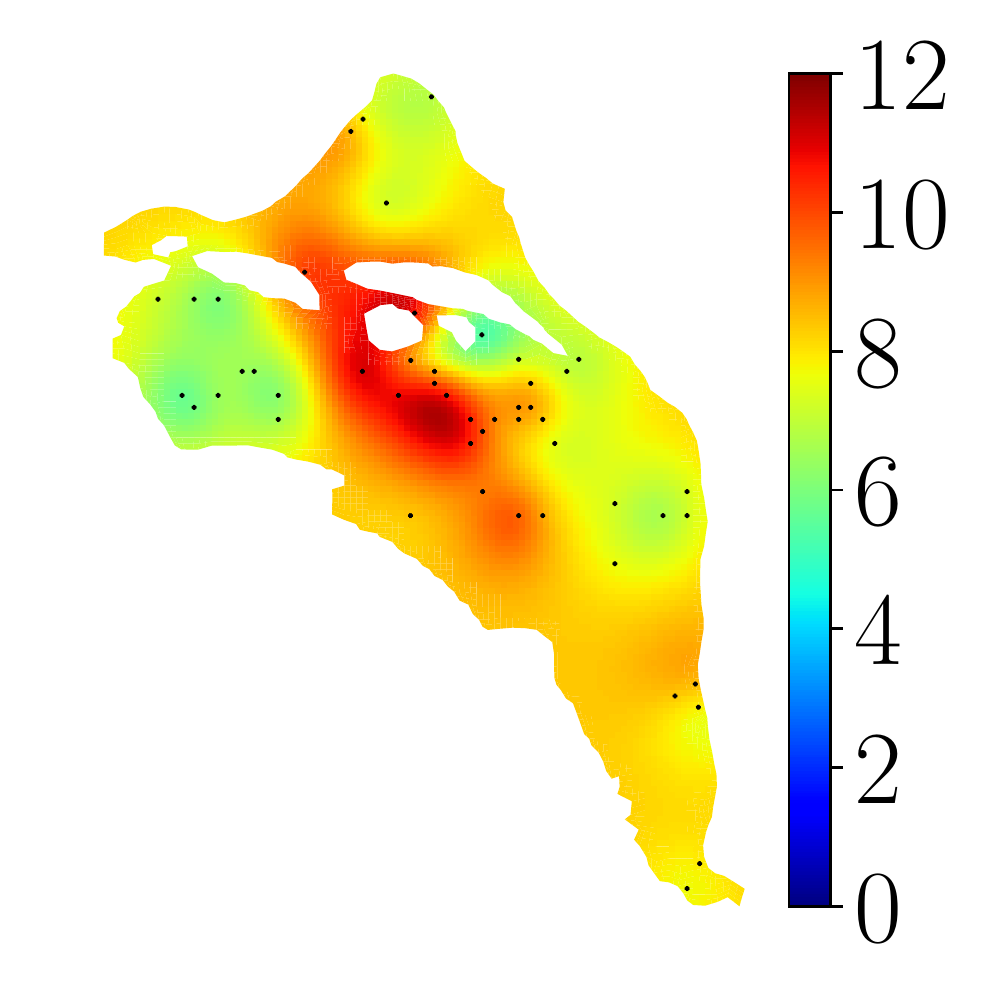}} & 100 &
        \includegraphics[scale=0.33]{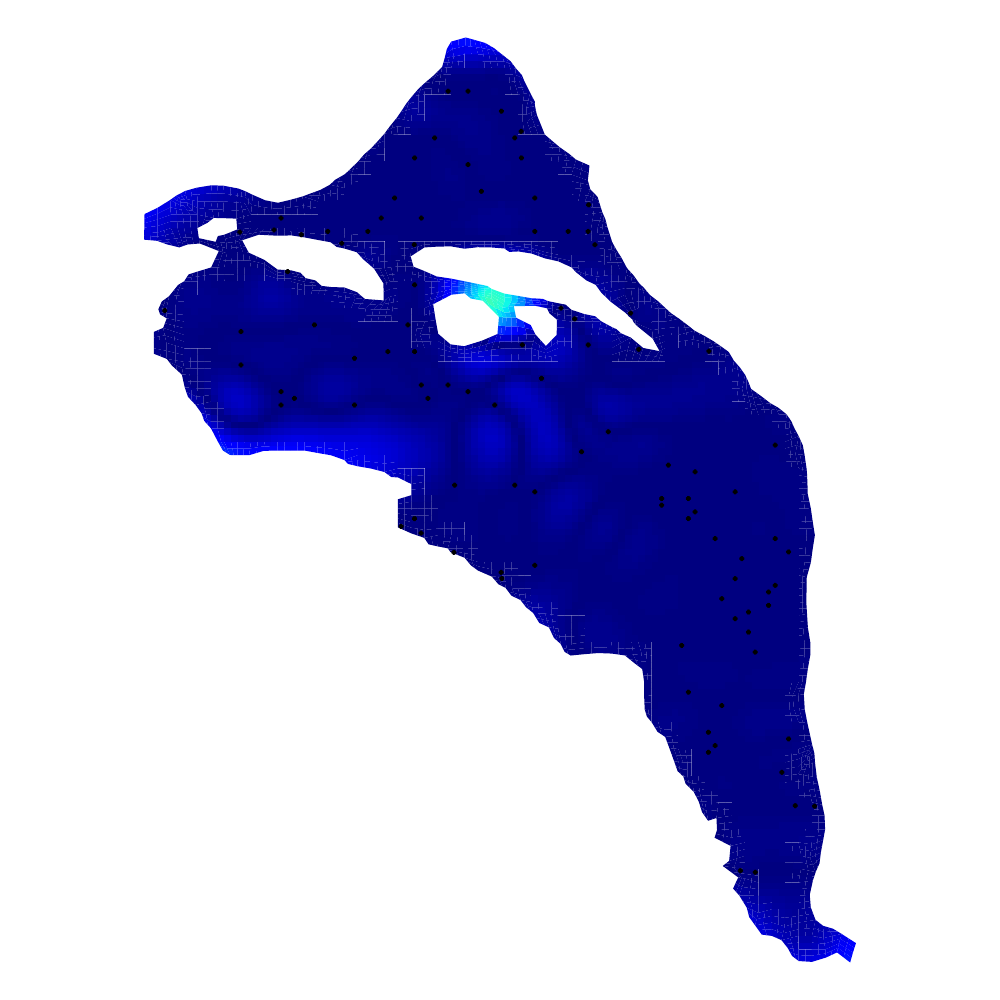} &
        \includegraphics[scale=0.33]{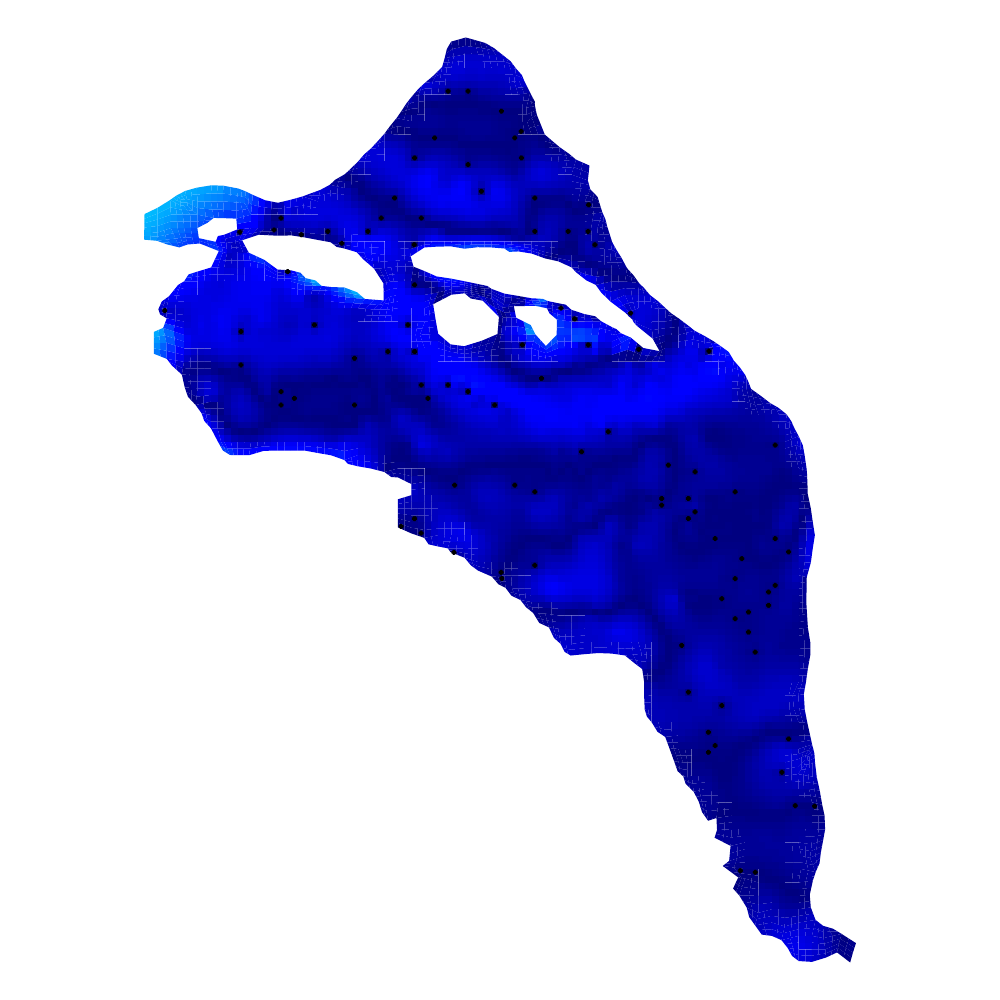} &
        \multirow{4}{*}{\includegraphics[scale=0.44]{{figures/Ydiff_colorbar_RF2}.pdf}} \\
        & 50 &
        \includegraphics[scale=0.33]{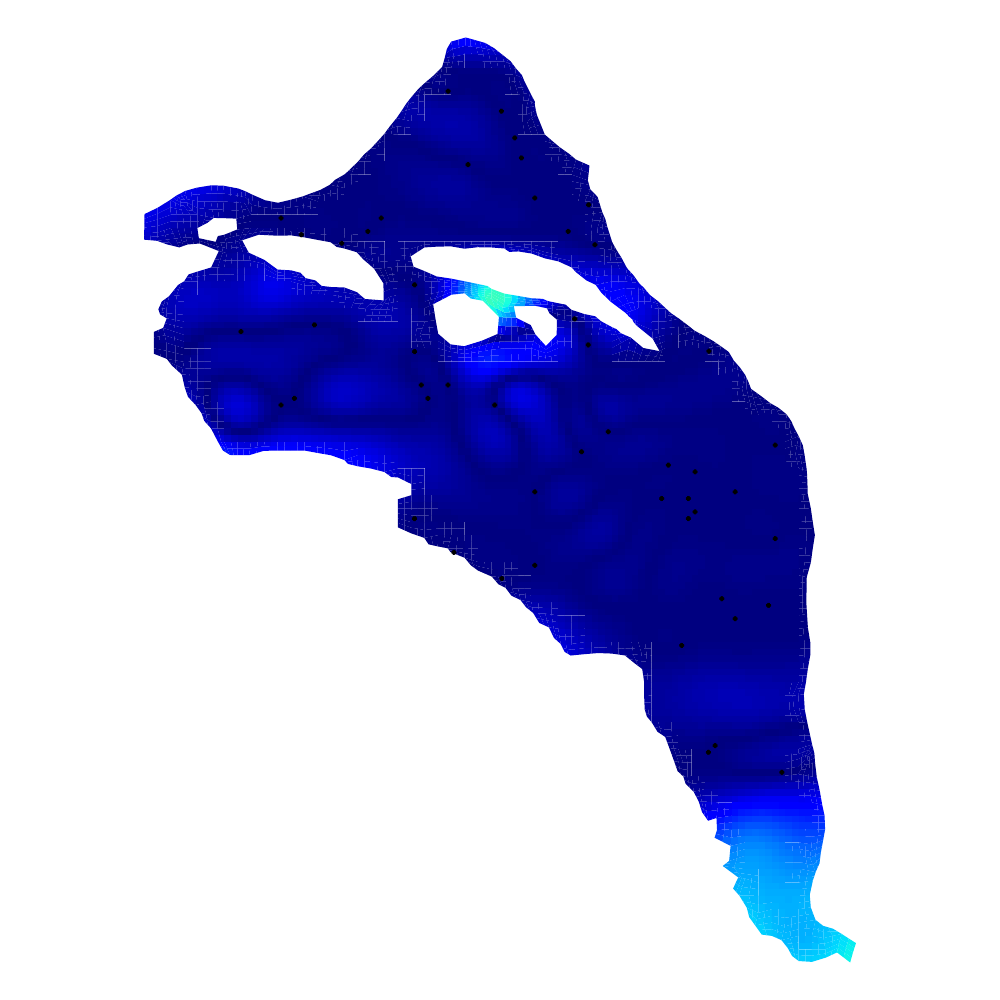} &
        \includegraphics[scale=0.33]{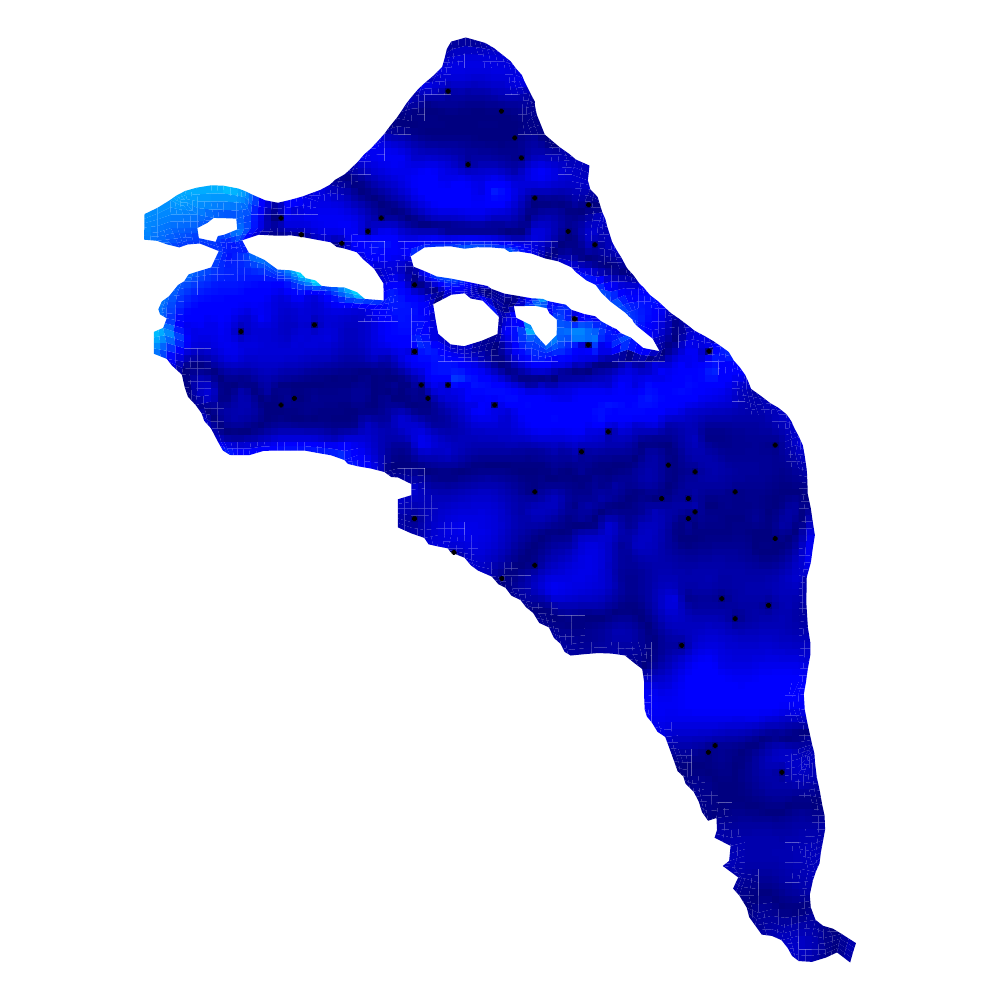} &
        \\
        & 25 &
        \includegraphics[scale=0.33]{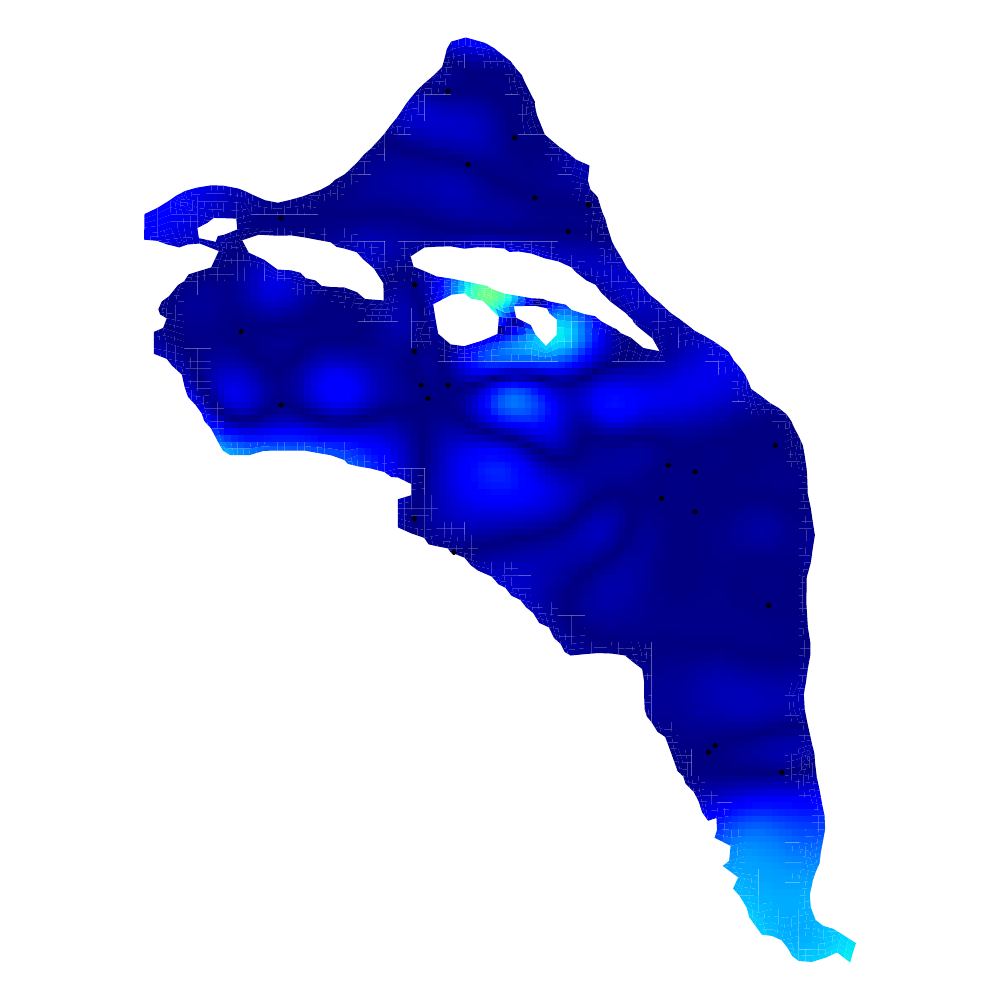} &
        \includegraphics[scale=0.33]{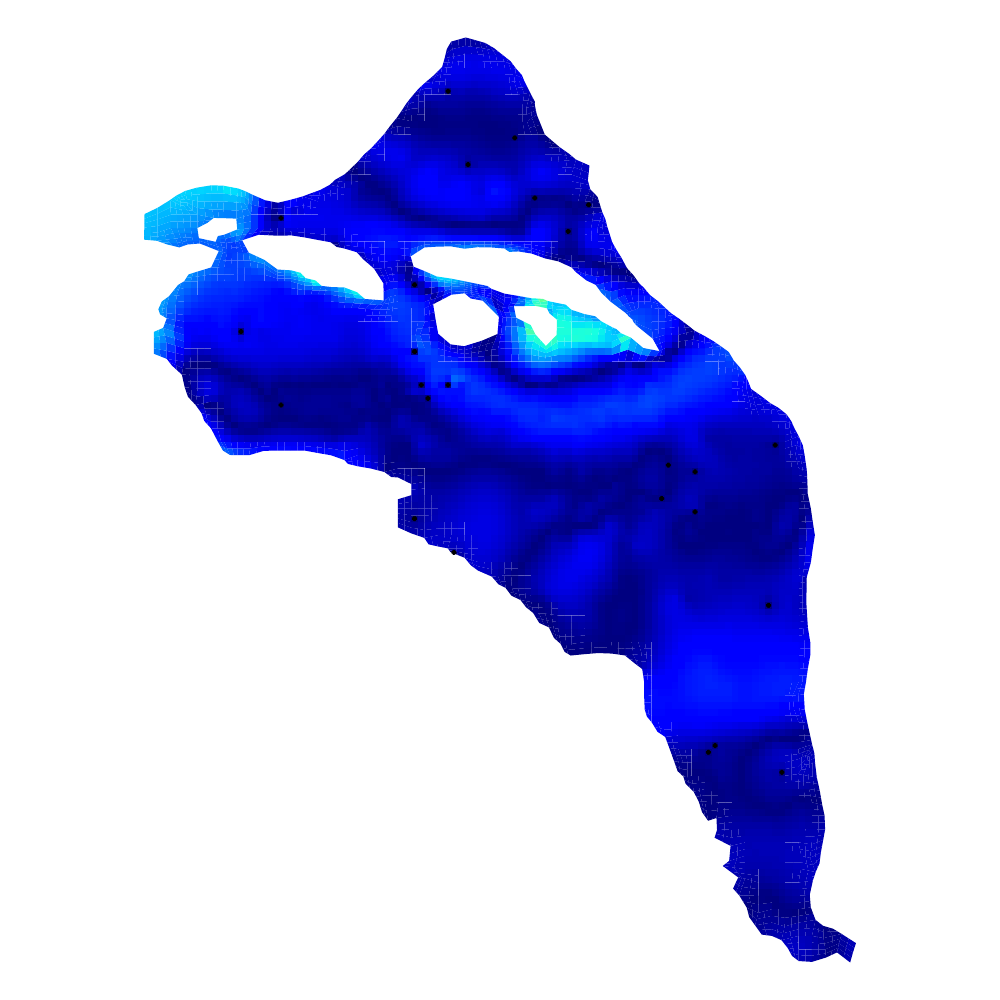} &
        \\
        & 10 &
        \includegraphics[scale=0.33]{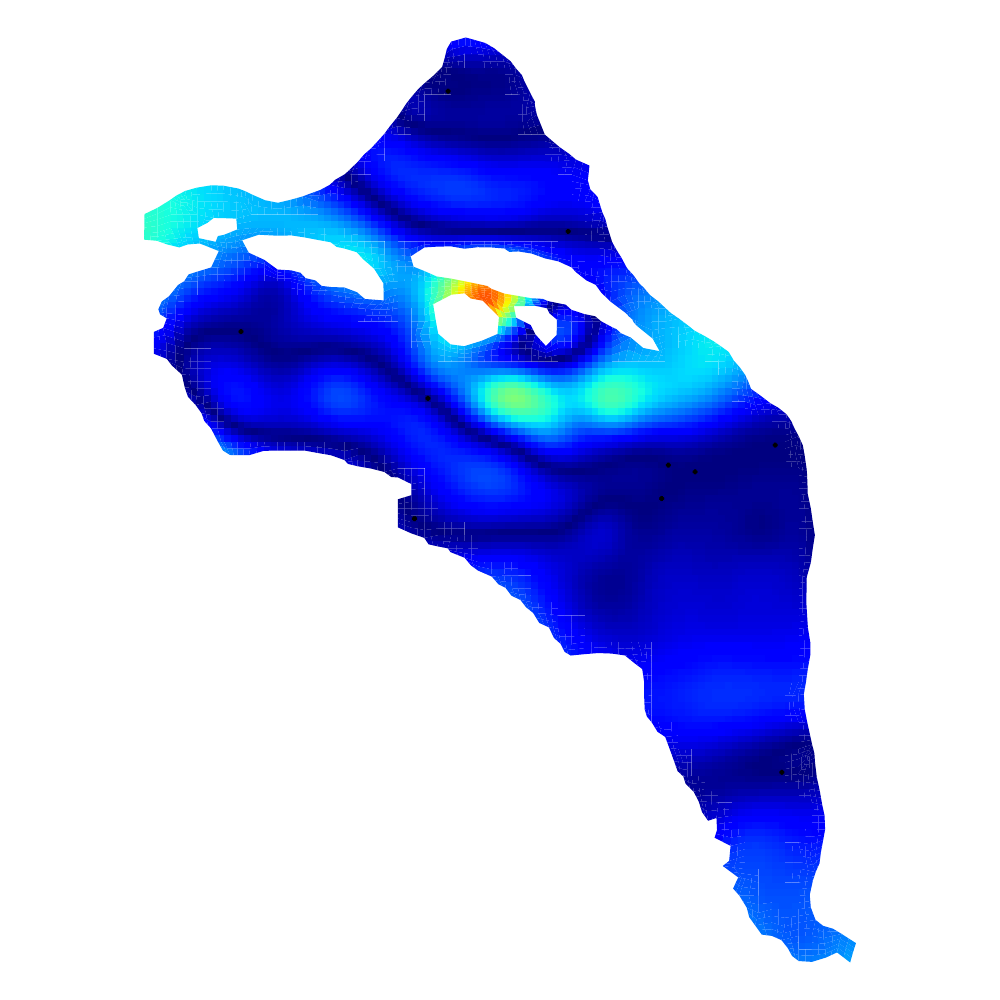} &
        \includegraphics[scale=0.33]{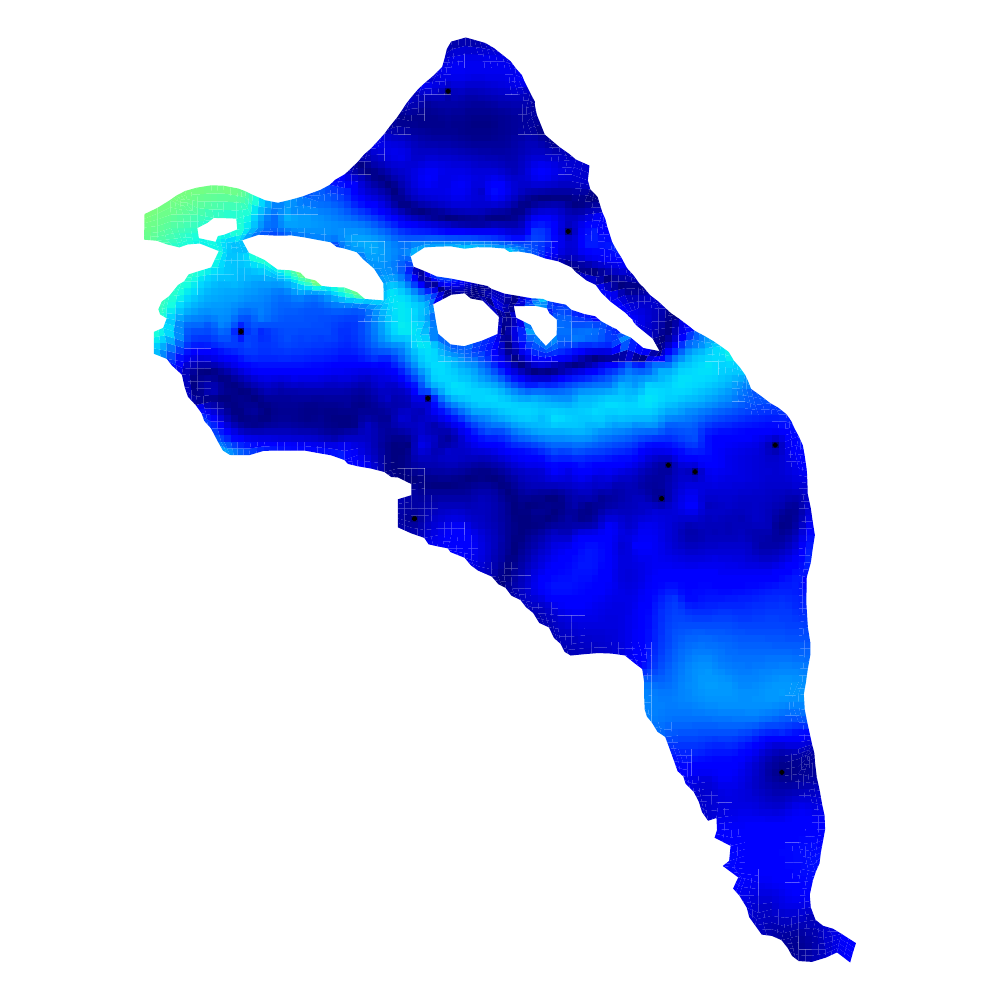} & \\
    \end{tabu}
    \caption{The fine-resolution ($N_{FV}=5900$) RF2 reference $y$ field and the point errors in the PICKLE and MAP estimates of the $y$ field as functions of $N_{\mathbf{y}_{\mathrm{s}}}$  given the unknown Neumann boundary conditions. The dots are the locations of the 100 observations of $y$.}
    \label{fig:hanford_4x}
\end{figure}
\begin{comment}
\begin{figure}[!htbp]
    \centering
    \includegraphics{pickle-hanford-paper/figures/Fig_Ydiff_RF2_4x.pdf}
    \caption{The fine-resolution ($N_{FV}=5900$) RF2 reference $y$ field and the point errors in the PICKLE and MAP estimates of the $y$ field as functions of $N_{\mathbf{y}_{\mathrm{s}}}$  given the unknown Neumann boundary conditions. The dots are the locations of the 100 observations of $y$.}
    \label{fig:hanford_4x}
\end{figure}
\end{comment}

Finally, we test the relative performance of the PINN and MAP methods as a function of the resolution of the flow model by estimating $y$ and $u$ using the finer mesh with $N_{VF} = 5900$.
Table~\ref{tab:RF2_4x_flux_results} lists the ranges of $\ell_2$ and $\ell_\infty$ errors in the PICKLE, GPR, and MAP estimates of $y$ as functions of $N_{\mathbf{y}_{\mathrm{s}}}$ as well as the execution times obtained from 10 different random distributions of measurements for each value of $N_{\mathbf{y}_{\mathrm{s}}}=10$. At this resolution,  PICKLE is more accurate than MAP for most considered configurations and numbers of measurements.  \cref{tab:RF2_4x_unknown_flux_results,tab:RF2_4x_known_flux_results} show results for unknown and known boundary conditions, respectively.  
Figure~\ref{fig:hanford_4x} shows the RF2 $y$ field with the resolution $N_{VF} = 5900$ and the point errors in the PICKLE and MAP estimates of this $y$ field obtained with $N_{\mathbf{y}_{\mathrm{s}}}=10$, 25, 50, and 100 and unknown flux boundary conditions.
It follows from Table~\ref{tab:RF2_4x_flux_results} that 
the PICKLE $\ell_2$ errors are smaller than those in MAP except the upper ranges of the errors for $N_{\mathbf{y}_{\mathrm{s}}}=10$ and 25. The lower bound of $\ell_\infty$ errors is lower in the PICKLE method (except for $N_{\mathbf{y}_{\mathrm{s}}}=10$) while  the upper bound is larger (except for $N_{\mathbf{y}_{\mathrm{s}}}=100$). The errors for unknown Neumann boundary conditions are slightly larger in both methods than those in the case with known Neumann boundary conditions.

\subsection{Scaling of the execution time with the problem size} 
\begin{figure}[!htbp]
    \centering
    \begin{tikzpicture}
        \begin{loglogaxis}[trend plot]
            \pgfplotstableread{figures/time_vs_size.txt}{\tabledata}
            \newcounter{yi}
            \pgfplotsinvokeforeach {RF1_PICKLE_t, RF1_MAP_t, RF2_PICKLE_t, RF2_MAP_t} {
                \pgfmathsetmacro{\nodepos}{ifthenelse(\the\value{yi}<2,"above","below")}
                \setv{nodepos}{#1}{\nodepos}
                \addplot+ table [x=cells, y=#1] {\tabledata};
                \label{plot:#1}
                \addplot+[mark=none] table [x=cells, y={create col/linear regression={y=#1}}] {\tabledata}
                    node[\getv{nodepos}{#1}, pos=.75, sloped] {
                        $\pgfmathprintnumber{\getv{intercept}{#1}} x^{\pgfmathprintnumber{\getv{slope}{#1}}}$
                    }
                    coordinate[pos=0] (A_#1) coordinate[pos=.5] (B_#1) coordinate[pos=1] (C_#1);
                \setv{slope}{#1}{\pgfplotstableregressiona}
                \pgfkeys{/pgf/fpu=true}
                \pgfmathparse{exp(\pgfplotstableregressionb)}
                \setv{intercept}{#1}{\pgfmathresult}
                \pgfkeys{/pgf/fpu=false}
                \pgfmathaddtocounter{yi}{1}
            }
            \pgfplotsinvokeforeach {RF1_MAP_t, RF2_MAP_t} {
                \draw[Maroon] (A_#1) -- (B_#1);
                \draw[Maroon, dashed] (B_#1) -- (C_#1);
            }
            \coordinate (legend) at (axis description cs:.2,.7);
        \end{loglogaxis}
        \matrix [
            matrix of nodes,
            font=\small,
            anchor=south,
        ] at (legend) {
             & PICKLE & MAP \\
            RF1 & \ref{plot:RF1_PICKLE_t} & \ref{plot:RF1_MAP_t} \\
            RF2 & \ref{plot:RF2_PICKLE_t} & \ref{plot:RF2_MAP_t} \\
        };
    \end{tikzpicture}
    \caption{Execution times of PICKLE and MAP versus the number of FV cells for the RF1 and RF2 reference fields. The execution times of MAP for the mesh with 23600 FV cells are estimated by extrapolation.}
    \label{fig:time}
\end{figure}
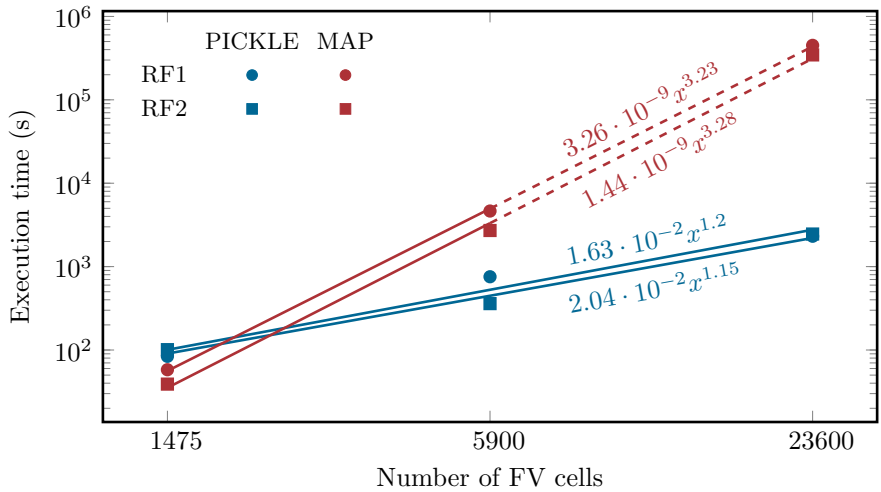

The comparison of Tables~\ref{tab:RF2_1x_flux_results} and \ref{tab:RF2_4x_flux_results} shows that 
the execution times of both PICKLE and MAP increase with the mesh resolution; however, the execution time of PICKLE increases slower than that of MAP.
To further study the dependence of the computational cost of  PICKLE and MAP, in Figure~\ref{fig:time} we plot the execution time of these methods as functions of $N_{FV}$ for both the RF1 and RF2 reference fields.  For these results, a mesh with  $N_{FV}=23600$ FV cells is generated by dividing each cell in the mesh with $N_{FV}=5900$) in four. The number of $y$ measurements in all simulations is set to  $N_{\mathbf{y}_{\mathrm{s}}}=100$.
Figure~\ref{fig:time} also shows the power-law model fits for both methods. We note that for $N_{FV}=23600$, the MAP method did not converge after running for two days. Therefore, the power law relationships for the MAP method are obtained based on the execution times for $N_{FV}=1475$ and $5900$ and used to estimate MAP's execution times for the highest resolution. 
From Figure~\ref{fig:time}, we see that the PICKLE and MAP execution times increase as $N_{FV}^{1.1}$ and $N_{FV}^{3.2}$, respectively, for both the RF1 and RF2 fields.
The close to linear dependence of PICKLE's execution time on the problem size gives it a computational advantage over the MAP method. 

\begin{comment}
\begin{figure}[!htbp]
    \centering
    \includegraphics{pickle-hanford-paper/figures/Fig_time_vs_size.pdf}
    \caption{Execution times of PICKLE and MAP versus the number of FV cells for the RF1 and RF2 reference fields. The execution times of MAP for the mesh with 23600 FV cells are estimated by extrapolation.}
    \label{fig:time}
\end{figure}
\end{comment}

\subsection{Modeling $u$ and $y$ measurements corresponding to arbitrary boundary conditions}\label{sec:varying_Dirichlet}

In many natural systems such as the Hanford Site, boundary conditions can change with time.
Once ``trained'' for one value of the boundary conditions, PICKLE can be used without additional retraining to assimilate data corresponding to any boundary condition, i.e., $C_u(x',x'')$, that is calculated from the MC simulations for certain boundary conditions ($q_\mathcal{N}^*(x), u_\mathcal{D}^*(x)$) can be used to estimate $y$ and $u$ using measurements that correspond to any values of ($q_\mathcal{N}(x), u_\mathcal{D}(x)$).
This is because for the deterministic (known) boundary conditions ($q_\mathcal{N}(x), u_\mathcal{D}(x)$),  $C_u(x',x'')$ does not depend on the values of ($q_\mathcal{N}(x), u_\mathcal{D}(x)$).
%\note[David]{Is this true? I thought that the dependence was weak but was there.}
For stochastic boundary conditions, $C_u(x',x'')$ depends only on the covariances of $q_\mathcal{N}(x)$ and $u_\mathcal{D}(x)$ and not on their mean values.

\begin{figure}[!htbp]
    \centering
    \includegraphics[scale=0.4]{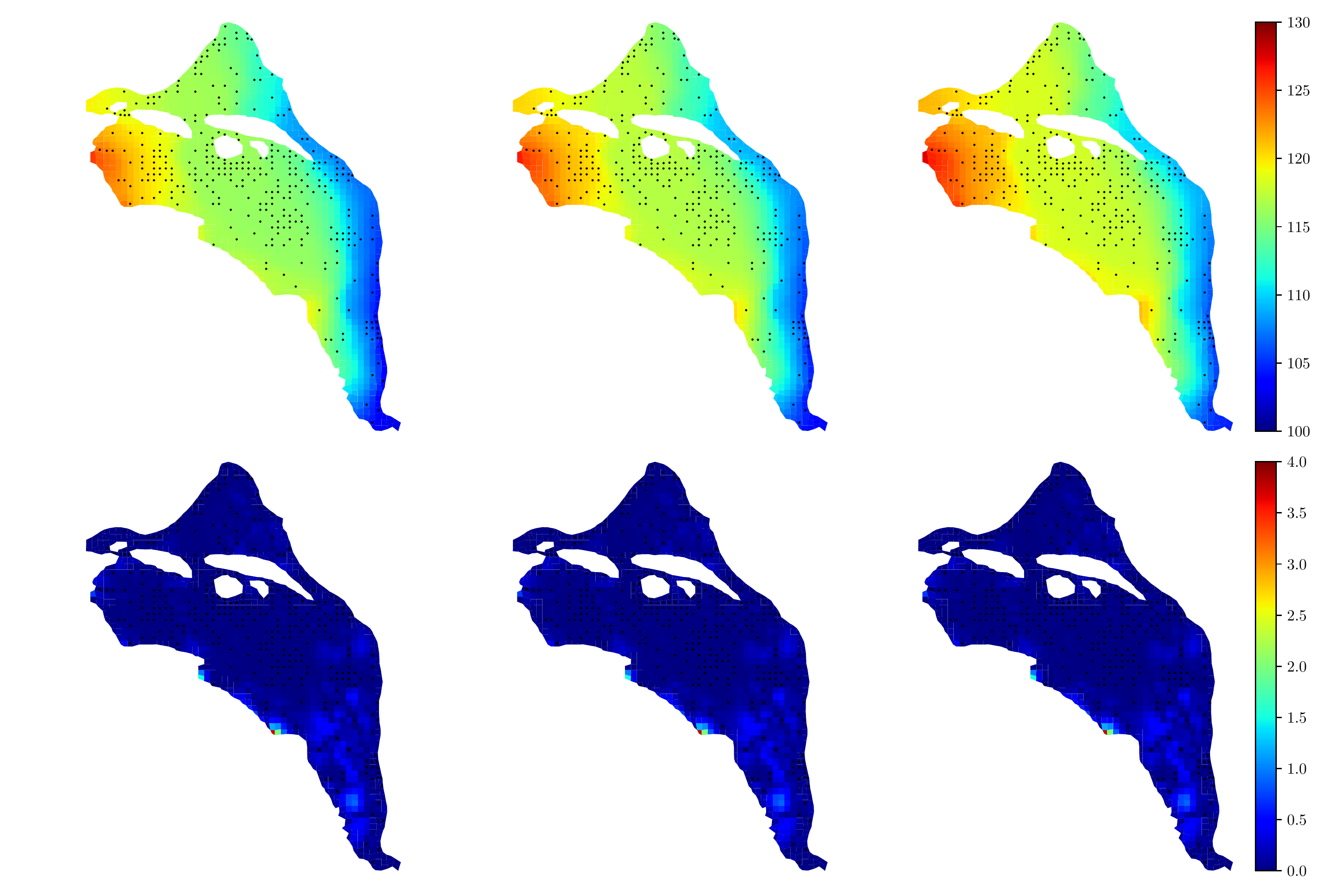}
    \caption{The PICKLE estimates $\hat{u}{(i)}$ ($i=1,2,3$) (top row) and the point errors with respect to the reference fields $u{(i)}$ (bottom row). The dots are locations of the wells where $u$ measurements are available.}
    \label{fig:uref_udiff_BCs}
\end{figure}

As an example, we consider a case where the Dirichlet boundary condition $u_\mathcal{D}(x)$ incrementally changes with time in the range [$u_\mathcal{D}^{min}$, $u_\mathcal{D}^{max}$] in response to the  changes in the water level in the Columbia and Yakima Rivers.
We denote  $u_\mathcal{D}^{(i)}(x)$ ($i=1,...,N_t$) as the Dirichlet BC at each of $N_{t}$ time intervals.
At the $i$th time interval, the $\mathbf{u}_\mathrm{s}^{i}$ measurements are collected at $N_{\mathbf{u}_{\mathrm{s}}}$ spatial locations.
The Neumann boundary conditions are assumed to be statistically known, i.e., the mean and covariance of $q_\mathcal{N}(x)$ are assumed to be known, and these statistical moments are assumed to be constant in time.
The $\mathbf{y}_s$ are available at $N_{\mathbf{y}_{\mathrm{s}}}$ locations ($\mathbf{y}_\mathrm{s}$ do not change in time).
To model $\mathbf{y}_\mathrm{s}$ and $\mathbf{u}_\mathrm{s}^{(i)}$ data, the covariance function $C_u(x',x'')$ can be found from the MC method as described in Section \ref{covariances} with $u_\mathcal{D}(x) = u_\mathcal{D}^*(x)$, where $u_\mathcal{D}^*(x)$ is any one BC from $\{ {u}_{\mathcal{D}}^{(i)} \}_{i=1}^{N_{t}}$.
We emphasize that $C_u(x',x'')$ should be computed only for one BC from $\{ {u}_\mathcal{D}^{(i)} \}_{i=1}^{N_{t}}$ and then can be applied to any boundary condition from $\{ {u}_\mathcal{D}^{(i)} \}_{i=1}^{N_{t}}$. 

To apply PICKLE to the $\mathbf{u}_\mathrm{s}^{(i)}$ measurements, in addition to $C_u(x',x'')$ we need to estimate  $\overline u^{(i)}(x)$ that, unlike the covariance, depends on $u_\mathcal{D}^{(i)}$.
Section \ref{covariances} describes the MC method for computing $\overline u^{(i)}(x)$ that could be expensive to perform for each ${u}_\mathcal{D}^{(i)}$ $(i=1,...,N_t)$.
Here, we propose to approximately compute the $\overline u^{(i)}(x)$ from the mean Darcy flow equation
\begin{align}
  \label{eq:pde-mean}
  \nabla \cdot \left [ \overline{T}^c(x) \nabla \overline u^{(i)}(x) \right ] & = 0, && x \in D,\\
  \label{eq:pde-flux-bc-mean}
  \overline{T}^c(x) \nabla \overline u^{(i)}(x) \cdot n(x) &= -\overline{q}_\mathcal{N}(x), && x \in \Gamma_\mathcal{N},\\
  \label{eq:pde-head-bc-mean}
  \overline{u}^{(i)}(x) &= u_\mathcal{D}^{(i)}(x), && x \in \Gamma_\mathcal{D},
\end{align}
where $\overline{T}^c = \exp (\overline{y}^c)$, and $\overline{y}^c(x)$ is given by Eq.~(\ref{eq:gpr-mean}).
Eq.~(\ref{eq:pde-flux-bc-mean}) is an approximation to the MC solution for $\overline{y}^c(x)$ that disregards the term $\overline{\nabla \cdot \left [ T'(x) \nabla  u^{(i)'}(x) \right ]}$, where $T'(x) = T^c(x)-\overline{T}^c(x)$ and $u^{(i)'}(x) = u^{(i)}(x) - \overline{u}^{(i)}(x)$.  

To test the proposed approximate PICKLE model, we assume that the reference $y$ field is given by the RF1 field from which we draw $N_{\mathbf{y}_{\mathrm{s}}}=100$ measurements of $y$ at random locations. Furthermore, we assume that $u(x)$ is sampled at $N_{\mathbf{u}_{\mathrm{s}}}=323$ locations, and that three measurements of $u(x)$ are available at each location forming three vectors of $u$ measurements $\mathbf{u}_\mathrm{s}^{(i)}$ ($i=1,2,3$).
The corresponding boundary conditions $u_\mathcal{D}^{(i)}(x)$ ($i=1,2,3$)  are constructed as follows: $u_\mathcal{D}^{(2)}(x)$ is given by the calibration study \citep{cole2001transient}, $u_\mathcal{D}^{(1)}(x) = u_\mathcal{D}^{(2)}(x) - 1$m, and $u_\mathcal{D}^{(3)}(x) = u_\mathcal{D}^{(2)}(x) + 1$m.
Three reference fields $u^{(i)}(x)$ are computed by solving Eqs.~(\ref{eq:pde})--(\ref{eq:pde-head-bc}) with $y(x)$ given by the RF1 $y$ field and subject to the Dirichlet BCs ${u}_D^{i}$ ($i=1,2,3)$.
The vector $\mathbf{u}_\mathrm{s}^{i}$ is drawn from the reference field $u^{(i)}(x)$. 
We compute $C_u(x',x'')$ (and $\overline{u}^{(2)}$) from MCS for $u_\mathcal{D}(x) = u_\mathcal{D}^{(2)}(x)$. 
The mean $u$ fields $\overline{u}^{(1)}$ and $\overline{u}^{(3)}$ are approximately computed from Eqs.~(\ref{eq:pde-mean})--(\ref{eq:pde-head-bc-mean}).
Figure~\ref{fig:uref_udiff_BCs} shows the PICKLE estimates $\hat{u}^{(i)}$ and the corresponding point errors with respect to the reference fields $u^{(i)}$ ($i=1,2,3$).
%\note[David]{Shouldn't this be a table instead? The figures all look the same to the eye}
For all three fields, the errors in the estimated $u$ fields are similar, with the average relative errors less than 0.5\% and maximum point errors less than 4\%. These results show that the PICKLE model trained for one boundary condition ( $u_\mathcal{D}^{(2)}(x)$ in this case) can be used to accurately predict the $u$ field for any boundary condition. 
Note that in general, the PICKLE estimate of $y$ from the $\mathbf{u}_\mathrm{s}^{i}$ and $\mathbf{y}_\mathrm{s}$ measurements could be different because the parameter estimation is an ill-posed problem.
However, for this test problem and the $\mathbf{u}_\mathrm{s}^{i}$ measurements, we find that the PICKLE estimates of $y$ are within 0.01\% of each other.

\section{Discussion and Conclusions}
\label{sec:conclusions}

We proposed the PICKLE method for assimilating data in models with unknown and time changing boundary conditions and used it to estimate the transmissivity and hydraulic head in the two-dimensional steady-state groundwater model of the Hanford Site with incrementally varying-in-time Dirichlet boundary conditions and uncertain Neumann boundary conditions.
The PICKLE method is based on the approximation of unknown parameters and state variables with CKLEs. The CKLE approximation of a field enforces (exactly matches) the field measurements and the covariance structure, that is, it models the field as a realization of the conditional Gaussian field with a prescribed covariance function. 
To test the applicability of CKLE-based approximations for natural systems such as the Hanford Site, we considered two reference log transmissivity fields, both representing the complexity of the Hanford Site.
The first transmissivity field, referred to as the RF1 field, is constructed by depth-averaging the conductivity field obtained in a previous calibration study that did not make any regularization Gaussianity assumptions.
The RF2 reference (natural-log-)transmissivity field was constructed using the Gaussian process regression (or Kriging) based on 50 values of the RF1 (natural-log-)transmissivity field with locations randomly selected from the 323 locations of the wells at the Hanford Site.
By construction, the RF2 field is smoother than the RF1 field. 
The comparison with the MAP method, a standard method for solving inverse problems, for RF1 and RF2 reference fields reveals the following relative advantages and disadvantages of the PICKLE method.
\begin{itemize}
\item 
  For the synthetic data generated with RF1 and RF2 $y$ fields, we demonstrated that the MAP and PICKLE execution times scale with the problem size as $N_{FV}^{1.15}$ and $N_{FV}^{3.27}$, respectively, where $N_{FV}$ is the number of FV cells.
  The close to linear dependence of PICKLE's execution time on the problem size gives PICKLE a computational advantage over the MAP method for large-scale problems. We consider this to be the main advantage of the PICKLE method. 
 \item For the same number of measurements, the accuracy of PICKLE and MAP depends on the measurements locations. The MAP method is on average more accurate for the RF1 field, and the PICKLE method is  more accurate for the RF2 field for most considered cases.   
 \item The execution time of PICKLE and MAP increases and the accuracy decreases as the roughness of the parameter field increases.
%\item As expected, the accuracy of both PICKLE and MAP increases as the number of $y$ and $u$ measurements increases.
\item In the PICKLE method, the execution time and accuracy increase with the increasing number of CKL terms. In this work, as a baseline we used $N_y=N_u=1000$ that corresponds to $\text{rtol}<10^{-6}$. We stipulate that this criterion is sufficient to obtain a convergent estimate of $y$ with respect to the number of CKL terms. 
%criterion is sufficient to obtain ides convergent Optimizing the number of terms in the conditional KL expansions. For example, in the scaling study we used 1000 terms in the CKL expansions of $u$ and $y$an unnecessarily large number of KL terms (in the simulations we used 1000 KL terms while according to Figure~\ref{fig:rel_errors_vs_kl}, 700 terms are sufficient to accurately represent the $y(x)$ and $u(x)$) is included in the CKLE representations of $y$ and $u$. Tables~\ref{tab:low_res_diff_kl_unknown_flux_results} and \ref{tab:low_res_diff_kl_known_flux_results} demonstrate that the computational cost of PICKLE decreases as the number of terms in the CKLE representations of $y$ and $u$ decreases. 
\item The training of the PICKLE model should be performed only for one value of the boundary conditions  and does not need to be updated as the boundary conditions change, which significantly reduces its cost.
\item The accuracy of the PICKLE method depends on the ability of the truncated CKLEs to accurately approximate $y(x)$ and $u(x)$, which requires a certain degree of smoothness of the considered fields. We demonstrated that for $y$ and $u$ fields that are representative of the Handford Site, the CKLE approximations of the fields lead to results that are comparable in accuracy to the MAP method. However,  CKL can also be used to approximate fields exhibiting step-like changes (e.g., at the boundaries of different geological formations) using a logistic function as was shown in~\citep{barajassolano-2019-pickle}.
\item In the PICKLE method, computing the covariance function of $u$ from MCS can become a computational bottleneck for large-scale problems. Two points should be made in this regard: (1) PICKLE's reported execution times in the scalability study (and everywhere else in this work) include the time to perform MCS and the execution time of MAP increases significantly faster than that of PICKLE; and (2) MCS can be replaced with more computationally efficient alternatives, including  the multilevel MC method, generative physics-informed machine learning models, Polynomial Chaos and other surrogate models, and the moment equation method.
\end{itemize}

\section{Acknowledgments}
This research was partially supported by the U.S. Department of Energy (DOE) Advanced Scientific Computing (ASCR) program. Pacific Northwest National Laboratory is operated by Battelle for the DOE under Contract DE-AC05-76RL01830. The data and codes used in this paper are available at  \url{https://github.com/yeungyh/pickle.git}.

\appendix
\section{Finite Volume Discretization}\label{sec:FVD}

\begin{figure}
    \centering
    \begin{tikzpicture}[
            thick, scale=1.5,
            centroid/.style={circle, fill, inner sep=0pt, minimum size=2pt}
        ]
        \coordinate (N1) at (-1, 1.5);
        \coordinate (N2) at (0, 1);
        \coordinate (N3) at (1, 1.2);
        \coordinate (N4) at (-2, -0.5);
        \coordinate (N5) at (0, -1);
        \coordinate (N6) at (2, -0.8);
        \coordinate (M) at ($(N2)!0.5!(N5)$);
        \draw (N2) -- (N1) -- (N4) -- node[above, sloped] {cell $i$} (N5) -- node[above, sloped] {cell $j$} (N6) -- (N3) -- cycle;
        \draw[very thick, OliveGreen] (N2) -- node[near end, right] {$e_{i, j}$} (N5);
        \node[centroid, Maroon, label={[text=Maroon]below:$p_i$}] (P1) at (barycentric cs:N1=1,N2=1,N4=1,N5=1) {};
        \node[centroid, MidnightBlue, label={[text=MidnightBlue]below:$p_j$}] (P2) at (barycentric cs:N2=1,N3=1,N5=1,N6=1) {};
        \draw[>=latex, Maroon] (P1) edge[->] node[above] {$\vec{c}_{i,j}$} (M) (M) edge[->] node[above] {$\vec{n}_{i,j}$} ++(0.5, 0);
    \end{tikzpicture}
    \caption{Two adjacent cells used to define the two-point finite-volume discretization model.}
    \label{fig:two_cells}
\end{figure}
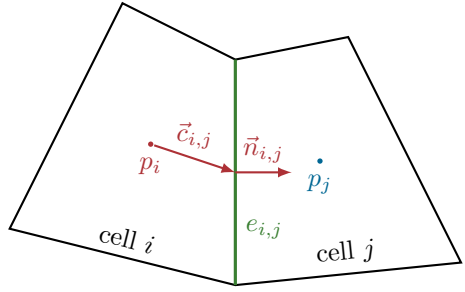
\begin{comment}
\begin{figure}
    \centering
    \includegraphics{pickle-hanford-paper/figures/Fig_TPFV.pdf}
    \caption{Two adjacent cells used to define the two-point finite-volume discretization model.}
    \label{fig:two_cells}
\end{figure}
\end{comment}

Figure~\ref{fig:two_cells} shows two adjacent cells in the finite-volume discretization model with their centers labeled as $p_i$ and $p_j$, respectively.
In this model, we assume that the transmissivity is linear  
within each cell $i$ and its average value $T_i$ is at its center.
The edge shared by the two cells, $e_{i,j}$, has dimension $|e_{i,j}|$.
The stiffness matrix $\mathbf{A}$ in Eq.~(\ref{eq:pde-discretized}) is defined as
\begin{equation}
    \mathbf{A}[i, j] = \begin{cases}
        \sum_k \mathcal{T}_{ik} &\text{if } i = j,\\
        -\mathcal{T}_{ij} &\text{if } i \ne j,
    \end{cases}
\end{equation}
where
\begin{align*}
    \mathcal{T}_{ij} &= \left[\tau_{i,j}^{-1} + \tau_{j, i}^{-1}\right]^{-1},\\
    \tau_{i, j} &= T_i |e_{i,j}|\dfrac{\vec{c}_{i,j}\cdot\vec{n}_{i,j}}{|\vec{c}_{i,j}|^2}.
\end{align*}
and $T_i$ is the transmissivity in cell $i$.
The right-hand side $\mathbf{b}$ describes the boundary conditions, and is defined as
\begin{equation}
    \label{eq:residual_rhs}
    \mathbf{b}[i] = \begin{cases}
        \tau_{i, \Gamma} \mathbf{u}_\mathcal{D}[i] & i\in\Gamma_\mathcal{D},\\
        \mathbf{q}_\mathcal{N}[i] & i\in\Gamma_\mathcal{N},\\
        0 & \text{otherwise},
    \end{cases}
\end{equation}
where $\tau_{i, \Gamma}$ is the transmissivity between cell $i$ and the boundary.

\section{Computing MAP estimates}\label{MAP}

In this work, we compute MAP estimates by recasting the PDE-constrained optimization problem of Eq.~(\ref{eq:pde-constrained-opt-reg}) into an unconstrained nonlinear least-squares problem. 
Specifically, we aim to solve the problem
\begin{equation}
  \label{eq:nonlinear-least-squares}
  \min_{\mathbf{p}} \quad \frac{1}{2} \| \mathbf{f}(\mathbf{p}) \|^2_2,
\end{equation}
with cost vector given by
\begin{equation}
  \label{eq:nonlinear-least-squares-vector}
  \mathbf{f}(\mathbf{p}) =
  \begin{bmatrix}
    \mathbf{u}_{\mathrm{s}} - \mathbf{H}_{\mathbf{u}} \mathbf{u}(\mathbf{y}, \mathbf{q})\\
    \mathbf{y}_{\mathrm{s}} - \mathbf{H}_{\mathbf{y}} \mathbf{y}\\
    \sqrt{\gamma} \, \mathbf{D} \mathbf{y}\\
    \sqrt{\gamma} \, \mathbf{q}
  \end{bmatrix}, \quad
  \mathbf{p} =
  \begin{bmatrix}
    \mathbf{y}\\
    \mathbf{q}
  \end{bmatrix},
\end{equation}
where $\mathbf{H}_{\mathbf{u}}$ and $\mathbf{H}_{\mathbf{y}}$ are the observation matrices in Eq~(\ref{eq:pde-constrained-opt-reg}).

Note that in Eq.~(\ref{eq:nonlinear-least-squares}), we fold the PDE constraint into the cost function by treating $\mathbf{u}$ explicitly as a function of the parameters $\mathbf{p}$.

For the least-square minimization problem, it is necessary to compute the Jacobian matrix of the cost vector with respect to the parameters $\mathbf{p}$, which is given by
\begin{equation}
  \label{eq:nonlinear-least-squares-jac}
  \mathbf{J}(\mathbf{p}) = \frac{\partial \mathbf{f}(\mathbf{p})}{\partial \mathbf{p}} = 
  \begin{bmatrix}
    - \mathbf{H}_{\mathbf{u}} \partial \mathbf{u}(\mathbf{y}, \mathbf{q}) / \partial \mathbf{y} & - \mathbf{H}_{\mathbf{u}} \partial \mathbf{u}(\mathbf{y}, \mathbf{q}) / \partial \mathbf{q}\\
    -\mathbf{H}_{\mathbf{y}} & 0\\
    \sqrt{\gamma} \, \mathbf{D} & 0\\
    0 & \sqrt{\gamma} \, \mathbf{I}
  \end{bmatrix}.
\end{equation}
It can be seen that the first block row of Eq.~(\ref{eq:nonlinear-least-squares-jac}) corresponds to the Jacobian of $\mathbf{u}$ with respect to the parameters $\mathbf{p}$, for which we derive a formula as follows.
Differentiating Eq.~(\ref{eq:pde-discretized}) with respect to the $i$th component of $\mathbf{p}$, we obtain
\begin{equation*}
  \frac{\partial \mathbf{l}}{\partial p_i} + \frac{\partial \mathbf{l}}{\partial \mathbf{u}} \frac{\partial \mathbf{u}}{\partial p_i} = \frac{\partial \mathbf{A}}{\partial p_i} - \frac{\partial \mathbf{b}}{\partial p_i} + \mathbf{A} \frac{\partial \mathbf{u}}{\partial p_i} = 0,
\end{equation*}
therefore,
\begin{equation}
    \label{eq:adjoint}
    \frac{\partial \mathbf{u}}{\partial p_i} = - \mathbf{A}^{-1} \left [ \frac{\partial \mathbf{A}}{\partial p_i} - \frac{\partial \mathbf{b}}{\partial p_i} \right ].
\end{equation}

\section{Solver Optimization}
\label{sec:optimization}

We implemented our solvers for both PICKLE and MAP in Python. In both solvers, TPFA is used as a finite volume model for the forward problem.
Although we did not parallelize the solvers used in this paper, we optimized the codes in several ways as follows.

\subsection{Precomputing matrices}
Because the properties of each cell, the observation locations of $\mathbf{u}_\mathrm{s}$,  $\mathbf{y}_\mathrm{s}$, and the topology of the cell connections are fixed, the structures of the matrices also remain unchanged throughout the least-squares minimization of \cref{eq:pde-constrained-opt-reg,eq:pickle-final}. Thus, the matrices representing these fixed properties including the observation matrices $\mathbf{H}_\mathbf{u}$, $\mathbf{H}_\mathbf{y}$, and the regularization matrix $\mathbf{D}$ can be precomputed in advance. For the stiffness matrix $\mathbf{A}$ in \cref{eq:pde-discretized} and the partial derivatives in the first block row of \cref{eq:nonlinear-least-squares-jac}, although their values change over each minimization iteration, their structures (the positions of nonzero components of the matrices) remain the same, and can be identified in advance. In addition, when the boundary conditions are known and constant in time, the aggregated contribution of the prescribed hydraulic head $u_\mathcal{D}$ and normal flux $q_\mathcal{N}$ to each FV cell $i$---$\mathbf{u}_\mathcal{D}[i]$ and $\mathbf{q}_\mathcal{N}[i]$ in \cref{eq:residual_rhs}---can also be precomputed. For MAP, the second to the fourth block rows and $\mathbf{H}_{\mathbf{u}}$ in the first block row of the Jacobian in \cref{eq:nonlinear-least-squares-jac} are also constant throughout minimization because they only depend on the topology of the mesh. Therefore, these elements can also be precomputed ahead of time.

\subsection{Sparsity}
Sparsity is maintained throughout the evaluations of the objective functions of both PICKLE and MAP, including the residual $\mathbf{l}(\mathbf{u}, \mathbf{y}, \mathbf{q})$ in \cref{eq:pde-discretized}, as well as their corresponding Jacobian matrices. This significantly reduces the storage and computation overhead because the increase in the resolution of the mesh quadruples the size of the matrices. However, the SciPy implementation of the sparse linear solver (\texttt{spsolve}) does not support sparse right-hand-side vectors and matrices. Furthermore, partial solvers that only compute solutions at measurement locations and do not have the ability to reuse sparse structural reordering are not supported by the package. Future optimization using these techniques would further reduce the execution times of both the MAP and PICKLE methods.

\bibliography{pickle-hanford-paper}

\end{document}